\documentclass[journal]{IEEEtran}

\newif\ifArXivVersion
\ArXivVersiontrue

\usepackage{times}
\usepackage[numbers]{natbib}
\usepackage{multicol}
\usepackage[bookmarks=true]{hyperref}
\usepackage{graphicx}
\usepackage{amsmath}
\usepackage{amssymb}
\usepackage{bm}
\usepackage{booktabs}
\usepackage{enumitem}
\usepackage{indentfirst}
\usepackage{makecell}
\usepackage{float}
\usepackage{caption}

\usepackage{multirow}
\usepackage{array}
\newcolumntype{L}[1]{>{\raggedright\arraybackslash}p{#1}}

\usepackage{tabularx}
\usepackage{xltabular}
\newcolumntype{T}{>{\raggedright\setlength{\parskip}{0.5\baselineskip}\arraybackslash}X}

\makeatletter
\let\LT@orig@endlongtable\endlongtable
\def\endlongtable{\crcr\noalign{\kern\z@}\LT@orig@endlongtable}
\newcommand{\entryrule}{\noalign{\ifnum\LT@rows=\z@\nobreak\else\penalty\z@\fi
  \vskip\aboverulesep\hrule\@height\lightrulewidth\nobreak\vskip\belowrulesep}}
\makeatother

\usepackage{xcolor}
\usepackage{soul}
\definecolor{behavioral_gen_color}{HTML}{CCE6FF}
\definecolor{visual_behavioral_gen_color}{HTML}{D4FFEE}
\newcommand{\hlb}[1]{{\sethlcolor{behavioral_gen_color}\hl{#1}}}
\newcommand{\hlvb}[1]{{\sethlcolor{visual_behavioral_gen_color}\hl{#1}}}

\usepackage[colorinlistoftodos]{todonotes}
\usepackage{comment}

\def\BenchmarkName{ContactBench~1.0}

\def\NumAssets{more than 40}
\def\NumTasks{36}
\def\ProjectURL{\href{https://contactbench.github.io/}{https://contactbench.github.io/}}
\def\PublicURL{\href{https://jingyi-xiang.github.io/contactbench}{https://jingyi-xiang.github.io/contactbench}}

\usepackage{graphicx}
\usepackage{capt-of}   

\makeatletter
\let\@oldmaketitle\@maketitle
\renewcommand{\@maketitle}{%
  \@oldmaketitle
  \centering
  {\normalsize
    \ifArXivVersion
    \PublicURL{}
    \else
    \ProjectURL{}
    \fi
  \par}
  \vspace{1.5em}
  \setcounter{figure}{0}
  \includegraphics[width=\textwidth]{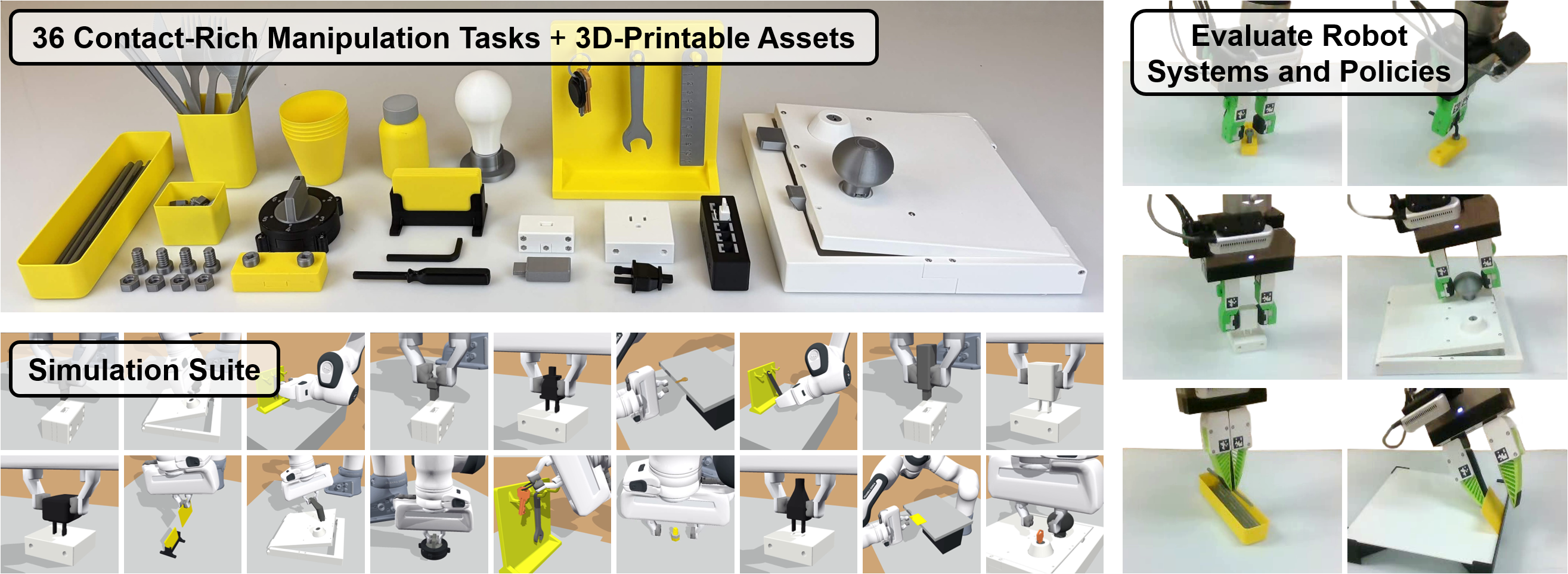}
  \captionof{figure}{We propose ContactBench 1.0, a benchmark for evaluating robot systems and policies on eight contact-rich manipulation skills. ContactBench 1.0 contains 36 tasks, more than 40 3D-printable assets, and their corresponding simulation components.}
  \vspace{-2em}
  \label{fig:teaser}
}
\makeatother

\begin{document}

\title{\BenchmarkName : A Benchmark for Everyday Contact-Rich Robotic Manipulation}
\ifArXivVersion
\author{Jingyi Xiang, Yixuan Wang, Jeannette Bohg, and Wenzhen Yuan%
\thanks{Jingyi Xiang, Yixuan Wang, and Wenzhen Yuan are with the University of Illinois Urbana-Champaign, Urbana, IL, USA (e-mail: \texttt{\{jingyix4, yixuan41, yuanwz\}@illinois.edu}).}
\thanks{Jeannette Bohg is with Stanford University, Palo Alto, CA, USA (e-mail: \texttt{bohg@stanford.edu}).}}
\else
\author{Anonymous Authors}
\fi
\maketitle

\begin{abstract}
Contact-rich manipulation underlies many everyday human activities, yet evaluating robots on such tasks remains difficult: the tasks involve diverse contact behaviors and are hard to reproduce across laboratories. Drawing on observations of everyday human manipulation, we identify eight essential skills that provide a practical framework for studying contact-rich manipulation. We introduce \BenchmarkName{}, an extensible benchmark that instantiates these skills through \NumTasks{} reference tasks and \NumAssets{} object assets. The assets capture task-relevant geometry and contact behavior, can be fabricated reproducibly on consumer-grade 3D printers, and are each paired with a matching simulation model. Controlled task variations in both the real world and simulation enable evaluation of how well robot systems generalize. We evaluate recent representative vision-only and visual-tactile policy learning methods on six benchmark tasks. The results reveal that while vision-only policies struggle to perceive or react to contact events, visual-tactile policies do not meaningfully improve their performance, highlighting that contact-rich manipulation remains a significant challenge for current policy learning methods. By providing common task definitions alongside physical and simulated assets, \BenchmarkName{} offers a shared testbed for developing, evaluating, and comparing contact-rich manipulation systems.
\ifArXivVersion
The benchmark will be open-sourced at \PublicURL{}.
\else
The benchmark will be open-sourced at \ProjectURL{}.
\fi
\end{abstract}



\section{Introduction}
\label{sec: intro}

Achieving human-level generality in robotic manipulation remains a long-standing goal of the robotics community. Recent progress in robot learning has produced increasingly capable systems, supported by large-scale demonstrations, expressive policy architectures, and broad task suites. Much of this progress has been demonstrated on tasks in which visually observable properties, such as object identity, appearance, and pose, provide strong guidance for action. Everyday manipulation, however, also requires robots to regulate forces, exploit environmental constraints, and respond to changes in contact. In these tasks, contact is the mechanism through which the robot accomplishes its goal. Developing robots that can effectively exploit contact is therefore essential to general-purpose manipulation.

This raises a core question: ``What constitutes success in contact-rich manipulation?" Although contact-rich manipulation has been widely studied, many existing works focus on insertion and peg-in-hole tasks~\cite{2-lee2020making, 3-dong2021tactile, 12-kim2022active, 82-higuera2023perceiving, 97-jin2024visual, peginbench} or tasks that require sustained force against a surface, such as wiping and peeling ~\cite{xue2025reactive, liu2025forcemimic, he2025foar}. Everyday human activities, however, involve a much broader range of contact modes, such as grasping challenging objects with various environmental constraints, leveraging contact between the object held in hand and the environment, and manipulating articulated objects based on force feedback. Progress toward general-purpose manipulation therefore requires systematic, reproducible evaluation across a diverse set of tasks that cover this breadth of contact interaction.

We sought answers to this question by observing everyday human manipulation. These observations suggest that, despite differences in objects' appearances and functions, most manipulation sequences involving contact can be broadly described in terms of eight skills:
grasping thin or small objects, grasping multiple objects, grasping from clutter, placing under environmental constraints, in-hand repositioning, extrinsic contact control, articulated object manipulation, and insertion and alignment. Many complex manipulation tasks can be understood as combinations of these skills. We therefore believe that mastering these skills is essential for robots to perform a broad range of contact-rich manipulation tasks.

To support the community's efforts to develop these capabilities, we introduce \BenchmarkName{}, an extensible benchmark that instantiates these skill categories through \NumTasks{} reference tasks and \NumAssets{} standardized object assets. The assets are designed to capture the functional contact interactions of everyday objects through their geometry and mechanical response. Each asset can be fabricated using 3D printing and is designed for reproducible fabrication on consumer-grade printers while preserving task-relevant contact behavior. Controlled variations in object geometry and internal mechanics support evaluation of policy generalization. Corresponding simulation models provide tunable resistance loads for studying related contact interactions in simulation. By providing shared task definitions and standardized physical and simulated assets, \BenchmarkName{} establishes common experimental ground for comparing methods across laboratories and supporting cumulative progress.

We demonstrate the utility of \BenchmarkName{} by evaluating recent vision-only policy learning methods (Diffusion Policy \cite{chi2025diffusion}, $\pi_{0.5}$ \cite{pi05}) and their tactile-augmented versions (Reactive Diffusion Policy \cite{xue2025reactive}, Foundation Tactile Policy \cite{ftp1}) on six tasks from our benchmark. The results reveal mixed performance across the evaluated methods. We observe that vision-only policies struggle to perceive and react to contact-dependent information that is difficult to infer from the vision modality alone. However, adding tactile input did not meaningfully improve performance for the policies we evaluated. Our analysis suggests that limited sensor coverage and difficulty translating tactile observations into appropriate actions may be the bottlenecks. These findings highlight challenges that extend beyond access to tactile sensing alone.

We open-source \BenchmarkName{} at 
\ifArXivVersion
\PublicURL{}.
\else
\ProjectURL{}.
\fi
The initial release focuses on rigid and articulated objects; deformable objects and objects with more challenging surface properties remain beyond its current scope, but will be a potential future extension direction. We envision \BenchmarkName{} as a growing community resource that can incorporate additional tasks, assets, and skills over time. By providing shared infrastructure and a systematic account of contact-rich skills, we aim to make physical interaction a measurable dimension of general-purpose robotic manipulation.


\section{Related Work}
\label{sec: related}





\subsection{Reproducible Physical Assets and Benchmarks}

Physical benchmarks pursue reproducibility in two main ways. Some standardize the hardware: YCB~\cite{calli2015ycb} provides everyday objects with geometric models, NIST task boards~\cite{nist-taskboard} define small-parts assembly tasks and protocols, and ManipulationNet~\cite{chen2026manipulationnet} supports distributed real-world evaluation with standardized hardware kits. Others release printable designs that any lab can fabricate: FurnitureBench~\cite{furniturebench} pairs printable furniture with demonstrations and a simulator for long-horizon assembly, Fabrica~\cite{fabrica} benchmarks dual-arm multi-part assembly, FMB~\cite{fmb} evaluates functional manipulation skills such as grasping, repositioning, and insertion, while POMDAR \cite{pomdar} introduces a benchmark for dexterity. Recent work also varies task conditions in a controlled way. MatchMaker~\cite{matchmaker} generates mating parts with user-specified clearances, and Peg-in-Bench~\cite{peginbench} varies peg geometry, insertion tolerance, spatial layout, and assembly structure. BusyBox~\cite{busybox} instead varies appearance, rearranging modules of buttons, switches, sliders, and other controls while preserving their affordances. Many of these tasks involve assembly or insertion, and the forces they require are either fixed by the asset or set by part clearance. Our benchmark spans a broader range of everyday contact-rich skills, and its printed compliant springs set the required force independently of fit and appearance.

\subsection{Simulation Benchmarks for Manipulation}

Simulation benchmarks support reproducible comparisons across tasks and learning methods. RLBench~\cite{rlbench}, Meta-World~\cite{meta-world}, and robosuite~\cite{robosuite} provide standardized task suites, ManiSkill2~\cite{maniskill2} adds wide object variation, and ManiSkill3~\cite{maniskill3} parallelizes simulation and rendering on the GPU. CALVIN~\cite{calvin} and LIBERO~\cite{libero} extend evaluation to long-horizon language-conditioned tasks and lifelong learning. Other environments focus on particular interactions. UniDoorManip~\cite{unidoormanip} simulates doors with diverse geometries and mechanisms. Several simulation platforms additionally support tactile sensing: Taxim~\cite{si2022taxim} renders vision-based tactile images, Tactile Gym 2.0~\cite{tactilegym2} compares optical tactile sensors through sim-to-real reinforcement learning, Robot Tactile Olympiad~\cite{roto2} benchmarks tactile reinforcement learning on four dexterous hands, while Manifeel \cite{luu2025manifeel}, UniVTac \cite{chen2026univtac}, and OpenViTac \cite{wu2026openvitac} offer standardized benchmarking platforms suitable for evaluating a wide range of visual-tactile policies. None of these environments, however, pairs its tasks with physical assets whose contact behaviors can be varied. Our simulation suite mirrors each printable mechanism in the benchmark and exposes that contact behavior as an explicit tunable parameter.

\section{ContactBench 1.0}
\label{sec: benchmark}

\begin{figure}
\textcolor{red}
    \centering
    \includegraphics[width=\columnwidth]{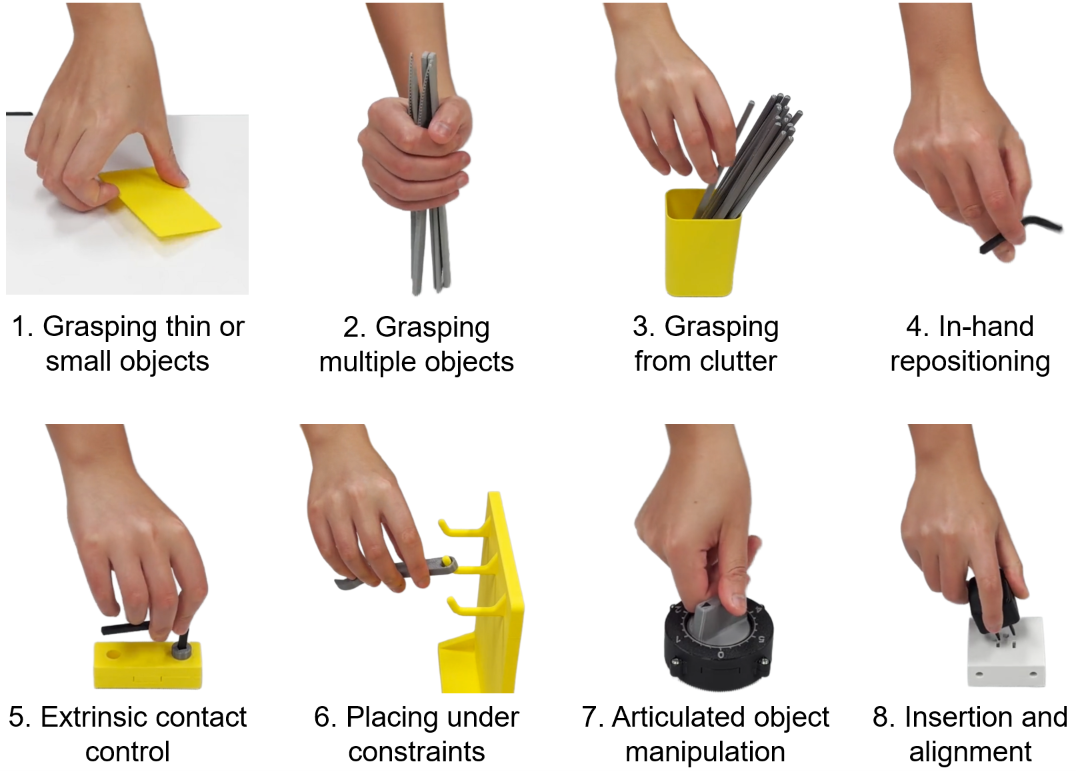}
    \caption{The eight contact-rich manipulation skills we identified for everyday contact-rich manipulation tasks. We built \BenchmarkName{} to instantiate these skills.
    }
    \label{fig:skills}
\end{figure}

Drawing on observations of everyday human manipulation, we identify eight key skills that underlie a broad range of contact-rich manipulation tasks. To enable reproducible evaluation of these skills across laboratories, we introduce \BenchmarkName{}. The benchmark comprises \NumTasks{} reference tasks organized around the eight skills and \NumAssets{} standardized, 3D-printable assets with corresponding simulation models. Shared task definitions, explicit success criteria, and reproducible assets provide a common basis for comparing robot systems. To evaluate generalization, we introduce controlled task and asset variations spanning both visually observable changes, such as object geometry, and changes that are not readily apparent from vision alone, such as contact forces. These variations enable evaluation of how manipulation systems respond to task changes that require adaptations in behaviors. The simulation counterparts of the assets and tasks will be introduced in Section ~\ref{sec:asset-simulation}.

\subsection{Contact-Rich Manipulation Skills}
\label{subsec:contact-rich-skills}



The eight skills we identified (shown in Figure \ref{fig:skills}) are drawn from observations of routine human activities and are chosen to span a broad range of contact-rich behaviors rather than to form a comprehensive or mutually exclusive taxonomy. Typically, humans use a combination of those skills to complete multi-stage tasks. 
We describe each skill in detail below.

\textbf{Grasping thin or small objects}. This skill involves acquiring a grasp on objects whose thin or small form factor demands more precision than objects of typical size and shape. Although pick-and-place has been studied extensively, the objects involved are often easy to grasp and require few considerations about grasp strategy. 
In contrast, many everyday objects, such as cards and keys, require not only precise grasp acquisition but also a carefully chosen grasp strategy, so that the resulting grasp supports the intended downstream manipulation. For example, a top-down grasp on a card’s thin edges is unstable and makes subsequent swiping or insertion actions difficult, whereas a grasp on its two faces is stable and well suited to these tasks. Achieving the latter often requires in-hand repositioning or a dedicated grasping strategy, such as sliding the card to the table edge to expose both faces, or using the thumb as a pivot while the index finger lifts the card off the surface.

\textbf{Grasping multiple objects}. This skill involves acquiring and holding several objects in a single grasp so that they can be transported together. It can substantially improve the efficiency of many everyday tasks: when clearing loose screws from a workspace, for example, collecting several in hand and depositing them into a bin together is much faster than picking and placing each one individually. Grasping multiple objects requires the end-effector to form and maintain an enclosure that keeps previously picked objects secure while acquiring new ones, which is a requirement that is difficult for parallel-jaw grippers and often benefits from high-dexterity end-effectors. Despite its practical value, multi-object grasping remains relatively unexplored in many general policy learning approaches, where methods often assume a single object per grasp \cite{zheng2026egoscale, luo2025being, dexwild}.

\textbf{Grasping from clutter}. This skill involves acquiring a grasp on a target object that is surrounded by and in contact with other objects. Objects in everyday environments are rarely presented in isolation: utensils sit mixed together in a drawer, screws and nuts of the same size share a storage bin, keys hang together on a key ring, cups tightly stack on top of each other to save storage space. Retrieving a target object in these settings requires locating it and then separating it from its neighbors, a process we refer to as singulation, without disturbing the scene more than necessary. The suitable manipulation strategy depends on how the clutter is arranged. In a loose clutter, such as a bin of screws, singulation often involves burrowing the gripper fingers into the pile to form a secure grasp on the target. In a tightly organized clutter, such as a deck of cards or a stack of cups, the target object must instead be extracted without displacing its neighbors.

\textbf{In-hand repositioning}. This skill involves changing how a held object is grasped, or its pose relative to the end-effector, without setting the object down. It is a routine part of human manipulation that bridges object acquisition and downstream usage, since an object is rarely picked up in the grasp configuration that the downstream task requires. Sometimes the grasp type itself needs to change \cite{grasp-taxonomy}: an Allen key might be most easily picked up with a pinch grasp, but using it to drive a screw usually requires a different grasp. Other times only the object pose needs to be adjusted: a key or USB stick picked up in the wrong orientation must be flipped before insertion. In-hand repositioning is most versatile with multi-fingered hands; parallel-jaw grippers are largely limited to sliding the object within the grasp, often relying on extrinsic contact with the environment \cite{touch-in-the-wild, 104-bronars2024texterity, 56-toskov2023hand}. While many prior works on dexterous in-hand manipulation focus on continuous rotation of blocky or spherical objects \cite{29-yuan2024robot, 54-yang2024anyrotate, 10-qi2023general, qi2023hand, chen2023visual}, the objects humans most often reposition, such as keys, cards, tools, and USB sticks, have more challenging form factors and remain difficult to effectively manipulate in-hand \cite{spin-pens}.

\begin{figure*}[t]
    \centering
    \includegraphics[width=\textwidth]{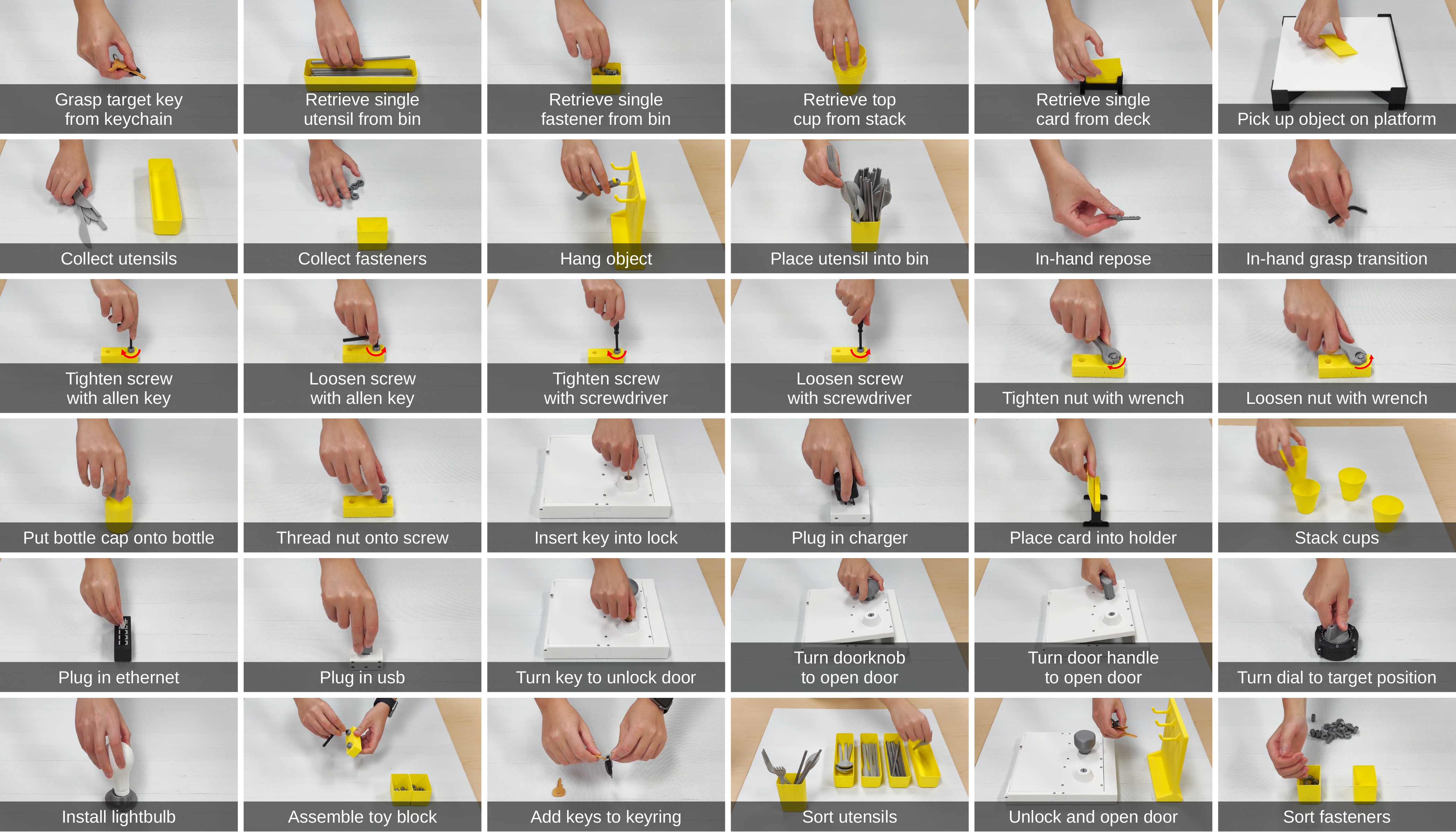}
    \caption{
    ContactBench 1.0 includes 36 manipulation tasks to concretely evaluate a robot system's proficiency on the eight skills. Many tasks require more than one skill to complete. 
    We record humans solving the tasks to illustrate successful task executions due to the difficulty of many tasks for current robot systems.
    }
    \label{fig:task_grid}
\end{figure*}

\textbf{Extrinsic contact control}. This skill involves regulating extrinsic contact, the contact made between an in-hand object and the environment, as opposed to intrinsic contact made between the gripper and the object it holds. It is especially prevalent in tool use: wiping a whiteboard with an eraser requires regulated pressure between the eraser and the board, and driving a screw with an Allen key requires maintaining the contact between the Allen key tip and the screw socket while transmitting torque. The defining challenge is that extrinsic contact is not a rigid joint. Without active regulation, it can slip, disengage, or transmit the wrong force. This distinguishes extrinsic contact control from articulated object manipulation, where the object’s joints maintain the constraint mechanically.

\textbf{Placing under environmental constraints}. This skill involves placing an object where the surrounding environment constrains how it must be positioned and released. Although pick-and-place has been studied extensively, placement is usually treated as dropping an object anywhere in a broad target region. Many everyday placements are more demanding: hanging requires precise actions to align the object with the hanging hook, stacking requires releasing the object without disturbing the stack beneath it, and placing into a cluttered region requires finding an unoccupied gap and depositing the object into it. In each case, the final alignment and release must be carefully controlled, since a premature or misaligned release fails the task even if the preceding approach was correct. We therefore treat constrained placement as a skill in its own right, rather than as the trivial final step of pick-and-place.

\textbf{Articulated object manipulation}. This skill involves operating objects whose rigid parts are connected by joints that constrain their relative motion. Such objects are abundant in human environments, including doors, drawers, buttons, switches, faucets, and levers. Therefore, operating articulated objects is a basic requirement for any robot working in human spaces. The defining challenge is complying with the kinematic constraint. A rigid position-controlled trajectory that wrongly estimates the kinematic constraint can rapidly build up damaging force on both the manipulator and the object. Therefore, effective manipulation of articulated objects must either estimate the constraint accurately or yield to it compliantly. 

\textbf{Insertion and alignment}. This skill involves bringing two parts into precise alignment and full engagement. Insertion has been widely studied as a canonical contact-rich problem with tasks such as peg-in-hole, electronics plugging, and gear meshing. It captures a distinct combination of challenges that involve tight tolerances, constrained insertion direction, and often substantial insertion force. Following these characteristics, we extended the category to include the initial alignment phase of threaded assembly, which involves similar tight tolerances for engaging the threads. Insertion can be further complicated by occlusion caused by the manipulated objects themselves: the pins of a charging block can be occluded by the charging block body, and the tip of a screw can be occluded by the nut being threaded onto it. The robot must therefore reason about geometry and contact it cannot directly observe visually.


\subsection{Benchmark Tasks}

To evaluate the contact-rich skills described above, we provide \NumTasks{} reference tasks with explicit task definitions and success criteria. As shown in Figure \ref{fig:task_grid}, the task set comprises 31 short-horizon tasks, each requiring one or two skills, and five long-horizon tasks that combine multiple skills across several stages. Since many of these tasks remain challenging for current robot systems, we record humans completing them to show what successful execution looks like. We intend these tasks to be attempted by any suitable robot embodiment. Together, these tasks expose open challenges for current robot systems and provide a common benchmark for measuring progress toward solving them. Table~\ref{tab:exp-tasks-with-specs} presents the six short-horizon tasks we used in our benchmarking experiments alongside their detailed specs, 
\ifArXivVersion
and the complete task list is shown in Appendix ~\ref{sec:full_task}.
\else
and the complete task list is available on the project website.
\fi

\subsection{Benchmark Task Variations}
\label{subsec:task-variations}

\begin{table}[!t]

\centering
\caption{Task variations supported by our benchmark. 
Blue rows denote \hlb{behavioral variations} that affect task behavior but are ambiguous to vision; green rows denote \hlvb{visual + behavioral variations} that affect both task behavior and appearance. Terminology follows Star-Gen \cite{gao2026taxonomy}.}
\label{tab:variation}
\small
\renewcommand{\arraystretch}{1.2}
\begin{tabular}{L{4.6cm} L{3.3cm}}
\toprule
\textbf{Task} & \textbf{Variations} \\

\midrule
\multirow{3}{=}{Turn doorknob to open door \\[0.2\baselineskip] Turn door handle to open door} & \hlb{Required rotation amount} \\
                                 & \hlb{Rotation force} \\
                                 & \hlvb{Object geometry} \\

\midrule
Plug in charger & \hlb{Insertion force} \\
Plug in USB & \hlvb{Object geometry} \\

\midrule
Put bottle cap onto bottle       & \multirow{2}{=}{\hlb{Required rotation amount} \\[0.2\baselineskip] \hlvb{Object geometry}} \\
Install lightbulb                & \\

\midrule
Tighten screw with Allen key   & \multirow{4}{=}{\hlb{Required rotation amount}} \\
Tighten screw with screwdriver & \\
Tighten nut with wrench        &  \\
Turn key to unlock door          & \\

\midrule
Grasp target key from keychain   & \multirow{6}{=}{\hlvb{Object geometry} \\[0.2\baselineskip] \hlvb{Number of objects}} \\
Retrieve single utensil from bin  & \\
Retrieve single fastener from bin & \\
Collect utensils & \\
Collect fasteners & \\
Place utensil into bin  & \\

\midrule
Pick up object from platform & \multirow{5}{=}{\hlvb{Object geometry}} \\
Hang object     & \\
In-hand repose              & \\
In-hand grasp transition              & \\
Turn dial to target position              & \\

\midrule
Retrieve top cup from stack   & \multirow{3}{=}{\hlvb{Number of objects}} \\
Retrieve single card from deck   & \\
Place card into holder  & \\
Stack cups  & \\

\bottomrule
\end{tabular}
\end{table}

We introduce task variations designed to evaluate behavioral generalization \cite{gao2026taxonomy}, which is the ability of a policy to adapt to novel task-relevant conditions that are ambiguous to vision. While recent work on generalist manipulation policies has demonstrated impressive zero-shot transfer to novel scenes, objects, and language instructions, behavioral generalization has not received as much attention, despite being especially prevalent in contact-rich manipulation. Humans can unlock a door regardless of whether the doorknob must be turned $45^{\circ}$ or $60^{\circ}$, plug in a charger regardless of whether the outlet is tight or loose, and tighten a bottle cap regardless of whether it takes two or three revolutions. Each of these variations is difficult to perceive visually but is central to successful task execution. Following the generalization taxonomy proposed in Star-Gen \cite{gao2026taxonomy}, the variations we consider fall into two categories: behavioral variations, which affect the behavior required to complete a task without altering its visual appearance, and visual + behavioral variations, which affect both the task behavior and visual appearance. 
Several standard variations, such as changes to object color, lighting, camera viewpoint, and initial object pose, are supported by our benchmark without dedicated asset designs, and we do not enumerate them here. 

Table \ref{tab:variation} lists supported benchmark task variations, including but not limited to required interaction forces, required rotation amounts, and object geometry. Figure \ref{fig:asset-exp}(c) shows examples of the various levels of interaction forces available as task variants, which are achieved by deliberate asset designs described in Section \ref{subsec:assets}. Many reference tasks support at least one variation, and several support both behavioral and visual + behavioral variations, allowing the same task to be evaluated under either or both axes.

\subsection{Benchmark Assets}
\label{subsec:assets}

ContactBench contains \NumAssets{} assets that support the evaluation of the manipulation skills and generalization capabilities discussed in Section \ref{subsec:contact-rich-skills} and Section \ref{subsec:task-variations}. Compared to benchmarks that standardize the assets through specific vendors and distributors \cite{calli2015ycb, nist-taskboard, chen2026manipulationnet}, we prioritize the accessibility of our assets to researchers worldwide. All assets are therefore 3D-printable in standard polylactic acid (PLA) filament, which is inexpensive and easy to print, and are assembled with standard button-head M3 screws and nuts. Asset files, print settings, and assembly instructions will be made available on the project website.

The accessibility of 3D printing comes with two challenges: how realistically the printed parts reproduce the contact forces of real objects, and how consistently they do so across different 3D printers. Existing 3D-printable benchmarks for contact-rich manipulation often use large clearances between mating parts, resulting in loose fits that generate little contact force \cite{furniturebench, fmb, fabrica}. This contradicts the demands of contact-rich manipulation, where applying and regulating contact forces is central to task execution. Tightening the clearance between parts can increase frictional contact forces, but printing accuracy varies across printers, so the forces obtained with a given clearance on one printer might not reliably transfer to another.

\begin{figure}
    \centering
    \includegraphics[width=0.95\columnwidth]{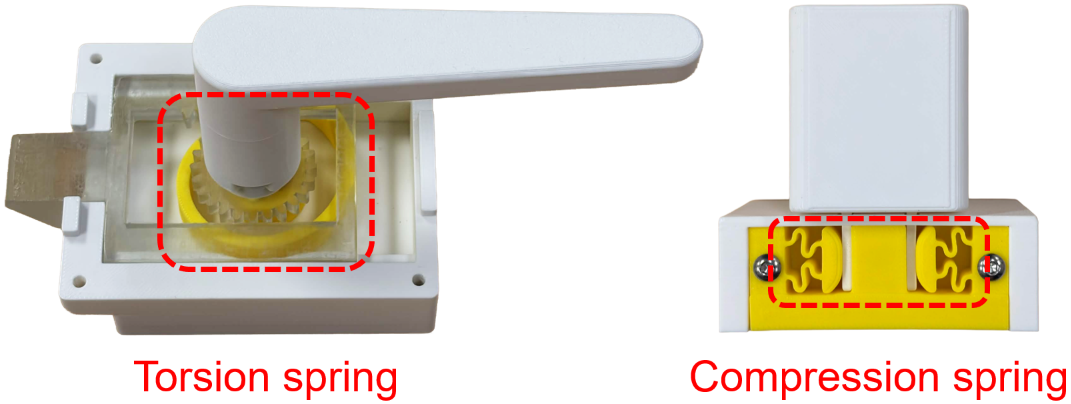}
    \caption{
    We design spring structures inside the assets to approximate their resistance force during manipulation. Here is an example of the torsional spring inside a door handle and a compression spring inside an outlet.   
    }
    \label{fig:compliant-mechanisms}
\end{figure}

We address both challenges by reducing the dependency of force generation on printing accuracy. Rather than relying on a tuned fit between parts, our assets exploit the PLA material compliance and generate interaction forces with compliant spring designs, such as the torsion spring in the door handle and the compression spring in the outlet shown in Figure \ref{fig:compliant-mechanisms}. Since a spring’s behavior is governed by its geometry and the material properties of PLA, the resulting forces depend on quantities that are more consistent across printers than part tolerances. Varying the spring geometry also lets us provide task variants that require different forces to complete, as described in Section \ref{subsec:task-variations}.

\begin{figure}
    \centering
    \includegraphics[width=\columnwidth]{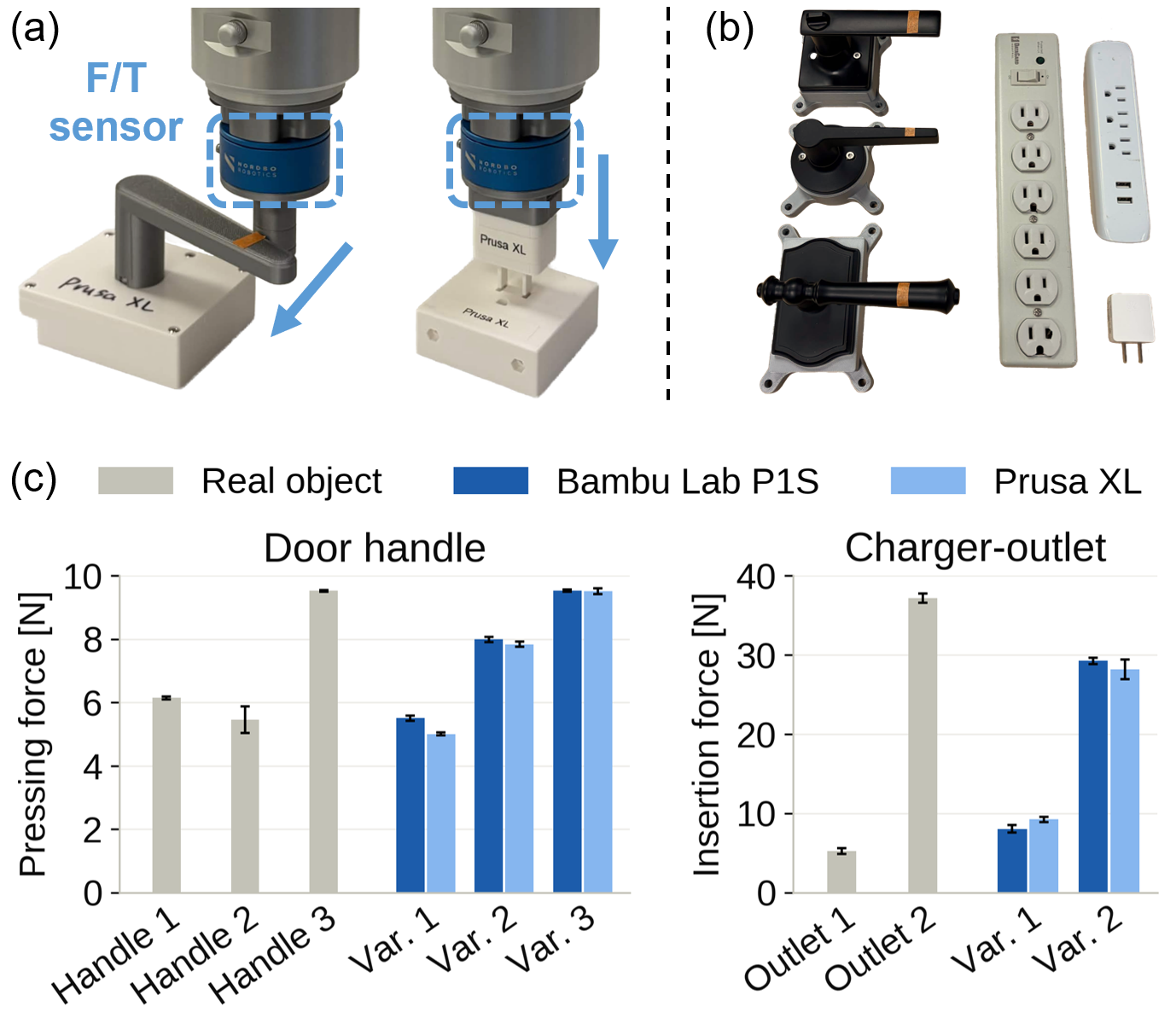}
    \caption{The resistance forces exhibited by our assets are comparable to those of real objects and are generally consistent across different 3D printers. We show experimental results over a door handle and a charger-outlet pair with their variants. (a) shows the experimental setup, (b) shows the real products used for comparisons, and (c) shows the measured max pressing and insertion forces across the real objects and the asset variants printed by two printers.}
    \label{fig:asset-exp}
\end{figure}

To validate our approach, we mount a force/torque (F/T) sensor on the robot end-effector and let the robot execute preprogrammed motions that press a door handle and insert a charger into an outlet, as illustrated in Figure \ref{fig:asset-exp}(a). 
As references, we first measure the forces produced by the three real door handles and two real outlets shown in Figure \ref{fig:asset-exp}(b). We then print three variants of our door handle and two variants of our charger-outlet assets on a Bambu Lab P1S and a Prusa XL and repeat the measurements. Both printers use the same filament and print settings. The force measurement is repeated ten times on each real and printed asset.  As shown in Figure \ref{fig:asset-exp}(c), the printed door handles produce 5.01N – 9.54N pressing forces, matching closely with the 5.45N – 9.53N produced by the real handles. The printed outlets produce 8.04N – 29.26N insertion forces, which fall within the 5.24N – 37.18 N range spanned by the real outlets. Across all tested variants, the forces generated by assets printed on the two printers differ by at most 0.5N for the door handle and 1.19N for the charger-outlet pair. These results indicate that our asset designs with compliant PLA springs produce interaction forces comparable to real objects and can achieve this reproducibly across printers.

\begin{figure*}[!t]
    \centering
    \includegraphics[width=\textwidth]{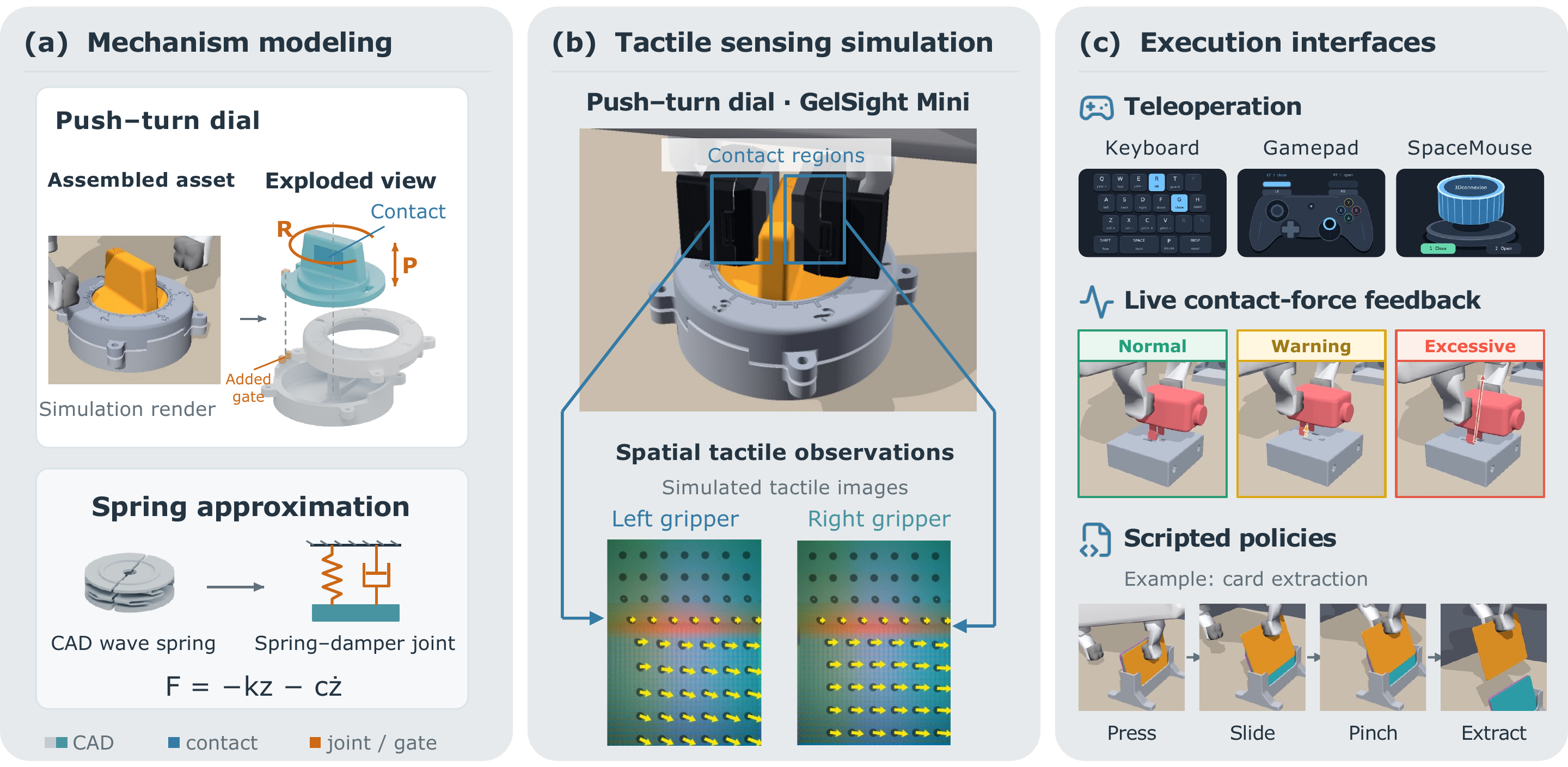}
    \caption{
        {Overview of the \BenchmarkName{} simulation suite.}
        The suite provides assets that match the 3D-printable versions in terms of shape and mechanisms
        \textbf{(a)}, a simulated GelSight Mini interface
        \textbf{(b)}, and a teleoperation interface for demonstrating manipulation policies through 
        keyboard, gamepad, or SpaceMouse \textbf{(c)}. We also provide a set of scripted policies for the contact-rich manipulation tasks.
    }
    \label{fig:simulation_overview}
\end{figure*}
\section{Simulation Benchmark}
\label{sec:asset-simulation}


\begin{figure*}[!t]
    \centering
    \includegraphics[width=\textwidth]{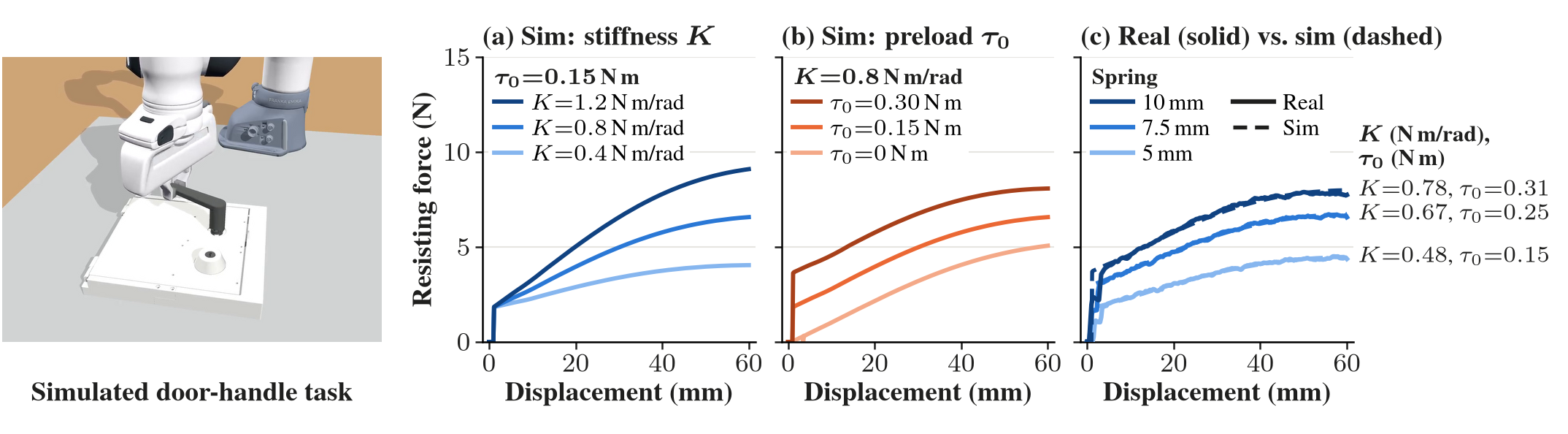}
    \caption{
    Our simulated asset models feature tunable force response. In this example of the door handle asset, we make a robot rotate it at a linear speed of 10\,mm/s. Here we show the measured force/torque value at the robot wrist.  
        \textbf{(a)}~Force response when changing the simulated door handle's parameter of revolute spring with stiffness $K$.
        \textbf{(b)}~Force response when changing the simulated door handle's parameter of pre-load torque $\tau_0$.
        \textbf{(c)}~Force response of simulated assets with varied settings in comparison with the response from real assets. The real ones are printed on a Bambu Lab P1S with 5, 7.5, and 10\,mm
        springs (solid, mean of 10 trials). The simulation reproduces the
        shape and ordering of the measured profiles.
    }
    \label{fig:sim-real-door-handle}
\end{figure*}

\begin{figure*}[!t]
    \centering
    \includegraphics[width=\textwidth]{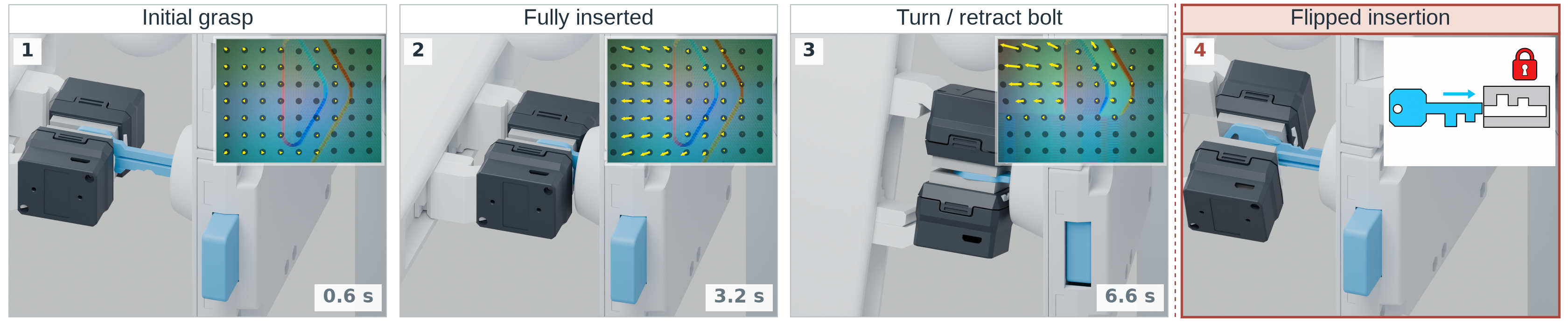}
    \caption{
        {Example of the scripted policy in simulation with tactile feedback: key insertion and turning.}
        The correctly oriented key is inserted and turned to retract the bolt (1–3).
        The keyway interlock rejects a flipped-key insertion attempt (4).
        Insets show simulated tactile images from the right fingertip;
        yellow arrows indicate marker displacement.
    }
    \label{fig:sim-key}
\end{figure*}

We develop the corresponding assets and task setups for \BenchmarkName{} in simulation, building on the Newton physics engine \cite{newton-physics} with the MuJoCo Warp solver \cite{todorov2012mujoco, mujocowarp2025}. The simulated assets reproduce the behavior of the real assets with an emphasis on force-realistic contact interactions: spring resistance, frictional retention, constrained motion, geometric interlocking, and changes in load during engagement and release. The simulation suite supports demonstration collection through scripted
policies and teleoperation, and includes an optional module for simulating a fingertip GelSight tactile sensor. Figure~\ref{fig:simulation_overview} provides an overview of the simulation suite. 

\subsection{Assets and contact modeling}
We export the source CAD of the assets as separate rigid parts and assemble mechanisms using their joint axes, travel limits, and fixed supports. Printed spring bodies are replaced by prismatic or revolute spring--damper drives, exposing stiffness and preload as explicit parameters. This retains the mechanism's functional resistance without simulating the printed spring's full deformation. 
Figure~\ref{fig:sim-real-door-handle} shows the force response of one asset, the door handle, with its printed spring modeled as a revolute spring with stiffness $K$ and preload $\tau_0$. Tuning these two parameters reproduces the measured profiles of handles printed with three different springs.

To balance the contact fidelity and computational efficiency, we adopt different strategies to model each asset. 
Contact-critical geometry, including threads and insertion passages, uses mesh signed-distance fields (SDFs); simpler collision primitives are used where detailed geometry is unnecessary. Thin features require sufficient SDF resolution and surface sampling. For example, the card-grasping implementation uses a 512-cell maximum SDF resolution for its 1.2\,mm-thick card. Sub-resolution features that require geometric approximations are documented per task.
Appendix~\ref{app-sec:sim-assets} shows how these modeling choices apply to representative mechanisms.

\begin{figure}[!t]
    \centering
    \includegraphics[width=\columnwidth]{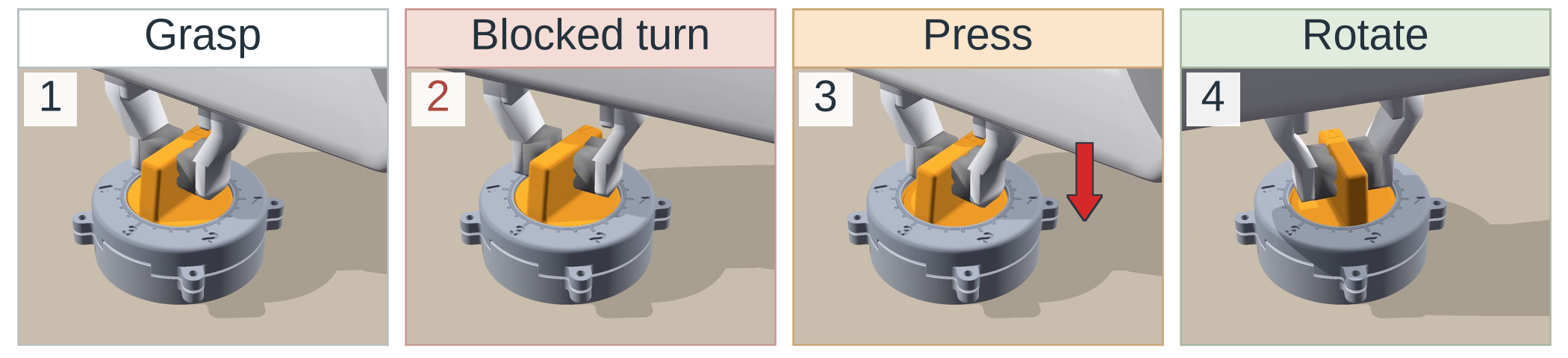}\\[2pt]
    \includegraphics[width=\columnwidth]{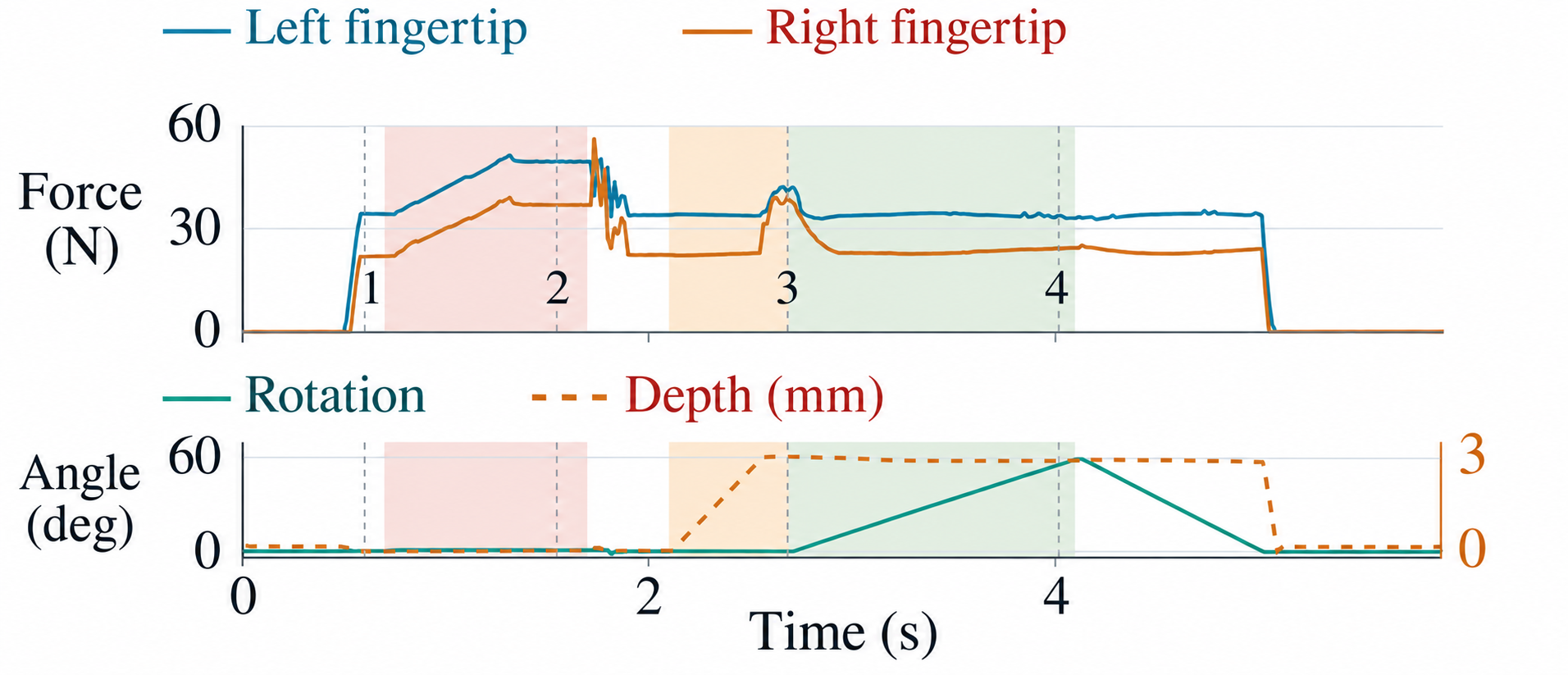}
    \caption{
        {Example of the scripted policy in simulation: push-and-turn dial operation.}
        The plots show fingertip forces, dial rotation, and axial depression during the process,
        with numbered guides matching the keyframes.
        The red, orange, and green bands denote the phase of blocked turning, pressing,
        and permitted rotation, respectively.
    }
    \label{fig:sim-dial}
\end{figure}

\subsection{Tactile sensing simulation}
To support potential tactile-augmented policy learning, we simulate GelSight Mini sensors mounted on the gripper fingers, including their compliant sensing pads and rigid housings. Contact is modeled using a hydroelastic pressure-field framework~\cite{elandt2019pressure}. The contact surface is rasterized into spatial pressure maps, with pressure scaling based on the normal load resolved by the dynamics solver. Sensor indentation is estimated through a nonlinear pressure--indentation relation that captures increased resistance under compression. In the observation model, relative tangential motion is accumulated into a shear displacement field, bounded by local pressure and friction.

The resulting deformation fields are rendered into tactile images using optical shading based on Taxim~\cite{si2022taxim}. Local surface gradients and image position determine color intensity, while marker displacement reflects shear deformation and changes in surface height. The module provides tactile images, pressure and indentation maps, shear displacement, and net finger forces. These outputs describe the spatial structure of contact and its evolution during manipulation.

\subsection{Teleoperation and data collection}
We provide teleoperation interfaces for keyboard, gamepad, and SpaceMouse to support interacting with the simulated assets and task setups. All three devices control end-effector translation, rotation, and gripper opening through a shared command interface, with adjustable speed and a fine-control mode for precise alignment. Live contact-force feedback allows operators to monitor loading and adjust their commands during execution (Figure~\ref{fig:simulation_overview}(c)). 
Collected trajectories align executed actions with observations and contact measurements, while retaining task outcomes and simulation configurations for potential policy training and evaluation.

\subsection{Scripted policies and task execution}
We provide task-specific scripts to demonstrate the intended interaction sequences in simulation and generate reference trajectories. Each script specifies a predefined sequence of motion phases, such as approach, grasp, engagement, manipulation, and release.  Depending on the task, execution follows a timed trajectory or updates motion targets using simulated object states and contact-force measurements. The successful rollouts can be recorded to generate seed demonstrations for potential subsequent policy learning.

Figure~\ref{fig:sim-key} and Figure~\ref{fig:sim-dial} show two examples: key-operated unlocking with simulated fingertip tactile images, and a push-and-turn dial that rotates only after it is pressed down. 
\section{Experiments}
\label{sec: experiments}

\begin{table*}[!t]
  \caption{Benchmark tasks used for policy learning experiments}
  \label{tab:exp-tasks-with-specs}
  \centering
  \renewcommand{\arraystretch}{1.2}
  \setlength{\aboverulesep}{4pt}
  \setlength{\belowrulesep}{4pt}
  \renewcommand{\tabularxcolumn}[1]{m{#1}}  

  \begin{tabularx}{\textwidth}{@{} >{\centering\arraybackslash}m{0.39\textwidth} T @{}}
    \toprule
    \textbf{Task Execution Sequence Using UMI} & \multicolumn{1}{c@{}}{\textbf{Task Specs}} \\


    \midrule
    \includegraphics[width=\linewidth]{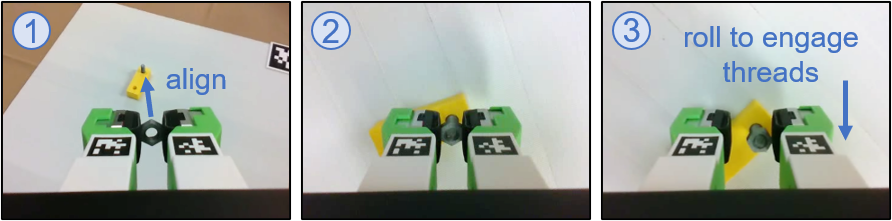}
    &
    \textbf{Task:} Thread nut onto screw

    \textbf{Skills:} Insertion and alignment, articulated object manipulation

    \textbf{Description:} Screw the nut held in hand onto the screw head placed in the workspace such that their threads fully engage.

    \textbf{Success criteria:} The nut and screw threads engage. The robot movement stops after terminating the rotation motion, moving away from, and breaking contact with task objects.
    \\


    \midrule
    \includegraphics[width=\linewidth]{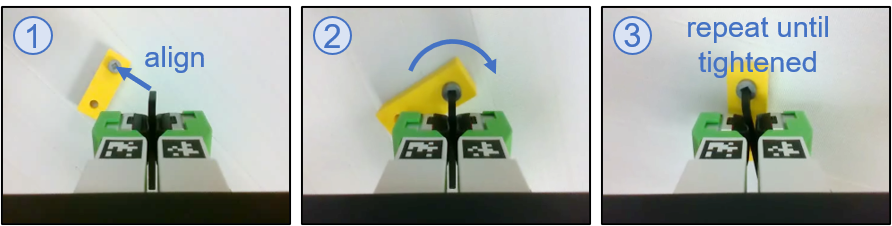}
    &
    \textbf{Task:} Tighten screw with Allen key

    \textbf{Skills:} Extrinsic contact control, insertion and alignment

    \textbf{Description:} A screw is placed into the screw slot of an assembly block in the workspace, but the threads are unengaged. Use the Allen key grasped in hand to tighten the screw.

    \textbf{Success criteria:} The screw is tightened and remains stationary when a small force is applied in an attempt to rotate the screw. The robot movement stops after terminating the tightening motion, moving away from, and breaking contact with the screw.
    \\


    \midrule
    \includegraphics[width=\linewidth]{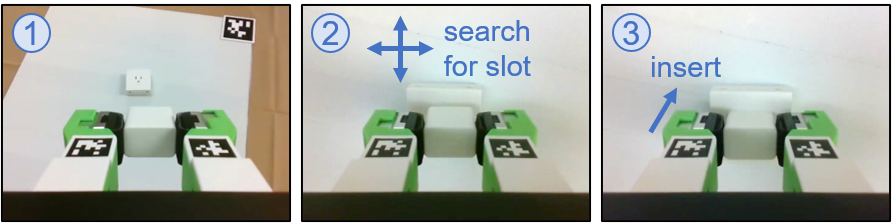}
    &
    \textbf{Task:} Plug in charger

    \textbf{Skills:} Insertion and alignment

    \textbf{Description:} An outlet is placed in the workspace. Fully insert the charger held in hand into the outlet.

    \textbf{Success criteria:} The charger is fully inserted into the outlet, such that it does not noticeably move when pushed further toward the outlet. The robot movement stops after terminating the insertion motion, moving away from, and breaking contact with task objects.
    \\


    \midrule
    \includegraphics[width=\linewidth]{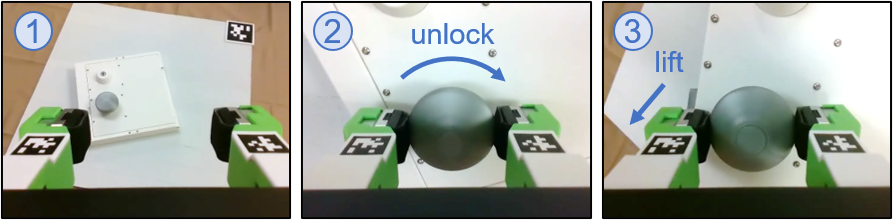}
    &
    \textbf{Task:} Turn doorknob to open door

    \textbf{Skills:} Articulated object manipulation

    \textbf{Description:} Turn the doorknob fully to unlock the door, then open the door.

    \textbf{Success criteria:} The door is opened by at least 5 degrees, such that it is visually obvious that the door has been unlocked.
    \\


    \midrule
    \includegraphics[width=\linewidth]{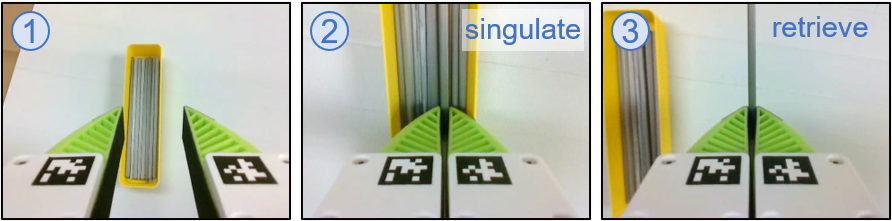}
    &
    \textbf{Task:} Retrieve single utensil from bin

    \textbf{Skills:} Grasping from clutter, grasping thin or small objects

    \textbf{Description:} A utensil storage bin is placed in the workspace. The target utensil is placed in the bin alongside multiple other utensils. Pick up a single copy of the target utensil from the cluttered storage bin and place it on the table without removing other objects from the bin.

    \textbf{Success criteria:} The robot movement stops after separating a single copy of the target utensil and placing it on the table. Other utensils should remain in the storage bin. The robot should not remain in contact with the task objects.
    \\


    \midrule
    \includegraphics[width=\linewidth]{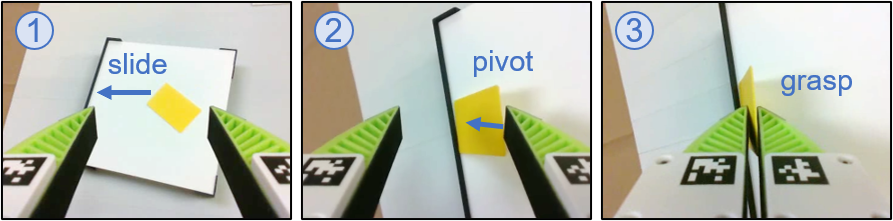}
    &
    \textbf{Task:} Pick up card from platform

    \textbf{Skills:} Grasping thin or small objects

    \textbf{Description:} Pick up the thin and/or small object placed on the flat platform.

    \textbf{Success criteria:} The target object is stably grasped by the robot and lifted off the platform.
    \\

    \bottomrule
  \end{tabularx}
\end{table*}

\BenchmarkName{} is designed to evaluate different manipulation systems, where the performance provides insight into the system's deficiencies in addressing contact-rich manipulation. In this section, we demonstrate the use of the real-world benchmark sets with four recent policy learning methods on six tasks, and analyze their behaviors. We expect the results to provide insight for the following questions: 
\begin{enumerate}
    \item What are the major challenges for vision-only policies on contact-rich tasks? 
    \item Do representative visual-tactile policies address these challenges?
    \item How do the evaluated policies generalize when task-relevant conditions change in ways that are difficult to perceive visually?
\end{enumerate}

\begin{figure}
    \centering
    \includegraphics[width=\columnwidth]{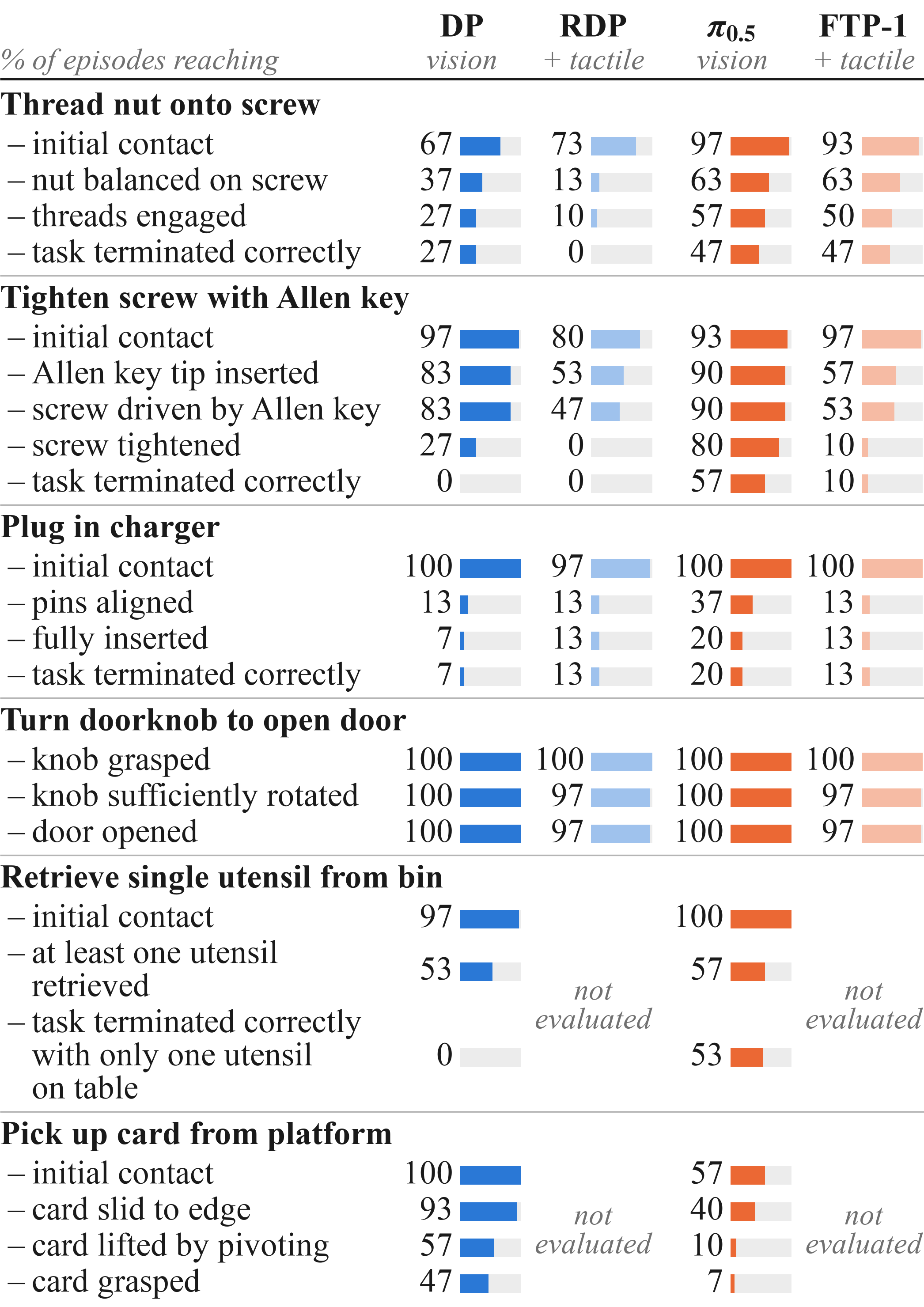}
    \caption{Task progress achieved by each policy, averaged over 30 evaluation trials.}
    \label{fig:checkpoint-prog}
\end{figure}

The policies we evaluate include two vision-only policy learning methods, Diffusion Policy (DP) \cite{chi2025diffusion} and $\pi_{0.5}$ \cite{pi05}, and their representative tactile-augmented variants, Reactive Diffusion Policy (RDP) \cite{xue2025reactive} and Foundation Tactile Policy (FTP-1) \cite{ftp1}. 
The performance of each policy across the selected tasks is shown in Table \ref{tab:results} and Figure \ref{fig:checkpoint-prog}. While the policies were able to make progress toward solving each task, the overall success rate is 56.7\% for the best policy-task combination, with the turn doorknob to open door task being the exception where all policies achieved near 100\% success rate. We found that perceiving and reacting to contact events remains challenging for vision-only policies in the form of establishing and maintaining contact, precise alignment, avoiding unsafe forceful interactions, and recognizing task-critical states. These challenges contribute to policies' inability to complete the tasks, taking noticeably longer and requiring more attempts to complete certain tasks compared to the human demonstrator, and generating excessive forces between the end-effector and the environment. 
The two representative visual-tactile policy learning methods we evaluated do not systematically address these challenges faced by vision-only policies, and oftentimes performed worse. On the other hand, in the generalization experiments, we found that vision-only policies showed fragility under behavioral perturbations, while visual-tactile policies adapted better. 

\newcommand{\sr}[1]{\ifnum#1<10 \phantom{0}\fi#1/30}  
\newcommand{\best}[1]{\textbf{#1}}           
\newcommand{\x}[1]{#1$\times$}          
\newcommand{\xfew}[1]{\x{#1}$^\dagger$}  
\newcommand{\nosucc}{--}              
\newcommand{\noteval}{\multicolumn{4}{c}{\textit{not evaluated}}}
\newcommand{\subhead}{%
  \makecell{SR\\$\uparrow$} &
  \makecell{Prog\\$\uparrow$} &
  \makecell{$>$40\,N\\$\downarrow$} &
  \makecell{Time\\$\downarrow$}}
 
\begin{table*}[t]
\centering
\caption{Policy performance on the six selected tasks (30 episodes per method per task)}
\label{tab:results}
\footnotesize
\setlength{\tabcolsep}{3pt}
\begin{tabular*}{\textwidth}{@{\extracolsep{\fill}} l c cccc cccc cccc cccc @{}}
\toprule
& Human
& \multicolumn{4}{c}{DP}
& \multicolumn{4}{c}{RDP}
& \multicolumn{4}{c}{$\pi_{0.5}$}
& \multicolumn{4}{c}{FTP-1} \\
\cmidrule(lr){2-2}\cmidrule(lr){3-6}\cmidrule(lr){7-10}\cmidrule(lr){11-14}\cmidrule(l){15-18}
Task & Time & \subhead & \subhead & \subhead & \subhead \\
\midrule
\makecell[l]{Thread nut\\onto screw}          & 17.5\,s & \sr{8}  & 0.39 & 33\% & \x{1.72}     & \sr{0}  & 0.24 & 27\% & \nosucc      & \best{\sr{14}} & \best{0.66} & 27\% & \x{1.22}     & \best{\sr{14}} & 0.63 & 83\% & \x{1.32} \\ \addlinespace
\makecell[l]{Tighten screw\\with Allen key}  & 19.7\,s & \sr{0}  & 0.58 & 10\% & \nosucc      & \sr{0}  & 0.36 & 0\%  & \nosucc      & \best{\sr{17}} & \best{0.82} & 7\%  & \x{1.50}     & \sr{3}  & 0.45 & 3\%  & \xfew{0.84} \\ \addlinespace
\makecell[l]{Plug in\\charger}               & 7.3\,s  & \sr{2}  & 0.32 & 80\% & \xfew{1.06}  & \sr{4}  & 0.34 & 83\% & \xfew{1.15}  & \best{\sr{6}} & \best{0.44} & 40\% & \x{0.89}     & \sr{4}  & 0.35 & 40\% & \xfew{0.95} \\ \addlinespace
\makecell[l]{Turn doorknob\\to open door}    & 7.6\,s  & \best{\sr{30}} & \best{1.00} & 10\% & \x{0.92}     & \sr{29} & 0.98 & 63\% & \x{0.91}     & \best{\sr{30}} & \best{1.00} & 7\%  & \x{0.87}     & \sr{29} & 0.98 & 43\% & \x{0.89} \\ \addlinespace
\makecell[l]{Retrieve single\\utensil from bin}  & 11.8\,s & \sr{0}  & 0.50 & 97\% & \nosucc      & \noteval                                  & \best{\sr{16}} & \best{0.70} & 47\% & \x{2.05}     & \noteval \\ \addlinespace
\makecell[l]{Pick up\\thin card}             & 7.5\,s  & \best{\sr{14}} & \best{0.74} & 57\% & \x{1.11}     & \noteval                                  & \sr{2}  & 0.28 & 60\% & \xfew{0.93}  & \noteval \\
\bottomrule
\end{tabular*}

\vspace{3pt}
{\raggedright\footnotesize
SR: number of successful episodes out of 30. Prog: mean fraction of task checkpoints reached per episode (0--1).
$>$40\,N: percentage of episodes with peak wrist force exceeding 40\,N.
Time: mean completion time of successful episodes; for the policies, shown as a multiple of the time taken by human demonstrations.
$^\dagger$Fewer than 5 successful episodes. A dash (--) marks no successful episodes.\par}
\end{table*}

\subsection{Experiment Setup}

\textbf{Policies}. We benchmark two vision-only policy learning methods together with their tactile-augmented variants. Diffusion Policy (DP) \cite{chi2025diffusion} models robot action sequence generation as a conditional denoising diffusion process, and has become one of the most widely adopted policy learning frameworks in manipulation research, serving as the backbone and comparison point for a large body of subsequent work. $\pi_{0.5}$ \cite{pi05} is a vision-language-action model that combines flow-matching action generation with vision-language pretraining, and is representative of the recent wave of large-scale generalist manipulation policies. Their tactile-augmented variants, Reactive Diffusion Policy (RDP) \cite{xue2025reactive} and Foundation Tactile Policy (FTP-1) \cite{ftp1}, extend the architecture to incorporate tactile feedback. Reactive Diffusion Policy uses a slow-fast architecture with a tactile-conditioned reactive policy on top of Diffusion Policy, and FTP-1 is a generalist tactile policy built on $\pi_{0.5}$ and pretrained on data from diverse tactile sensors and robot embodiments. 

We collect training data with a customized Universal Manipulation Interface (UMI) \cite{umi} hand-held gripper. A single wrist camera view serves as the RGB camera input for all policies, while the marker displacements and tactile images from the GelSight Mini sensors are used as tactile inputs for RDP and FTP-1 respectively. We use the official implementations for all policies and train them to predict future actions relative to the current robot pose.
We focus on evaluating a policy’s ability to learn fine-grained, contact-rich manipulation skills. Therefore, we finetune $\pi_{0.5}$ and FTP-1 on each selected task and effectively treat them as single-task specialist policies, leaving evaluation of their reasoning and language understanding capabilities in the contact-rich setting to future work. 
Data collection and policy training details and hyperparameters can be found in Appendix \ref{sec:policy-training-details}.

\textbf{Tasks}. 
We select six tasks from the benchmark for policy evaluation: \textit{thread nut onto screw}, \textit{tighten screw with Allen key}, \textit{plug in charger}, \textit{turn doorknob to open door}, \textit{retrieve single utensil from bin}, and \textit{pick up card from platform}. Table \ref {tab:exp-tasks-with-specs} shows the detailed specs for each task and their typical UMI demonstration sequence. Together, these tasks cover five of the six skills introduced in Section \ref{subsec:contact-rich-skills} that a single robot arm with a parallel-jaw gripper can complete. We selected them because we believe they are among the most challenging tasks in the benchmark that are feasible for our robot setup.

\textbf{Robot hardware setup}. All experiments use a UR5e robot equipped with a WSG50 parallel-jaw gripper and a RealSense D435i wrist camera. For \textit{thread nut onto screw}, \textit{tighten screw with Allen key}, \textit{plug in charger}, and \textit{turn doorknob to open door}, we use custom 3D-printed fingers equipped with GelSight Mini tactile sensors. We cover the sensing area of each GelSight Mini with a layer of 3M gripper material to reduce gel pad wear and tear. For \textit{retrieve single utensil from bin} and \textit{pick up card from platform}, we use compliant Fin-Ray fingers instead, since these tasks frequently generate contact that is outside of GelSight Mini’s fingertip-only sensing coverage and GelSight Mini's bulky form factor is not well suited to these tasks. Consequently, tactile signals are not recorded during demonstrations for these two tasks, and only the vision-only policies are evaluated on them. 

\textbf{Demonstration collection}. All demonstrations were collected with a customized UMI hand-held gripper. To achieve the high trajectory tracking accuracy required for contact-rich manipulation, we replace the wide-angle fisheye camera with a RealSense D435i camera, and the SLAM-based localization with HTC VIVE-based motion tracking using OpenVR \cite{valvesoftware_openvr}.
We calibrate the HTC VIVE tracking system to collect demonstration trajectories in the robot base frame. 
We prevent the human operator from relying on depth perception or third-person visual cues that the policies would not have access to during rollouts by requiring them to complete the tasks based on the wrist camera stream and their own tactile feedback. In principle, this matches the observations available to tactile-augmented policies at rollout time.
We note, however, that human tactile perception remains richer than what the GelSight Mini tactile sensors can capture. While GelSight Mini provides high-resolution contact information only within the fingertip contact patch, the human hand holding the UMI gripper can perceive contact happening across the entire gripper body.

\begin{figure}
    \centering
    \includegraphics[width=\columnwidth]{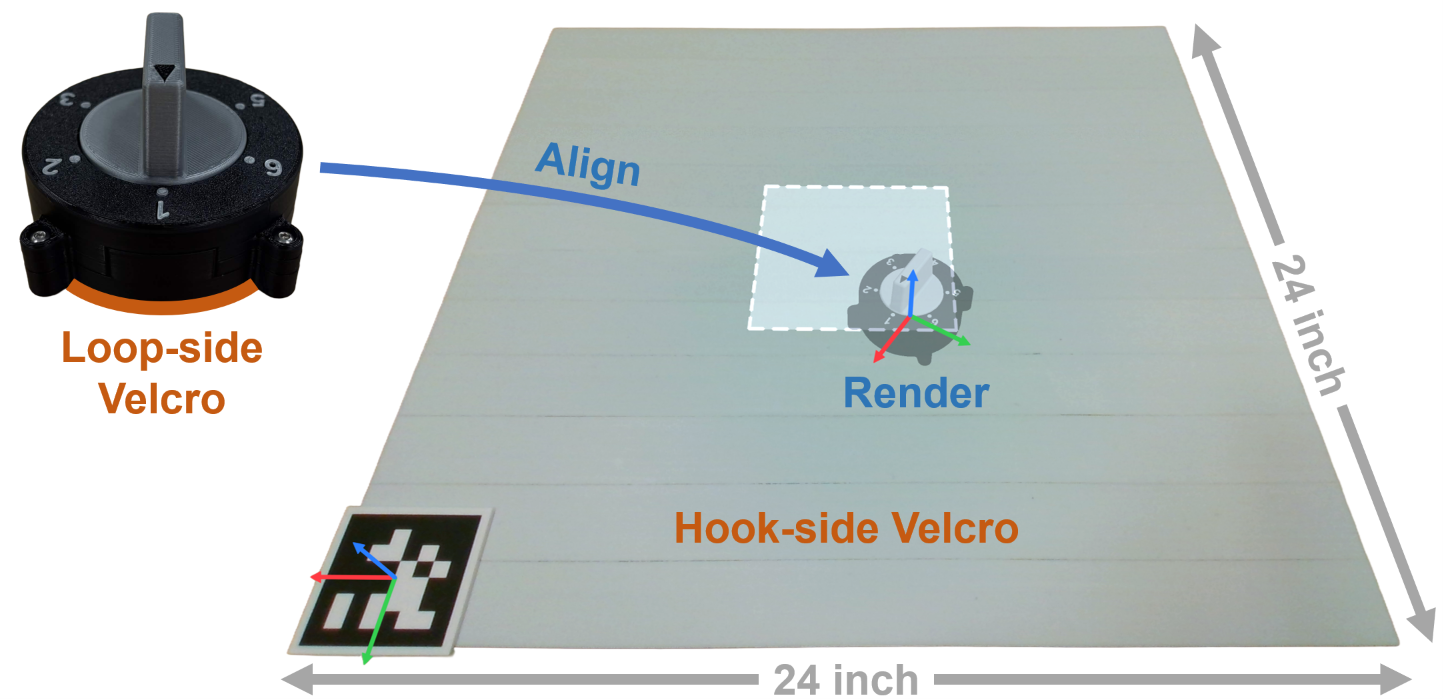}
    \caption{We systematically sample task object initial positions within a specified region and render them in a 3rd-person camera view to guide the human operator in task object placement. When necessary, the task objects are attached to the workspace with Velcro.}
    \label{fig:real-world-task-setup}
\end{figure}

\textbf{Evaluation protocol}. To generate task initial conditions systematically and make them easy to reproduce across research labs, we sample each object's initial pose from a fixed random seed and let the human operator place the object at that location. 
To hold objects at their sampled poses, we use a 24 inch by 24 inch acrylic board covered with hook-side Velcro. We apply loop-side Velcro to task-relevant objects (assembly block with screws, outlet, door, storage bin with utensils, platform) so they can be fixed at any location on the board. This method of fixing objects in the workspace also lets a single robot arm perform tasks that would otherwise require a second arm to stabilize the object. We sample the pose of each object's origin from a uniform distribution within a 0.15m by 0.15m area centered on the board. To provide visual guidance to the operator placing the objects, we render objects at their sampled poses with PyBullet and overlay them onto the board in a 3rd-person camera view. This procedure, shown in Figure \ref{fig:real-world-task-setup} and similar to the setups in \cite{TRI_LBM}, \cite{chen2026manipulationnet}, and \cite{SceneReplica}, is used for both demonstration collection and rollouts. For each task, we collect 100 demonstrations and evaluate 30 rollouts per policy, with robot initial poses during evaluation sampled from a Gaussian distribution fit to the UMI gripper initial poses in the demonstrations; the same 30 object and robot initial poses are used across all four policies for fair comparison. Each rollout has a one-minute time limit and is evaluated against the task's success criteria. For the six tasks we use for experiments, we additionally measure task progress as the fraction of task-specific checkpoints completed.

\subsection{What Are the Major Challenges for Vision-Only Policies on Contact-Rich Tasks?}
\label{subsec:vision-policy-challenges}

\begin{figure*}[!t]
    \centering
    \includegraphics[width=\textwidth]{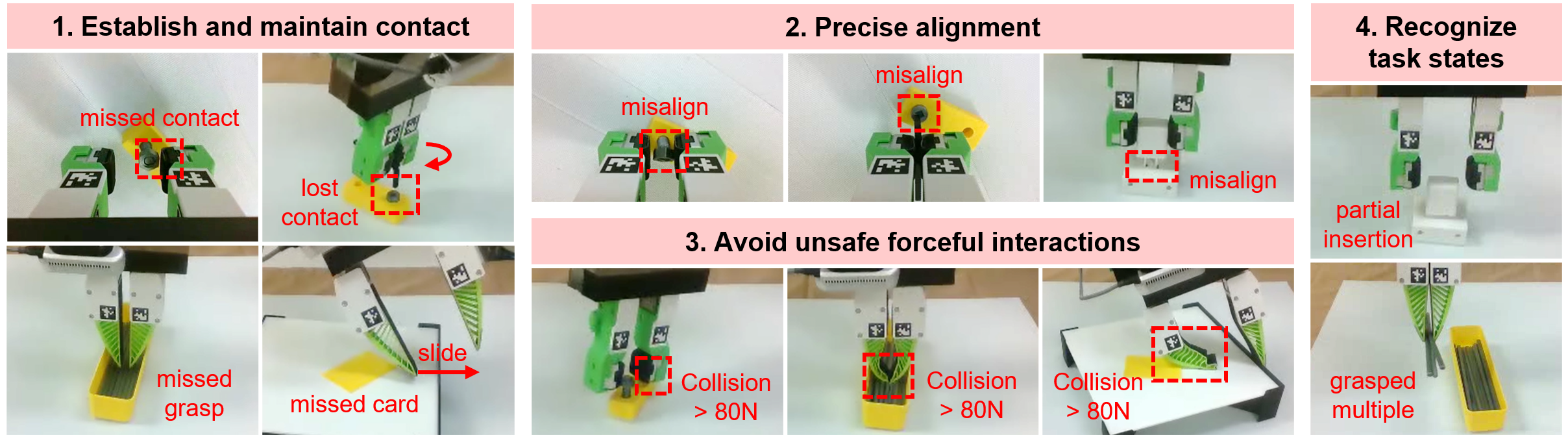}
    \caption{Examples of the four typical challenges vision-only policies encountered on the selected tasks.    
    }
    \label{fig:vision-failure-patterns}
\end{figure*}

We discuss the performance of the vision-only policies, Diffusion Policy (DP) and $\pi_{0.5}$, in this subsection. Excluding \textit{turn doorknob to open door}, where both policies were able to solve the task with a 100\% success rate, policy performance varies substantially across tasks. In the remaining tasks, except for \textit{pick up card from platform}, we observe that $\pi_{0.5}$ outperforms DP by a significant margin, likely due to it having a more powerful model as well as a large and diverse pretraining data mix. \textit{Plug in charger} appears to be the most difficult task, where $\pi_{0.5}$ leads with only 20\% success rate. The rollouts we examined suggest that the low performance of vision-only policies across different tasks traces to several recurring challenges tied to the policies' inability to perceive or react to contact events. We describe these challenges below, with references to how each manifests across DP and $\pi_{0.5}$ rollouts on the six selected tasks. 
Figure \ref{fig:vision-failure-patterns} shows the challenges and their typical examples across tasks. 

\textbf{Establishing and maintaining contact.} Every manipulation process requires contact to be established or maintained between the robot fingers and the target object, the target object and the environment, or both. However, vision-only policies frequently fail to sustain such required contact. In \textit{thread nut onto screw}, policies sometimes miss the contact between the gripper finger and the nut when trying to roll it, leading to more attempts to properly engage the threads and longer task completion time. In \textit{tighten screw with Allen key}, the Allen key can slip in-hand naturally over the course of tightening or when the tip becomes misaligned with the screw socket; in either case, it loses the contact needed to transmit torque when the robot fails to adjust its action to account for the shifted Allen key in-hand pose. 
In \textit{retrieve single utensil from bin}, the policies are prone to making grasping attempts inside the utensil bin without properly contacting the chopsticks, causing either task failure or significantly longer, on average $2\times$, task completion time compared to human demonstrations. In \textit{pick up card from platform}, the robot often fails to establish or maintain contact with the card during either sliding the card on the platform or pivoting the card to lift it up, leading to policies being unable to complete the task.

\textbf{Precise alignment.} Several tasks require the robot to precisely align an in-hand object with a target feature for the subsequent contact-driven step to succeed.
The tolerances involved in such tasks and the required alignment adjustment actions are usually visually subtle. For \textit{thread nut onto screw}, alignment errors are the dominant cause of failure, as even small offsets can cause the nut to fall off before the threads engage. When attempting to insert the Allen key tip into the screw socket in \textit{tighten screw with Allen key}, misalignment between them leads to the Allen key colliding with the screw head and slipping in hand, making it more difficult to engage the Allen key with the screw properly. For \textit{plug in charger}, alignment errors between the charging block pins and the outlet holes again dominate the failure cases. During demonstrations, the human operator often relies on tactile feedback to help guide and adjust the various alignment processes, especially for \textit{plug in charger} where the pins are occluded by the charging block body in the wrist camera view. Across these tasks, such small adjustment-after-contact behaviors were not observed in the policy rollouts, hence presumably to not have been learned when the policy input includes the vision modality alone.

\textbf{Avoiding unsafe forceful interactions.} Unless required by the task, excessive contact forces during task execution are usually undesired and can cause task failure or safety concerns. Vision-only policies struggle to avoid forceful collisions with task objects and environments across multiple tasks we evaluated them on. In \textit{thread nut onto screw}, the gripper fingers can collide with the base block, which leads to spikes in contact forces. In \textit{plug in charger}, the policy kept pushing the charging block into the outlet even when its pins are misaligned with the outlet holes and the insertion is blocked. In \textit{retrieve single utensil from bin} and \textit{pick up card from platform}, the robot can collide with the utensil bin or the flat platform and generate excessive forces. As shown in Table \ref{tab:results}, many rollouts generated contact forces above 40N. 

\textbf{Recognizing task-critical states.} Certain task-critical states can be ambiguous to the vision modality alone. In \textit{plug in charger}, whether the charger has been fully inserted can be ambiguous from the wrist camera view. Therefore, the policies sometimes fail the task even when the charging block is properly aligned with the outlet due to stopping the insertion motion before the charging block pins make contact with the end of the outlet slot, leaving the charging block only partially inserted.  In \textit{retrieve single utensil from bin}, vision-only policies are prone to grasping onto multiple chopsticks instead of just one, lacking the capability to recognize the number of chopsticks currently in between gripper fingers.

We also noticed that in \textit{tighten screw with Allen key}, the policies tend to keep repeating the tightening motion after the screw has already been tightened. Although there is a strong visual signal at the moment the screw becomes tightened (Allen key tilts in-hand), a tightened vs. untightened screw looks visually the same otherwise. Since neither DP nor $\pi_{0.5}$ has memories of task stages completed, both tend to continue trying to tighten the screw by repeating the lift Allen key $\rightarrow$ align with screw $\rightarrow$ turn Allen key steps as if the screw has not been tightened. In general, we observe that DP is more prone to getting stuck in a loop and not terminate task execution compared to $\pi_{0.5}$, as DP also gets stuck in a loop in \textit{retrieve single utensil from bin}. We view lack of memory as a crucial challenge for many types of manipulation tasks and not unique to contact-rich manipulation.

\begin{figure}
    \centering
    \includegraphics[width=\columnwidth]{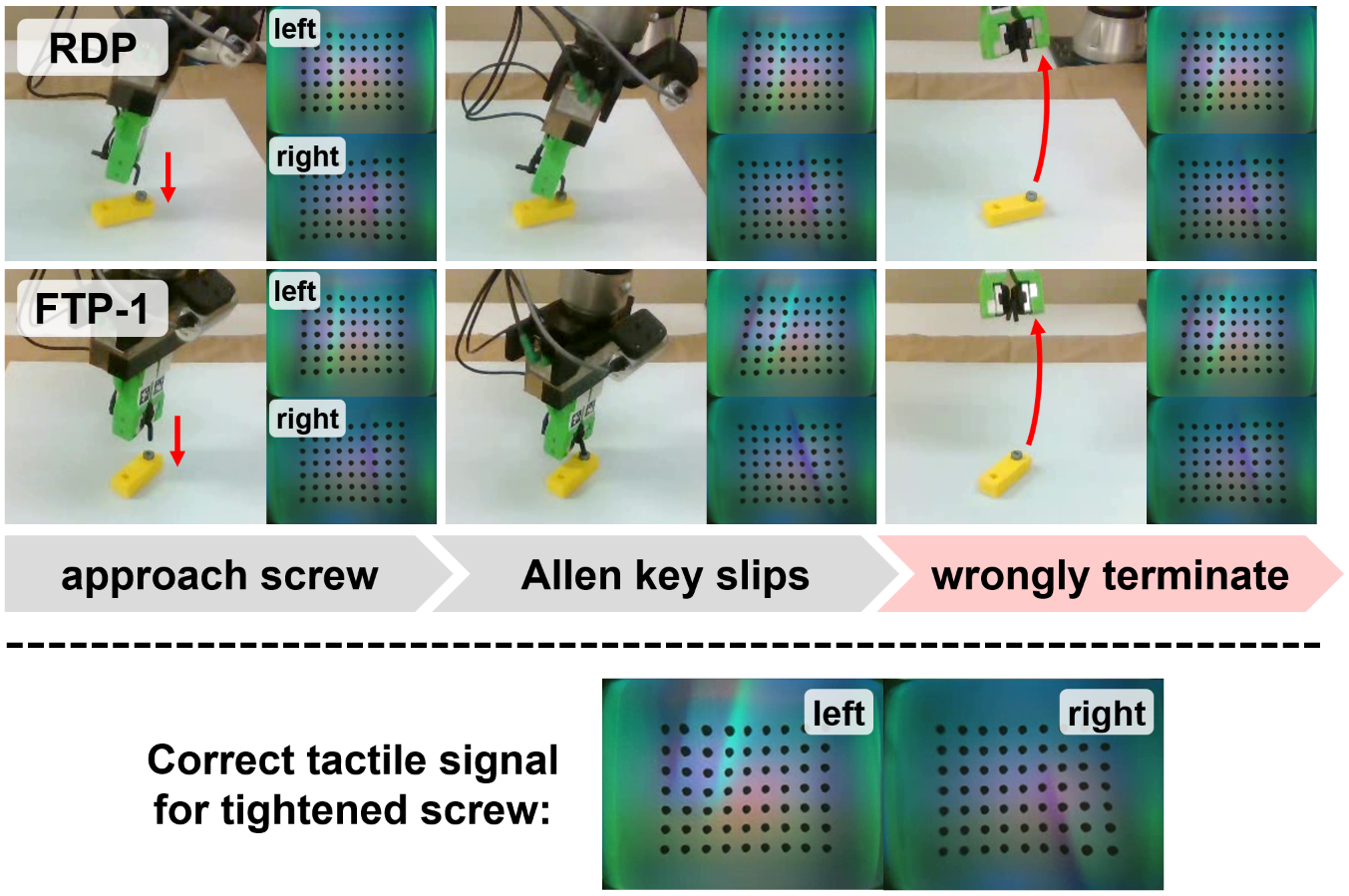}
    \caption{Top: Both RDP and FTP-1 are prone to terminating the tightening motion without tightening the screw after the Allen key slips in hand. Bottom: Although in-hand slip can happen naturally as the task progresses, the correct tactile signal associated with a tightened screw has a large in-plane torque.}
    \label{fig:allen-key-signal}
\end{figure}

\subsection{Do Visual-Tactile Policies Address These Challenges?}

\begin{figure*}[!t]
    \centering
    \includegraphics[width=\textwidth]{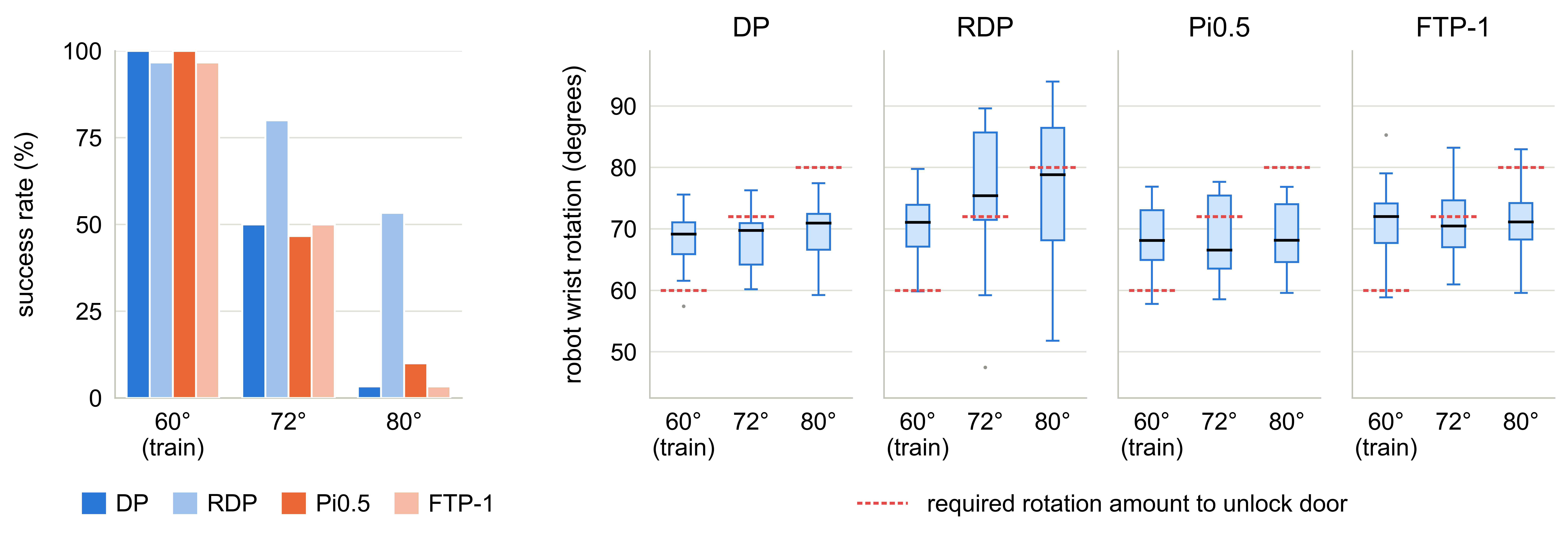}
    \caption{Left: Success rates across task variants for \textit{turn doorknob to open door}. Right: Robot wrist rotation angle distribution for each policy. Only RDP shows behavior adaptation effort when the task demands larger rotation angles to unlock the door.}
    \label{fig:wrist-rotation}
\end{figure*}

Tactile sensing is widely considered important for contact-rich manipulation, and intuitively, effective integration of force or tactile sensing can address the challenges discussed in Section \ref{subsec:vision-policy-challenges}. In this subsection, we examine whether Reactive Diffusion Policy (RDP) \cite{xue2025reactive} and Foundation Tactile Policy (FTP-1) \cite{ftp1}, two recent representative visual-tactile policy learning methods, address such challenges encountered by their vision-only base policies. We used the official implementations provided by RDP and FTP-1, which both natively support the GelSight Mini tactile sensor used in our experiments. Table \ref{tab:results} and Figure \ref{fig:checkpoint-prog} compare RDP and FTP-1 with their vision-only base policies, DP and $\pi_{0.5}$, on the four tasks for which we collected tactile data. On these tasks, RDP and FTP-1 did not meaningfully improve policy success rate, average task progress, interaction force, or completion time. Across policy rollouts on the four tasks, we observed three aspects that limited the benefit of the added tactile input from the GelSight Mini tactile sensor, which we discuss below.

\textbf{Sensor coverage limited the tactile information available to the policy.} In some cases, the policy performance is suboptimal not because the policies fail to properly react to the tactile input, but because the sensor never captured the relevant contact. In \textit{thread nut onto screw}, both vision-only and visual-tactile policies produced forceful collisions between the back of the gripper finger and the object. This region is outside the GelSight Mini's sensing area, which covers only the inner fingertip surface. As a result, the tactile input does not capture these collision events, and the visual-tactile policies have no additional information to avoid them compared to their vision-only counterparts. Tasks \textit{retrieve single utensil from bin} and \textit{pick up card from platform} illustrate the same limitation. Although vision-only policies performed poorly on both tasks, we could not collect visual-tactile demonstrations or train visual-tactile policies for them because GelSight Mini’s fingertip-only coverage and bulky form factor made it unsuitable for either task.

\textbf{Policies fail to learn the tactile-driven behaviors present in demonstrations}. 
During data collection, the demonstrator relied on tactile feedback for behaviors such as establishing and maintaining contact and tactile-driven alignment, which vision-only policies struggled to reproduce as discussed in Section \ref{subsec:vision-policy-challenges}. These signals are captured by tactile sensors and are used as input to RDP and FTP-1 during both training and inference. However, the two policies exhibited similar failures to their vision-only base policies, including improper engagement between the Allen key tip and the screw socket in \textit{tighten screw with Allen key}, and misalignment between task objects in \textit{thread nut onto screw}, \textit{tighten screw with Allen key}, and \textit{plug in charger}. This gap was most evident in \textit{plug in charger}, where the charging block pins and outlet holes are visually occluded during insertion, and the demonstrator relied on tactile feedback to search for the holes and align the pins with them. As noted in Section \ref{subsec:vision-policy-challenges}, the vision-only policies did not show this search behavior, and the same failure also occurred in the tactile-augmented policies despite receiving tactile signals that may have helped them avoid it.

\textbf{Policies acted on incidental tactile signals rather than task-relevant ones.} When the visual-tactile policies did react to tactile input, they sometimes reacted to signals that do not reliably indicate the current task state. In \textit{tighten screw with Allen key}, both visual-tactile policies tended to stop the tightening motion prematurely when the Allen key slipped enough in the gripper, as shown in Figure \ref{fig:allen-key-signal}. Such in-hand slip can occur naturally as the task progresses or from misalignment between the Allen key tip and the screw socket, hence it does not reliably indicate that the screw has been fully tightened. This premature termination contributed substantially to both visual-tactile policies underperforming their vision-only base policies on this task.

\subsection{Case Studies on Behavioral Generalization}

Section \ref{subsec:task-variations} introduces task variations designed to evaluate behavioral generalization capabilities of trained policies. Here we use \textit{turn doorknob to open door} to conduct two case studies on behavioral generalization. We select this task since it is the only one for which all four policies achieve near-perfect success rates, making it the ideal candidate where a drop in success rate under a variation can be attributed to the variation rather than the base task difficulty. We first test a behavioral variation in the amount of rotation required to unlock the door. We use doors that require approximately 72-degree and 80-degree rotations to unlock, where the base door used in training requires 60 degrees. The 72-degree and 80-degree doors are visually indistinguishable from the base 60-degree door. We also test a visual + behavioral variation in the doorknob geometry, where the original doorknob is replaced with one that is 5mm smaller in diameter.

\subsubsection{Behavioral variation: required rotation amount to unlock door} Figure \ref{fig:wrist-rotation} reports success rates across 30 rollouts for each of the four policies under the three rotation-amount conditions. While all four policies show degraded performance as the required rotation increases, the extent of degradation differs substantially. While the success rates of DP, $\pi_{0.5}$, and FTP-1 dropped by half for the 72-degree door and dropped to almost zero for the 80-degree door, RDP is able to partially retain its performance at 80\% and 53.3\% success rates.

To understand what is driving the performance gap between policies, we compute the actual amount by which each policy rotates the doorknob during rollouts, shown in Figure \ref{fig:wrist-rotation}. 
For DP, $\pi_{0.5}$, and FTP-1, their wrist rotation distributions remain largely unchanged across the base, 72-degree, and 80-degree conditions. This indicates that these policies rotate the doorknob by roughly the same amount regardless of how much rotation is required to unlock the door, and the successes that do occur under the variations correspond to rollouts where the natural spread of the wrist rotation distribution happens to reach the increased rotation requirement. RDP, by contrast, has its wrist rotation distribution shifted upward as the required rotation increases, indicating it is adapting to the task variations whereas the other three policies are not. 

These results show that tactile-augmented policies can achieve better behavioral generalization even when they offer no gains in the in-distribution task settings. However, how tactile feedback is integrated matters. FTP-1, which also uses tactile sensing, does not exhibit RDP’s adaptation behavior: its wrist rotation distribution remains flat across the three conditions, similar to the vision-only policies. Thus, for the task variations we tested here for \textit{turn doorknob to open door}, incorporating tactile feedback alone is insufficient, and RDP’s integration mechanism appears to be more beneficial.

\begin{table}[t]
\centering
\caption{Gripper width reached (mm) while trying to turn the doorknob, reported as mean $\pm$ std across episodes (30 episodes for the 60\,mm knob, 10 for the 55\,mm knob).}
\label{tab:knob-size-variation}
\footnotesize
\setlength{\tabcolsep}{3.5pt}
\begin{tabular}{lcccc}
\toprule
Variant & DP & RDP & $\pi_{0.5}$ & FTP-1 \\
\midrule
\begin{tabular}[c]{@{}l@{}}60\,mm knob\\(train)\end{tabular}
  & $58.9 \pm 0.12$ & $59.0 \pm 0.13$ & $58.8 \pm 0.13$ & $59.2 \pm 0.14$ \\
55\,mm knob
  & $59.0 \pm 0.11$ & $59.0 \pm 0.11$ & $58.7 \pm 0.11$ & $58.9 \pm 0.14$ \\
\bottomrule
\end{tabular}
\end{table}

\subsubsection{Visual + behavioral variation: smaller doorknob diameter} We replace the base 60mm doorknob with one 5mm smaller in diameter, subtly altering both the doorknob's appearance and the grasping width required to hold it. Across 10 rollouts per policy, all four policies achieved 0\% success rate. We further compute the gripper width each policy converged at for gripping the doorknob in Table \ref{tab:knob-size-variation}, which shows that all policies attempted to grip the smaller 55mm doorknob with the same gripper width used for the 60mm doorknob in the base configuration the policy was trained on. This causes the robot to miss the smaller doorknob entirely and fail to establish a grasp. 
Because the failure was immediate and consistent across every rollout of every policy, we did not run the full 30-rollout protocol on this variation.
This consistent failure to adjust the grasping width indicates that all four policies failed to learn a key aspect of grasping: adapting to the object’s size to achieve stable contact.
\section{Discussion}
\label{sec: discussion}

As a first release, \BenchmarkName{} makes several design choices that shape what it can and cannot evaluate. Here we discuss its coverage of tasks and materials, the limitations introduced by 3D-printed assets, and the broader evaluation of robot systems it is intended to support.

\textbf{Task and material coverage.}
The initial release of \BenchmarkName{} focuses on rigid and articulated objects. This scope reflects the time and resources available for the current study and captures only part of the diversity of everyday contact-rich manipulation. We envision the benchmark as a growing community effort rather than a fixed collection of tasks. Future extensions can increase the number of tasks and assets and introduce broader variations within existing tasks, enabling more comprehensive evaluation of skill acquisition and generalization. 
Beyond rigid and articulated objects, an important direction could be to incorporate different forms of deformable objects, fluids, and granular materials. These extensions will require corresponding physical assets, simulation capabilities, and evaluation protocols that account for the distinct manipulation challenges of these materials.

\textbf{Surface properties of printed assets.}
Although 3D printing facilitates reproducible fabrication, it also introduces surface characteristics that may differ from those of everyday manufactured objects. In particular, layer-induced surface texture can produce dense tactile features that are observable by sensors such as GelSight. 
This limitation concerns the transferability of learned policies beyond the benchmark assets. Furthermore, surface friction is not systematically characterized or varied in the current benchmark. Future work can investigate new formats of reproducible assets that can better leverage variance in surface texture and properties. 


\textbf{What evaluation will \BenchmarkName{} enable?}
\BenchmarkName{} is designed to evaluate robotic systems across a broad range of contact-rich manipulation tasks. Our experiments illustrate its use in evaluating manipulation policies, but performance also depends on other aspects of a robotic system, including mechanical design, embodiment selection, and sensor design, placement, and coverage. The benchmark can therefore support comparisons across these dimensions, helping researchers investigate how hardware and sensing choices affect manipulation performance. We envision \BenchmarkName{} as a shared experimental resource for evaluating complete robotic systems and informing the joint design of hardware, sensing, and control.



\section{Conclusion}
\label{sec: conclusion}

We presented \BenchmarkName{}, a benchmark for contact-rich robotic manipulation grounded in observations of everyday human activities. Organized around eight core manipulation skills, the benchmark provides \NumTasks{} reference tasks and \NumAssets{} standardized, 3D-printable assets with corresponding simulation models. Experiments across different 3D printers and comparisons between physical and simulated assets validate fabrication reproducibility and the correspondence of task-relevant mechanical responses, establishing a foundation for reproducible experimentation across laboratories. 
Beyond asset models, we provide a simulation suite integrating the benchmark assets, a GelSight tactile simulator, a teleoperation interface for data collection, and scripted policies.
Evaluating representative vision-only and visual-tactile policies on six benchmark tasks, we find that contact-rich manipulation remains challenging for the policy learning methods we tested.
We hope \BenchmarkName{} will serve as a shared testbed for developing, evaluating, and comparing robot systems that address these challenges.




\ifArXivVersion
{\footnotesize
\bibliographystyle{IEEEtran}
\bibliography{IEEEabrv,references}
}
\fi


\ifArXivVersion
  \clearpage  
\fi
\appendices
\makeatletter
\@addtoreset{figure}{section}
\@addtoreset{table}{section}
\@addtoreset{equation}{section}
\makeatother
\renewcommand{\thefigure}{\thesection\arabic{figure}}
\renewcommand{\thetable}{\thesection\arabic{table}}
\renewcommand{\theequation}{\thesection\arabic{equation}}
\def\theHfigure{appendix.\thesection.\arabic{figure}}
\def\theHtable{appendix.\thesection.\arabic{table}}
\def\theHequation{appendix.\thesection.\arabic{equation}}
\makeatletter
\setlength{\@fptop}{0pt}
\setlength{\@fpsep}{14pt}
\setlength{\@fpbot}{0pt plus 1fil}
\setlength{\@dblfptop}{0pt}
\setlength{\@dblfpsep}{20pt}
\setlength{\@dblfpbot}{0pt plus 1fil}
\makeatother

\section{Simulation Details}
\label{app-sec:mechanics}


\subsection{Asset modeling}
\label{app-sec:sim-assets}

\begin{figure}
    \centering
    \includegraphics[width=\columnwidth]{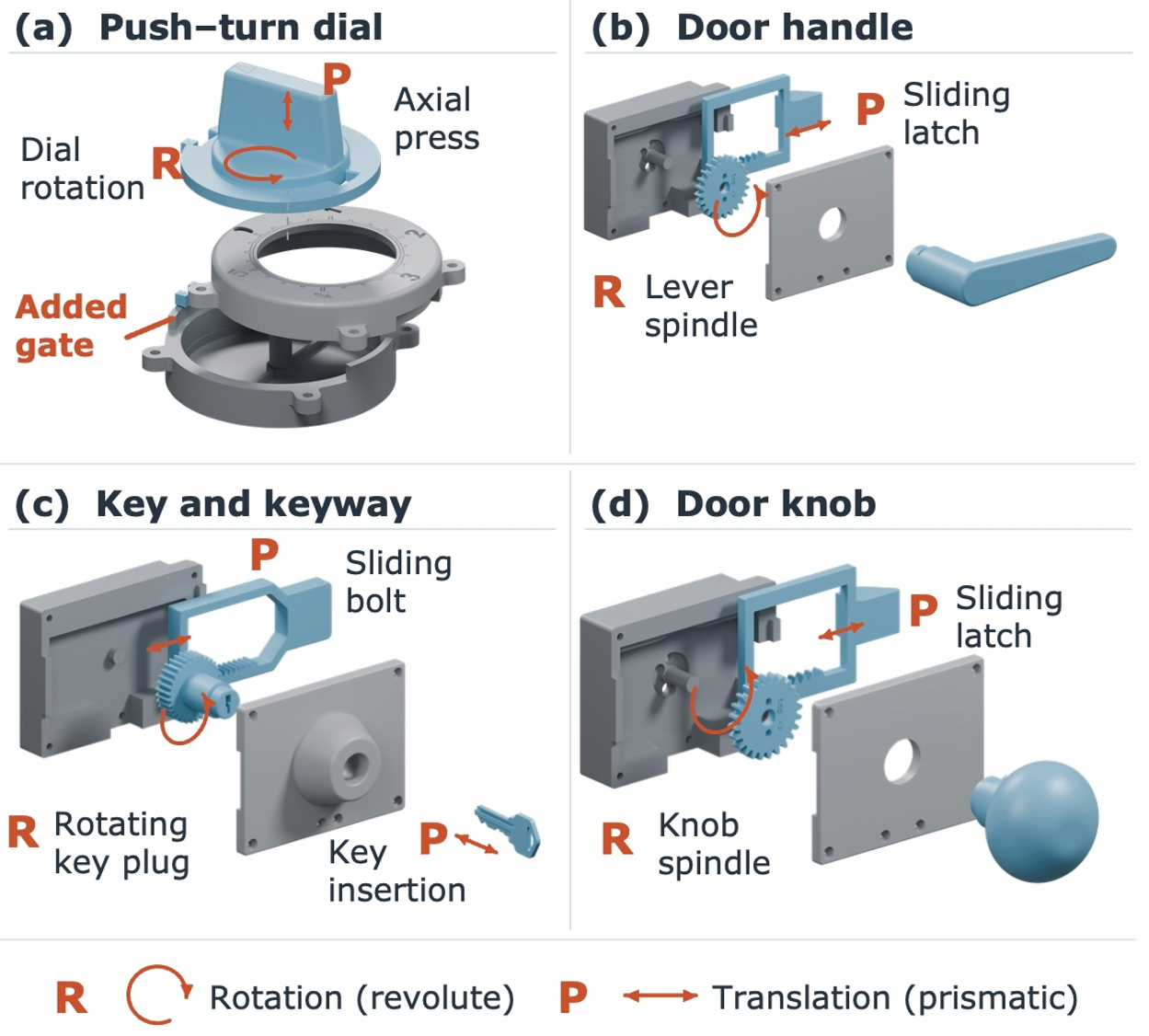}
    \caption{
        \textbf{Examples of CAD-derived simulation assets.}
        The assets approximate the corresponding physical mechanisms using
        rigid parts, joints, and simplified representations of compliance
        and mechanical constraints.
        Exploded views of
        \textbf{(a)} a push-and-turn dial,
        \textbf{(b)} a door handle,
        \textbf{(c)} a key-operated lock,
        \textbf{(d)} a doorknob, and
        \textbf{R} and \textbf{P} denote revolute and prismatic
        degrees of freedom, respectively.
    }
    \label{fig:sim-mechanisms}
\end{figure}

Fig.~\ref{fig:sim-mechanisms} shows how the modeling choices of Section~\ref{sec:asset-simulation} apply to four representative mechanisms; in each case, the model preserves the task-relevant contact behavior and motion constraints rather than the exact printed geometry.
The \emph{push-and-turn dial} pairs an axially compliant joint with a pin-and-post gate that approximates the small printed blocking features on the real part. Pressing the dial moves the pin downward, clear of the posts, so the dial can rotate; releasing it at the home orientation lets the spring push it back up. The \emph{door handle} retracts a sliding latch when turned and returns to its rest position through a revolute spring--damper. The \emph{key and keyway} uses axial sliding joint represents key insertion. The \emph{door knob} similarly converts spindle rotation into latch retraction, with contact between the latch and strike plate keeping the door shut until the latch clears. For these three door mechanisms, we use a reduced transmission model that maps spindle rotation to rack displacement and reflects any load on the rack back to the spindle as torque.


\subsection{Tactile sensing simulation}
\label{app-sec:sim-tactile}

This section gives the formulation of the tactile sensing module summarized in Section~\ref{sec:asset-simulation}. Within the hydroelastic pressure-field framework~\cite{elandt2019pressure}, the gel follows the pressure--indentation relation
\begin{equation}
    p_{\mathrm{law}}(\delta)
    = k_n \delta
      \left[1+\alpha\left(\frac{\delta}{h}\right)^2\right],
    \label{eq:sim-gel-pressure}
\end{equation}
where $\delta \geq 0$ is local indentation, $h$ is gel thickness, $k_n$ is the normal foundation stiffness, and $\alpha \geq 0$ controls stiffening under compression. Setting $\alpha=0$ recovers a linear response.

\begin{figure}
    \centering
    \includegraphics[width=\columnwidth]{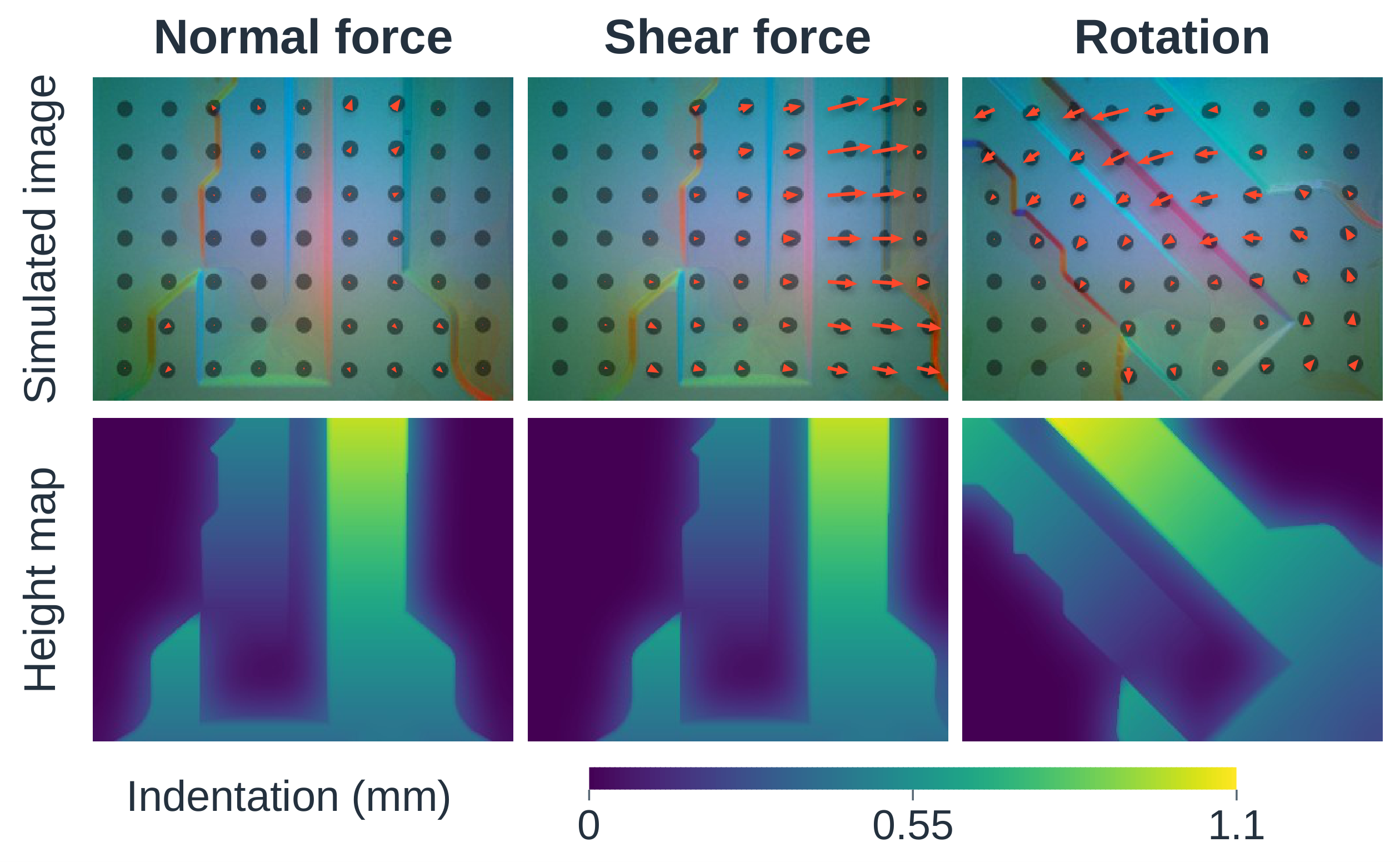}
    \caption{
        Simulated GelSight Mini responses to a key under normal loading,
        shearing, and rotation. Top: rendered tactile images, with red arrows
        indicating marker displacement. Bottom: gel indentation height maps on
        a common color scale.
    }
    \label{fig:sim-tactile}
\end{figure}

The contact surface is rasterized into a normal-pressure map $p_0(\mathbf{x})$. For nonzero contact, this map is normalized as
\begin{equation}
    \widehat{p}(\mathbf{x})
    = \frac{F_n^{\mathrm{sim}}}
           {\int_{\Omega} p_0(\mathbf{x})\,\mathrm{d}A}
      p_0(\mathbf{x}),
    \label{eq:sim-force-normalization}
\end{equation}
where $\Omega$ is the sensing domain and $F_n^{\mathrm{sim}}$ is the solver-resolved normal load on the gel pad. Pressure exceeding the full-compression level is redistributed to neighboring cells, yielding the pressure map $p$ used for subsequent deformation computation.

At each contacting surface location, tangential displacement is updated by
\begin{equation}
\begin{aligned}
    \mathbf{u}_{t+\Delta t}
    &= \Pi_{\|\mathbf{u}\|\leq u_{\mathrm{lim}}}
       \left(\mathbf{u}_t+\mathbf{v}_{\parallel}\Delta t\right),\\
    u_{\mathrm{lim}}
    &= \min\left(\frac{\mu p}{k_s},\,u_{\mathrm{cap}}\right),
\end{aligned}
\label{eq:sim-gel-shear}
\end{equation}
where $\mathbf{v}_{\parallel}$ is the local relative tangential velocity, $k_s$ is the shear foundation stiffness, $\mu$ is the friction coefficient, and $u_{\mathrm{cap}}$ limits shear displacement. The operator $\Pi$ projects onto the specified displacement bound, ensuring that the shear traction $\boldsymbol{\tau}=k_s\mathbf{u}$ satisfies $\|\boldsymbol{\tau}\|\leq\mu p$.

The resulting deformation fields are rendered into tactile images with the Taxim-based optical shading~\cite{si2022taxim} described in Section~\ref{sec:asset-simulation}. Together, the module's outputs expose contact geometry and surface texture, distributed force and torque, incipient slip, and the aggregate wrench on each finger.

Figure~\ref{fig:sim-tactile} compares the simulated sensor responses
to normal loading, tangential shearing, and rotation of a key,
shown from left to right. The top row shows rendered tactile images,
with red arrows indicating the direction and relative magnitude of
marker displacement. The bottom row shows the corresponding height
maps of gel indentation in millimeters, using a common color scale
across all three conditions.

\onecolumn
\section{Policy Training Details}
\label{sec:policy-training-details}

\edef\PTDparindent{\the\parindent}%
\noindent
\begin{minipage}[t]{\dimexpr(\textwidth-\columnsep)/2\relax}
\vspace{0pt}
\setlength{\parindent}{\PTDparindent}%
We use the official implementations of the four policy learning methods we evaluated in Section~\ref{sec: experiments}:
\begin{itemize}[nolistsep]
    \item \href{https://github.com/real-stanford/diffusion_policy/tree/5ba07ac6661db573af695b419a7947ecb704690f}{Diffusion Policy (commit 5ba07ac)}
    \item \href{https://github.com/xiaoxiaoxh/reactive_diffusion_policy/tree/8ed82a351648cb3ea3d3a2ae626874150fef148a}{Reactive Diffusion Policy (commit 8ed82a3)}
    \item \href{https://github.com/Physical-Intelligence/openpi/tree/15a9616a00943ada6c20a0f158e3adb39df2ccac}{$\pi_{0.5}$ (commit 15a9616)}
    \item \href{https://github.com/michaelyuancb/ftp1-policy/tree/89fa681d6c014cce28300946b7526db808e0b1c1}{Foundation Tactile Policy (commit 89fa681)}
\end{itemize}

We show our real-world setup in Fig.~\ref{fig:policy-training-hardware-setup}, policy training details in Table~\ref{tab:policy_training}, and demonstration statistics and the language instructions used for $\pi_{0.5}$ and FTP-1 in Table~\ref{tab:demo_stats}.
\end{minipage}\hfill
\begin{minipage}[t]{\dimexpr(\textwidth-\columnsep)/2\relax}
\vspace{0pt}%
\centering
\includegraphics[width=0.95\columnwidth]{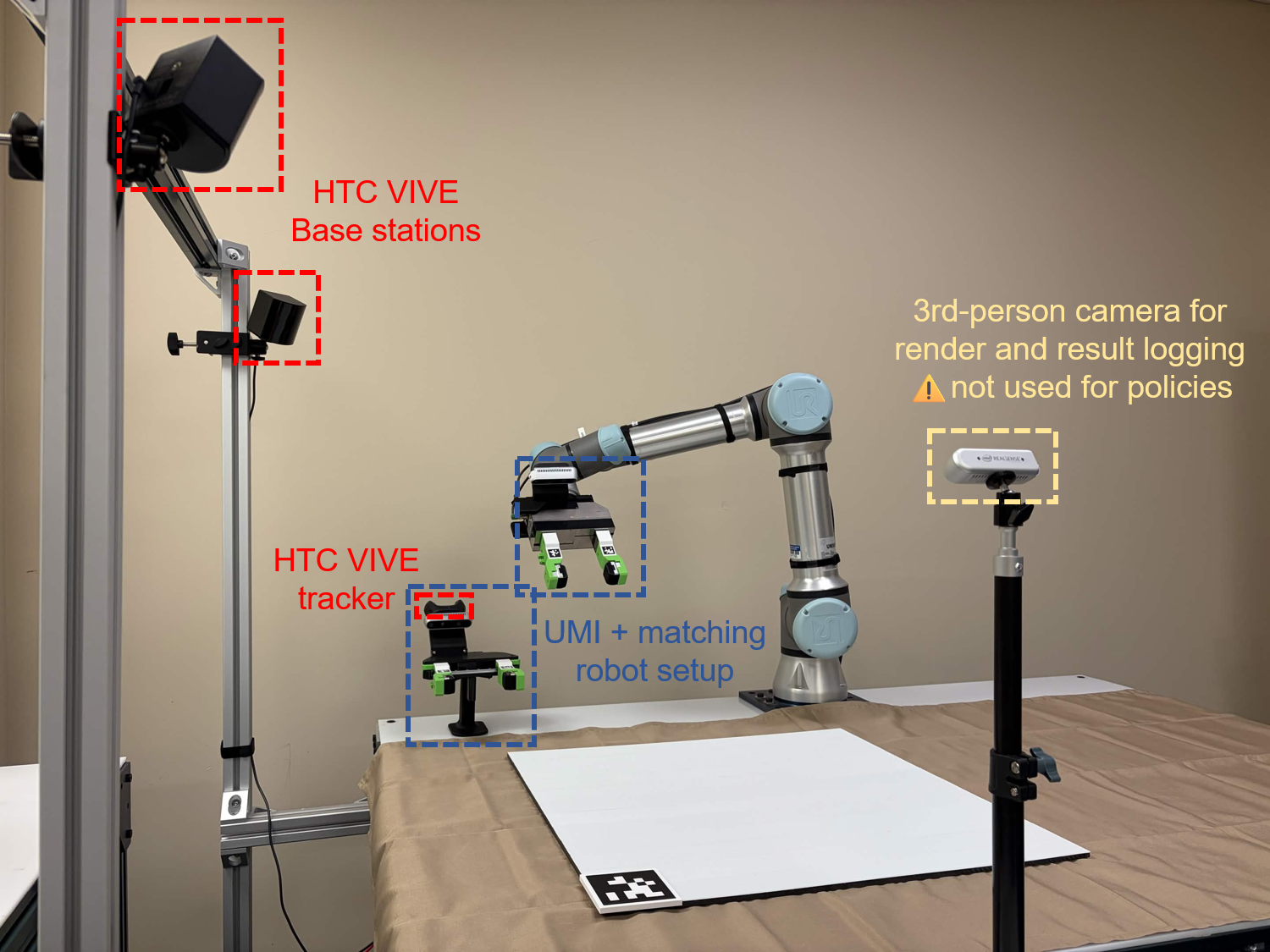}
\captionof{figure}{Our setup for demo collection and policy rollout.}
\label{fig:policy-training-hardware-setup}
\end{minipage}

\begin{table}[!htb]
\centering
\caption{Training details of the evaluated policies.}
\label{tab:policy_training}
\footnotesize
\renewcommand{\arraystretch}{1.05}
\setlength{\tabcolsep}{4pt}
\newcommand{\RowSep}{\specialrule{\cmidrulewidth}{0pt}{1pt}}
\begin{tabularx}{\textwidth}{@{}l*{4}{>{\centering\arraybackslash}X}@{}}
\toprule
 & \textbf{DP} & \textbf{RDP} & {\boldmath$\pi_{0.5}$} & \textbf{FTP-1} \\
\midrule
Initialization
  & From scratch
  & From scratch
  & \texttt{pi05\_base}, full fine-tuning
  & \texttt{ftp1\_v0426\_50kstep}, full fine-tuning \\
\RowSep
Data rate
  & 12\,Hz
  & 12\,Hz (slow policy), 24\,Hz (fast policy)
  & 24\,Hz
  & 24\,Hz \\
\RowSep
Visual input
  & Wrist RGB, $240\times320$
  & Wrist RGB, $240\times320$
  & Wrist RGB, $240\times320$, resized then zero-padded to $224\times224$
  & Wrist RGB, $240\times320$, resized then zero-padded to $224\times224$ \\
\RowSep
Tactile input
  & --
  & $2\times$ GelSight Mini marker movements, each projected to 15-D with the PCA basis released with RDP
  & --
  & $2\times$ GelSight Mini images, each resized to $224\times224$; tactile encoder initialized from FTP-1's pretrained GelSight Mini checkpoint \\
\RowSep
Proprioceptive input
  & EEF pose (relative to current), absolute gripper width (normalized)
  & EEF pose (relative to current), absolute gripper width (normalized)
  & Absolute gripper width (normalized); EEF pose set to identity
  & Absolute gripper width; EEF pose set to identity \\
\RowSep
Observation history
  & 2 steps
  & 2 steps
  & 1 step (current only)
  & 1 step (current only) \\
\RowSep
Action
  & \multicolumn{4}{c}{10-D: position (3) and 6D rotation (6) relative to the current EEF pose, absolute gripper width (1)} \\
\RowSep
Prediction horizon
  & 16 steps (1.33\,s)
  & 32 steps (1.33\,s)
  & 32 steps (1.33\,s)
  & 32 steps (1.33\,s) \\
\RowSep
Batch size
  & 64
  & 64
  & 64
  & 64 \\
\RowSep
Peak LR / schedule
  & $1\times10^{-4}$; cosine decay, 500 warmup steps
  & AT: $1\times10^{-3}$, cosine decay, 100 warmup steps; LDP: $1\times10^{-4}$, cosine decay, 500 warmup steps
  & $5\times10^{-5}$; cosine decay to $5\times10^{-6}$, 500 warmup steps
  & $5\times10^{-5}$; cosine decay to $5\times10^{-6}$, 500 warmup steps \\
\RowSep
Training length
  & 501 epochs
  & AT: 601 epochs; LDP: 501 epochs
  & 40k steps for \textit{thread nut onto screw} and \textit{tighten screw with Allen key}, 20k for the rest
  & 40k steps for \textit{thread nut onto screw} and \textit{tighten screw with Allen key}, 20k for the rest \\
\RowSep
Hardware
  & 1$\times$ RTX 5090
  & 1$\times$ RTX 5090
  & 2--4$\times$ GH200
  & 2$\times$ GH200 \\
\bottomrule
\end{tabularx}

\vspace{3pt}
\parbox{\textwidth}{\footnotesize EEF: end effector. AT: asymmetric tokenizer (RDP's fast policy). LDP: latent diffusion policy (RDP's slow policy).}
\end{table}

\begin{table}[!htb]
\centering
\caption{Demonstrations used for training and language instructions given to $\pi_{0.5}$ and FTP-1.}
\label{tab:demo_stats}
\label{tab:language_instructions}
\footnotesize
\renewcommand{\arraystretch}{1.15}
\begin{tabular*}{\textwidth}{@{\extracolsep{\fill}}lcccl@{}}
\toprule
Task & Demos & Mean (s) & Min--max (s) & Language instruction \\
\midrule
Thread nut onto screw            & 100        & 19.6 & 14.0--24.0 & ``screw the nut onto the screw'' \\
Tighten screw with Allen key     & 100        & 21.0 & 11.9--32.7 & ``tighten the screw with allen key'' \\
Plug in charger                  & 100        & 8.4  & 7.0--10.0  & ``plug the charger into the outlet'' \\
Turn doorknob to open door       & 99$^{*}$   & 8.8  & 7.5--11.0  & ``turning the door knob to open the door'' \\
Retrieve single utensil from bin & 100        & 12.9 & 8.1--27.5  & ``pick up a single utensil from the storage bin'' \\
Pick up card from platform       & 100        & 8.6  & 7.0--10.5  & ``pick up the thin object on the table'' \\
\bottomrule
\end{tabular*}

\vspace{3pt}
\parbox{\textwidth}{\footnotesize All demonstrations were recorded at 24\,Hz with the UMI gripper. $^{*}$One of the 100 recorded demonstrations was dropped due to not passing data quality checks for the extracted UMI gripper widths. FTP-1 uses the same instructions with a terminating period.}
\end{table}

\ifArXivVersion
\onecolumn  
\section{Detailed Benchmark Task Descriptions}
\label{sec:full_task}
\begingroup
  \footnotesize  
  \renewcommand{\arraystretch}{1.2}
  \setlength{\aboverulesep}{4pt}
  \setlength{\belowrulesep}{4pt}
  \renewcommand{\tabularxcolumn}[1]{m{#1}}  

  \begin{xltabular}{\textwidth}{@{} >{\centering\arraybackslash}m{0.4\textwidth} T @{}}

    \caption{Base benchmark tasks}
    \label{tab:base-benchmark-tasks} \\
    \toprule
    \textbf{Task Execution Sequence} & \multicolumn{1}{c@{}}{\textbf{Task Specs}} \\
    \endfirsthead

    \toprule
    \textbf{Task Execution Sequence} & \multicolumn{1}{c@{}}{\textbf{Task Specs}} \\
    \endhead

    \bottomrule
    \endfoot
    \bottomrule
    \endlastfoot


    \entryrule
    \includegraphics[width=\linewidth]{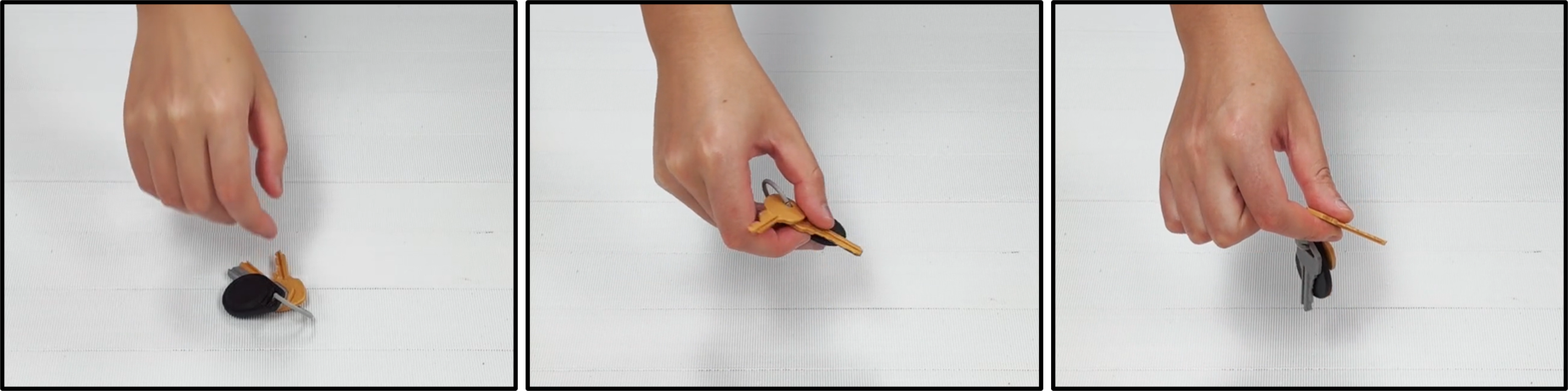} &
    \textbf{Task:} Grasp target key from keychain

    \textbf{Skills:} Grasping from clutter, grasping thin or small objects

    \textbf{Description:} For a keychain in the workspace with multiple keys, find and singulate the specified target key among all keys on the keychain.

    \textbf{Success criteria:} The robot movement stops with only the specified target key stably grasped within the robot end-effector.
    \\


    \entryrule
    \includegraphics[width=\linewidth]{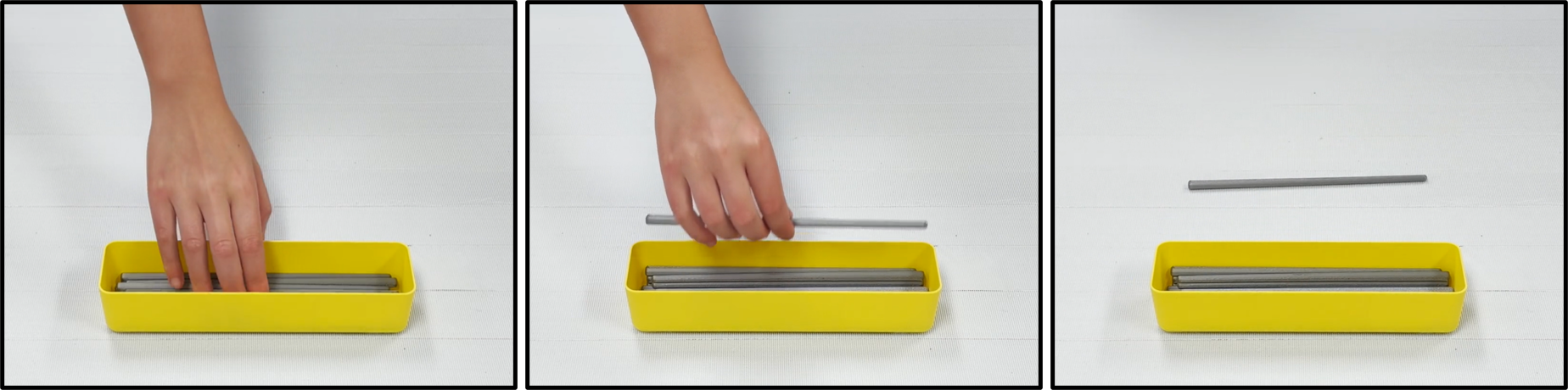} &
    \textbf{Task:} Retrieve single utensil from bin

    \textbf{Skills:} Grasping from clutter, grasping thin or small objects

    \textbf{Description:} A utensil storage bin is placed in the workspace. The target utensil is placed in the bin alongside multiple other utensils. Pick up a single copy of the target utensil from the cluttered storage bin and place it on the table without removing other objects from the bin.

    \textbf{Success criteria:} The robot movement stops after separating a single copy of the target utensil and placing it on the table. Other utensils should remain in the storage bin. The robot should not remain in contact with the task objects.
    \\


    \entryrule
    \includegraphics[width=\linewidth]{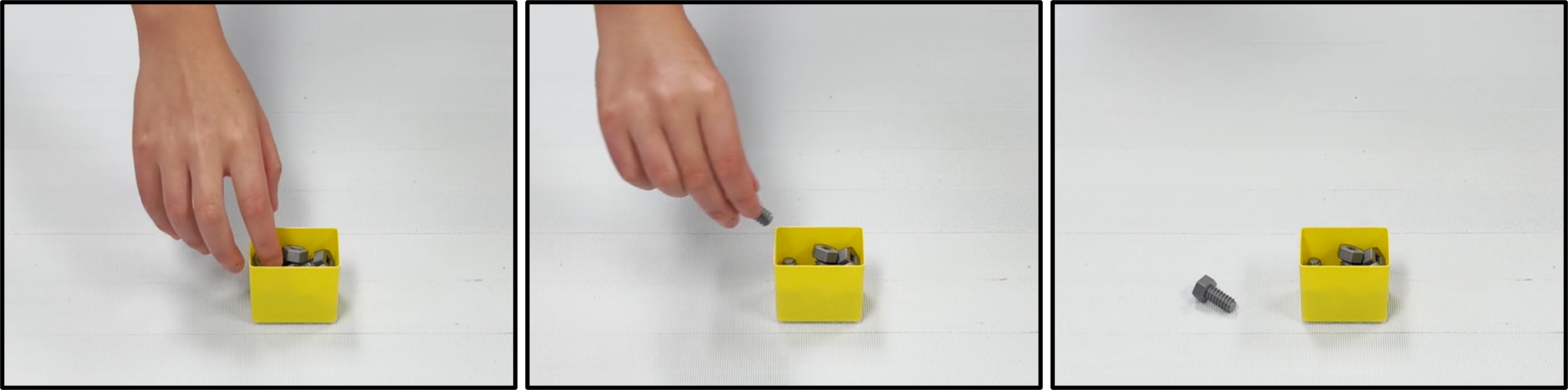} &
    \textbf{Task:} Retrieve single fastener from bin

    \textbf{Skills:} Grasping from clutter, grasping thin or small objects

    \textbf{Description:} A fastener storage bin is placed in the workspace. The target fastener is placed in the bin alongside multiple other fasteners. Pick up a single copy of the target fastener from the cluttered storage bin and place it on the table without removing other objects from the bin.

    \textbf{Success criteria:} The robot movement stops after separating a single copy of the target fastener and placing it on the table. Other fasteners should remain in the storage bin. The robot should not remain in contact with the task objects.
    \\


    \entryrule
    \includegraphics[width=\linewidth]{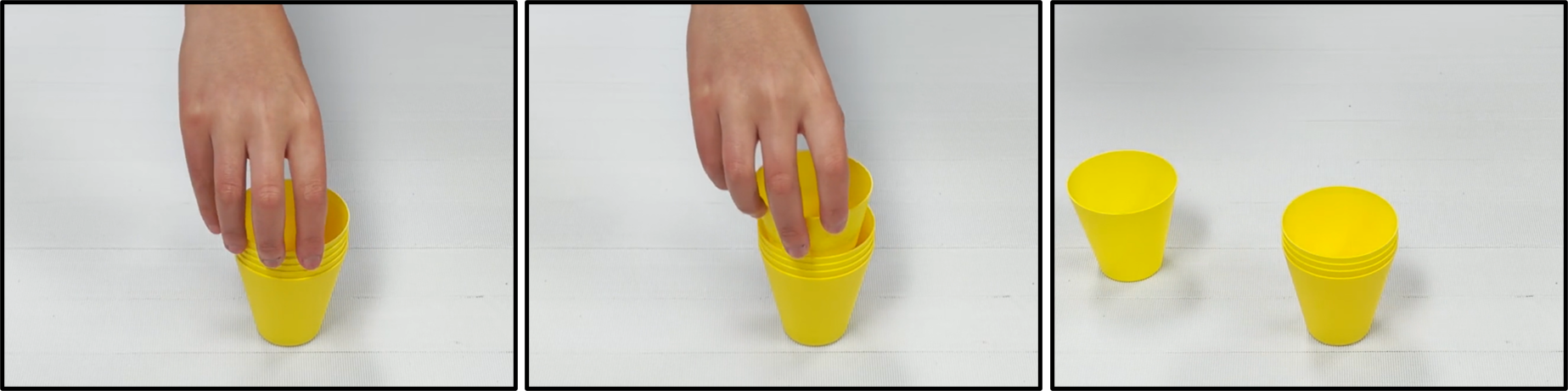} &
    \textbf{Task:} Retrieve top cup from stack

    \textbf{Skills:} Grasping from clutter

    \textbf{Description:} A neatly organized stack of cups is placed in the workspace. Pick up the top cup from the stack and place it on the table without disturbing the rest of the stack.

    \textbf{Success criteria:} The robot movement stops after separating the top cup from the stack and placing it on the table. The remaining cups should remain in a neat stack. The robot should not remain in contact with the task objects.
    \\


    \entryrule
    \includegraphics[width=\linewidth]{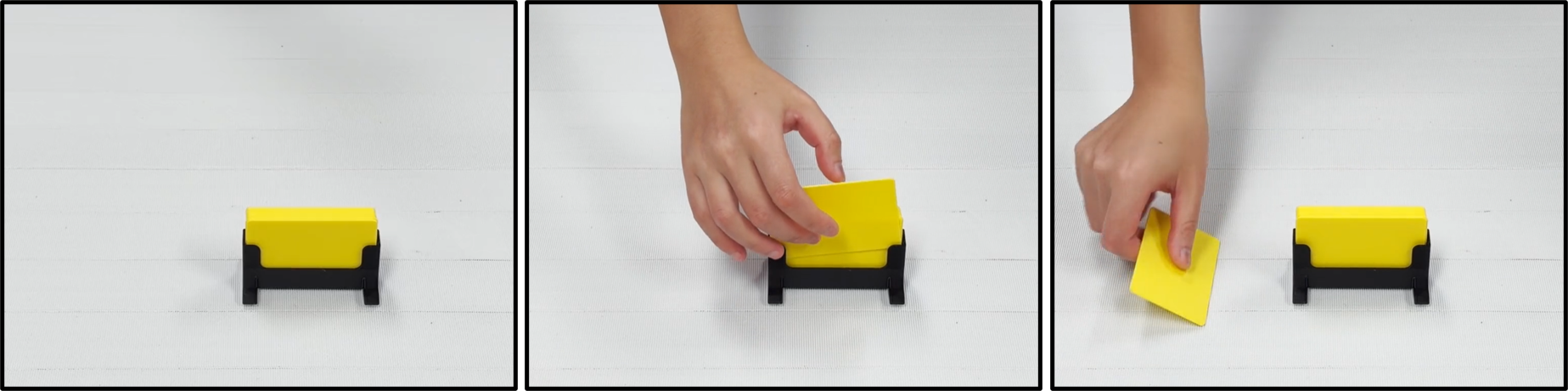}
    &
    \textbf{Task:} Retrieve single card from deck

    \textbf{Skills:} Grasping from clutter, grasping thin or small objects

    \textbf{Description:} A neatly organized deck of cards is placed in a cardholder in the workspace. Pick up a single card and place it on the table without disrupting the rest of the card deck.

    \textbf{Success criteria:} The robot movement stops after separating a single card from the deck and placing it on the table. The remaining cards should remain a neat deck in the cardholder. The robot should not remain in contact with the task objects.
    \\


    \entryrule
    \includegraphics[width=\linewidth]{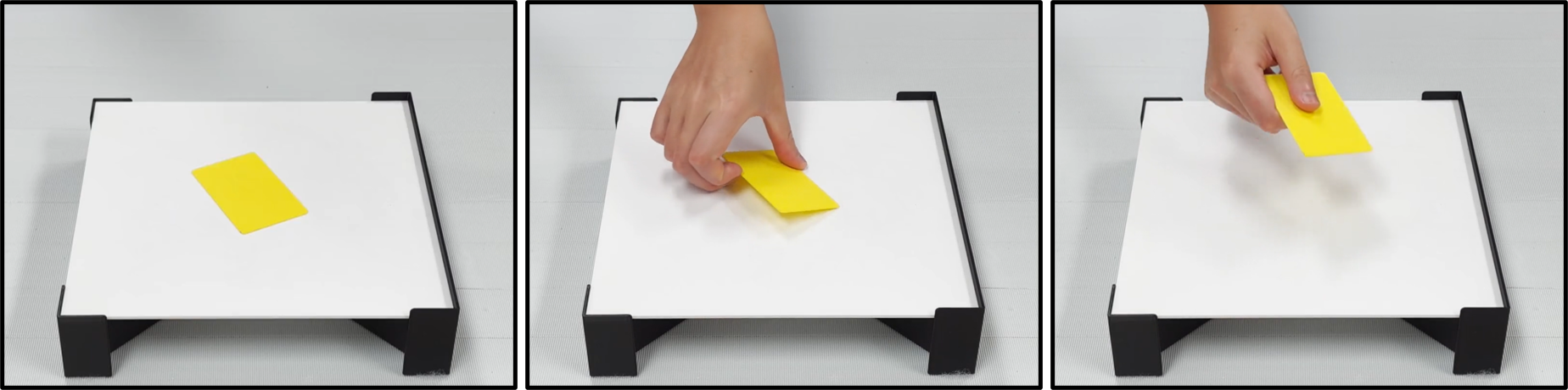}
    &
    \textbf{Task:} Pick up object from platform

    \textbf{Skills:} Grasping thin or small objects

    \textbf{Description:} Pick up the thin and/or small object placed on the flat platform.

    \textbf{Success criteria:} The target object is stably grasped by the robot and lifted off the platform.
    \\


    \entryrule
    \includegraphics[width=\linewidth]{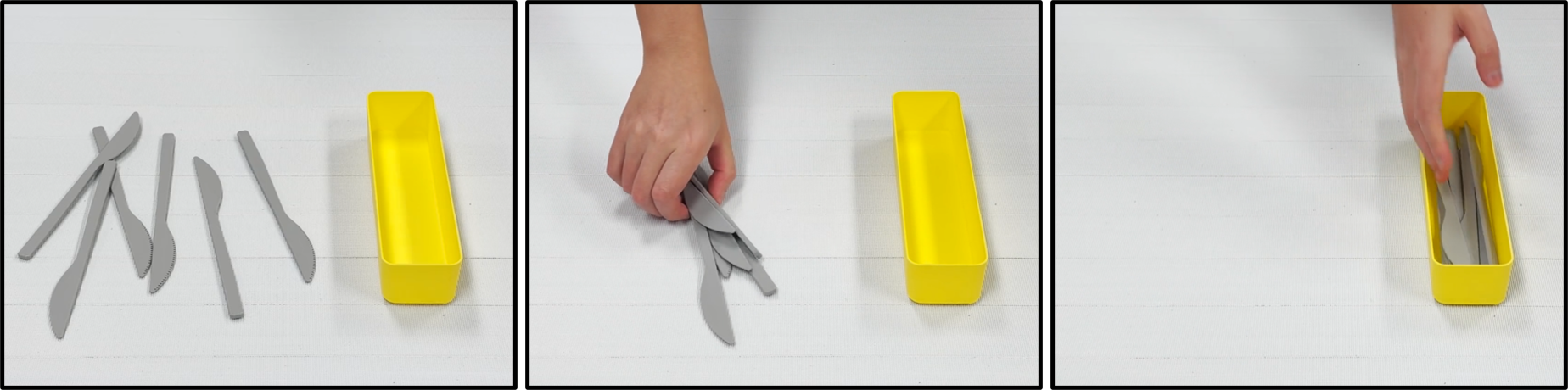}
    &
    \textbf{Task:} Collect utensils

    \textbf{Skills:} Grasping multiple objects, placing under environmental constraints

    \textbf{Description:} Collect all utensils scattered across the workspace and place them into the storage bin. Being able to hold multiple utensils simultaneously will greatly increase task completion efficiency.

    \textbf{Success criteria:} The robot movement stops after placing all utensils inside the storage bin. The robot should not remain in contact with the task objects.
    \\


    \entryrule
    \includegraphics[width=\linewidth]{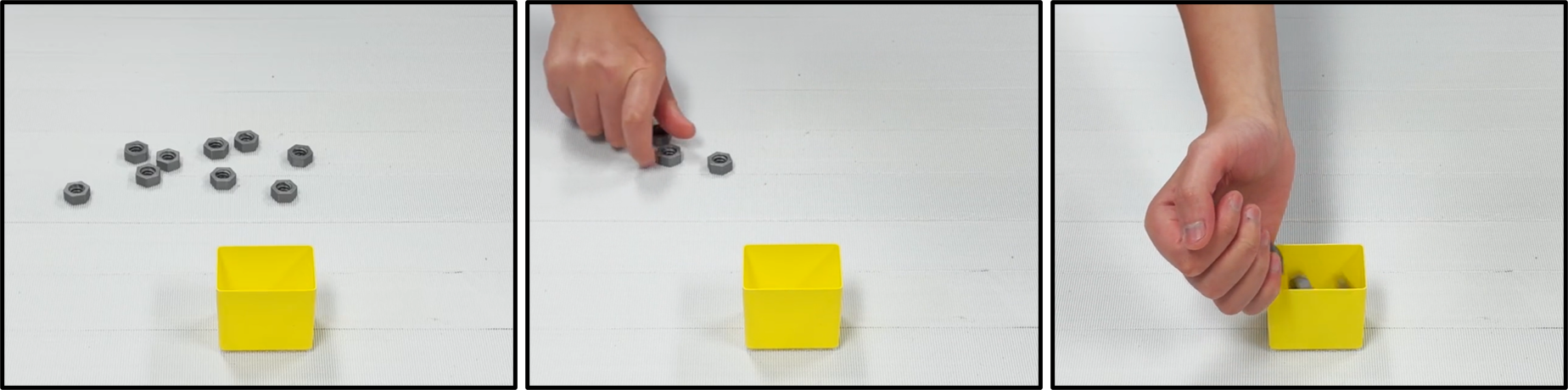}
    &
    \textbf{Task:} Collect fasteners

    \textbf{Skills:} Grasping multiple objects, placing under environmental constraints

    \textbf{Description:} Collect all fasteners scattered across the workspace and place them into the storage bin. Being able to hold multiple fasteners simultaneously will greatly increase task completion efficiency.

    \textbf{Success criteria:} The robot movement stops after placing all fasteners inside the storage bin. The robot should not remain in contact with the task objects.
    \\


    \entryrule
    \includegraphics[width=\linewidth]{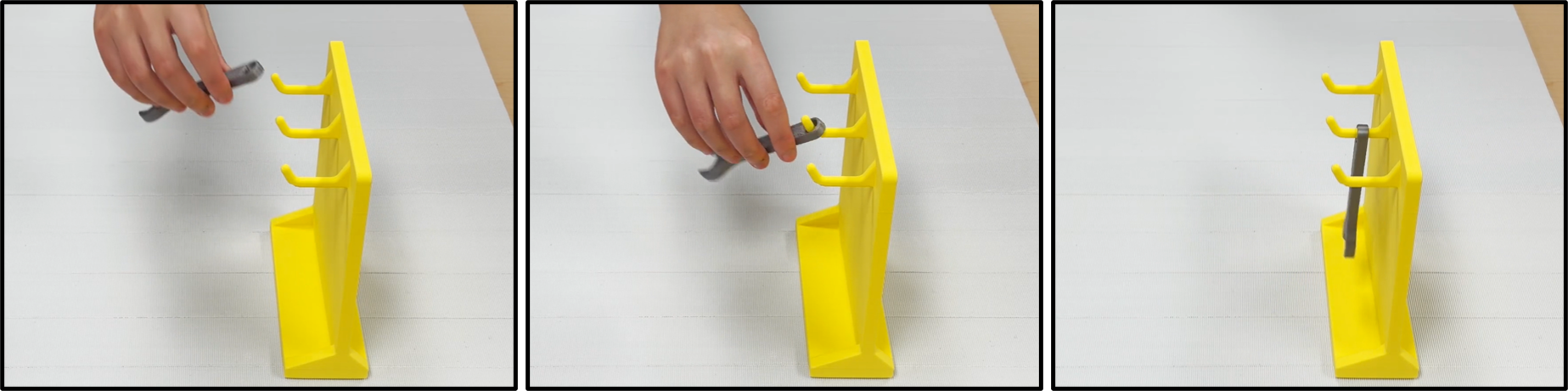}
    &
    \textbf{Task:} Hang object

    \textbf{Skills:} Placing under environmental constraints

    \textbf{Description:} Hang the object in-hand onto an empty hanging board hook.

    \textbf{Success criteria:} The robot movement stops after placing the target object onto a hanging hook. The robot should not remain in contact with the task objects.
    \\


    \entryrule
    \includegraphics[width=\linewidth]{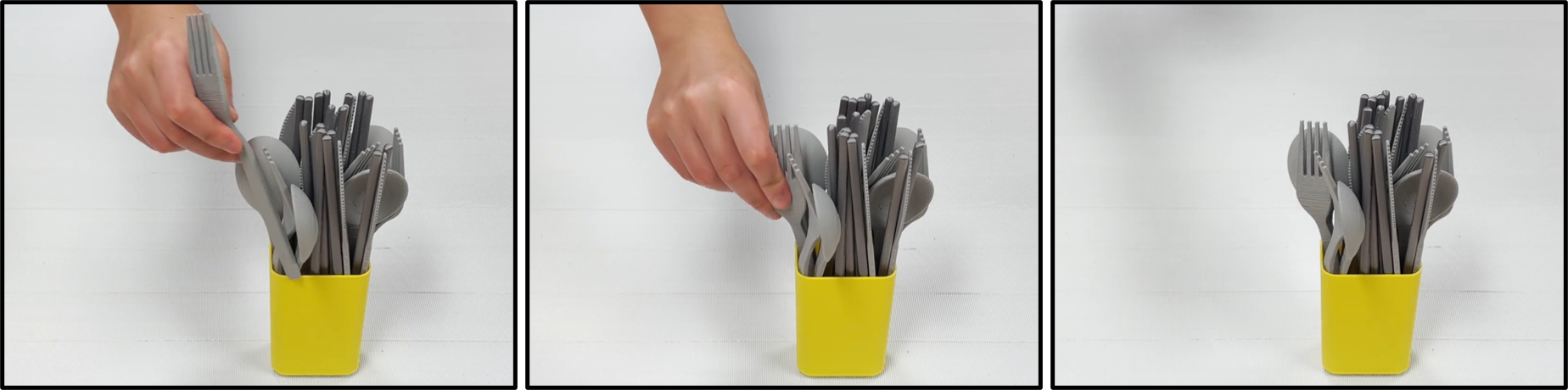}
    &
    \textbf{Task:} Place utensil into bin

    \textbf{Skills:} Placing under environmental constraints

    \textbf{Description:} A utensil bin with utensils already stored in it is placed in the workspace. Starting with a utensil held in hand, find an open pocket in the cluttered utensil storage bin and place the held utensil in it.

    \textbf{Success criteria:} The robot movement stops after placing the target object into the storage bin. The robot should not remain in contact with the task objects.
    \\


    \entryrule
    \includegraphics[width=\linewidth]{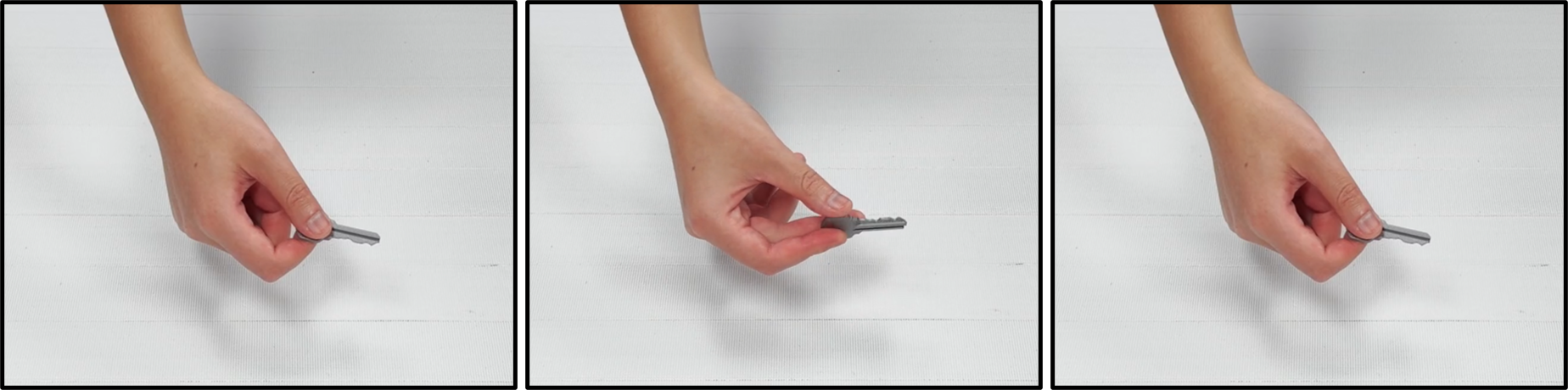}
    &
    \textbf{Task:} In-hand repose

    \textbf{Skills:} In-hand repositioning

    \textbf{Description:} Manipulate the target object in-hand, such that it ends up in the desired pose relative to the hand. Examples include flipping the orientation of a key or a USB stick to align with the slot.

    \textbf{Success criteria:} The target object pose converges and stabilizes to the desired pose, and remains within the robot end-effector throughout the execution without being dropped.
    \\


    \entryrule
    \includegraphics[width=\linewidth]{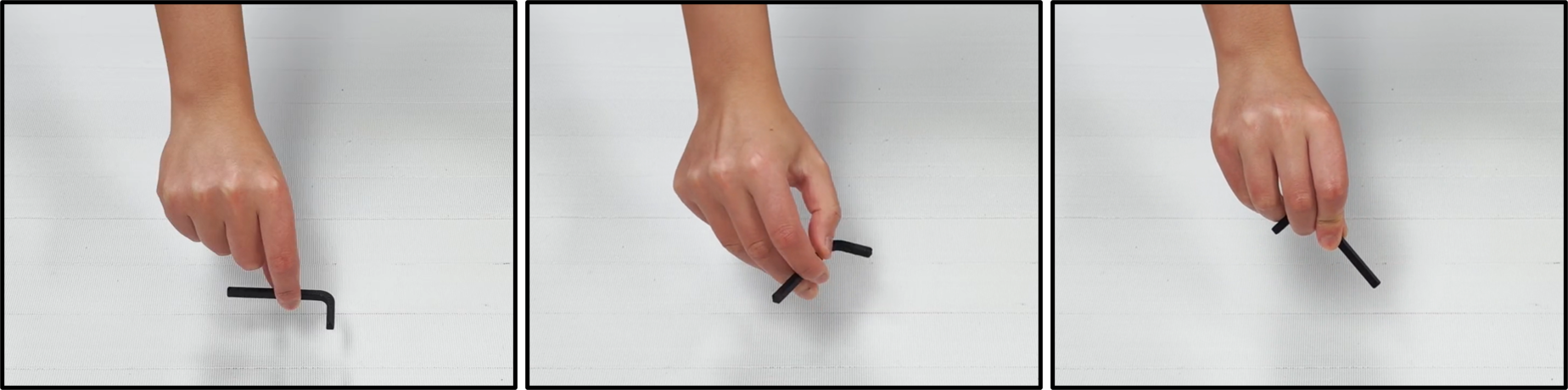}
    &
    \textbf{Task:} In-hand grasp transition

    \textbf{Skills:} In-hand repositioning

    \textbf{Description:} Manipulate the target object in-hand, such that it ends up with the desired grasp type (see grasp types defined in the Grasp Taxonomy \cite{grasp-taxonomy}).

    \textbf{Success criteria:} The target object pose converges and stabilizes with the desired grasp type, and remains within the robot end-effector throughout the execution without being dropped.
    \\


    \entryrule
    \includegraphics[width=\linewidth]{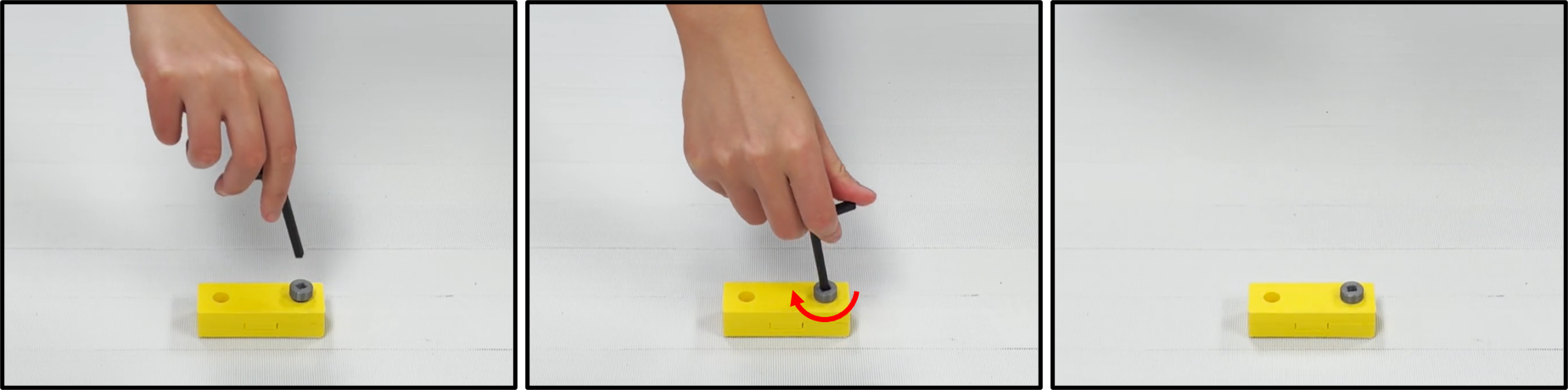}
    &
    \textbf{Task:} Tighten screw with Allen key

    \textbf{Skills:} Extrinsic contact control, insertion and alignment

    \textbf{Description:} A screw is placed into the screw slot of an assembly block in the workspace, but the threads are unengaged. Use the Allen key grasped in hand to tighten the screw.

    \textbf{Success criteria:} The screw is tightened and remains stationary when a small force is applied in an attempt to rotate the screw. The robot movement stops after terminating the tightening motion, moving away from, and breaking contact with the screw.
    \\


    \entryrule
    \includegraphics[width=\linewidth]{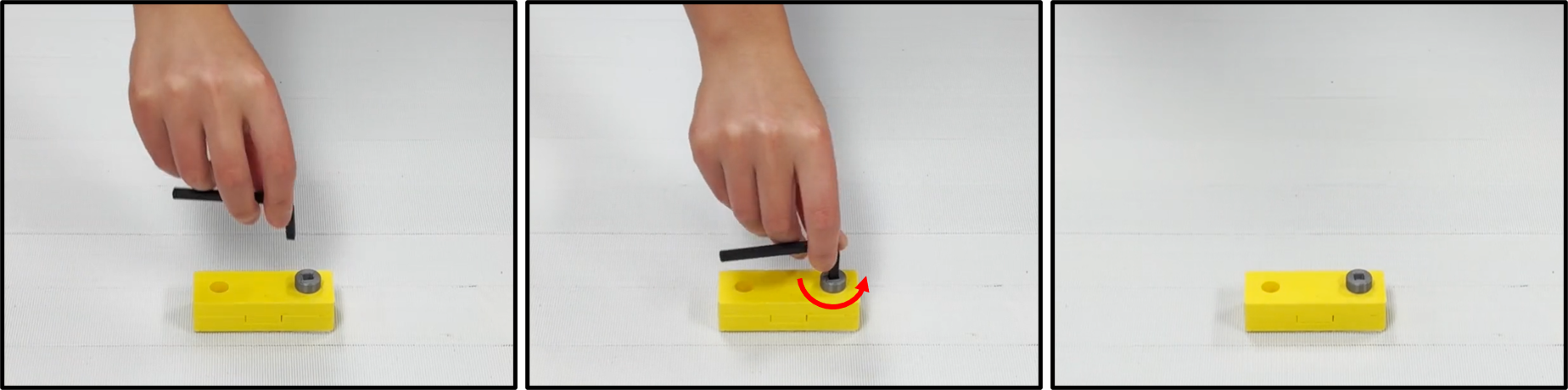}
    &
    \textbf{Task:} Loosen screw with Allen key

    \textbf{Skills:} Extrinsic contact control, insertion and alignment

    \textbf{Description:} A screw is placed into the screw slot of an assembly block in the workspace and is fully tightened. Use the Allen key grasped in hand to loosen the screw.

    \textbf{Success criteria:} The screw is loosened and can rotate freely when a small force is applied in an attempt to rotate the screw. The robot movement stops after terminating the loosening motion, moving away from, and breaking contact with the screw.
    \\


    \entryrule
    \includegraphics[width=\linewidth]{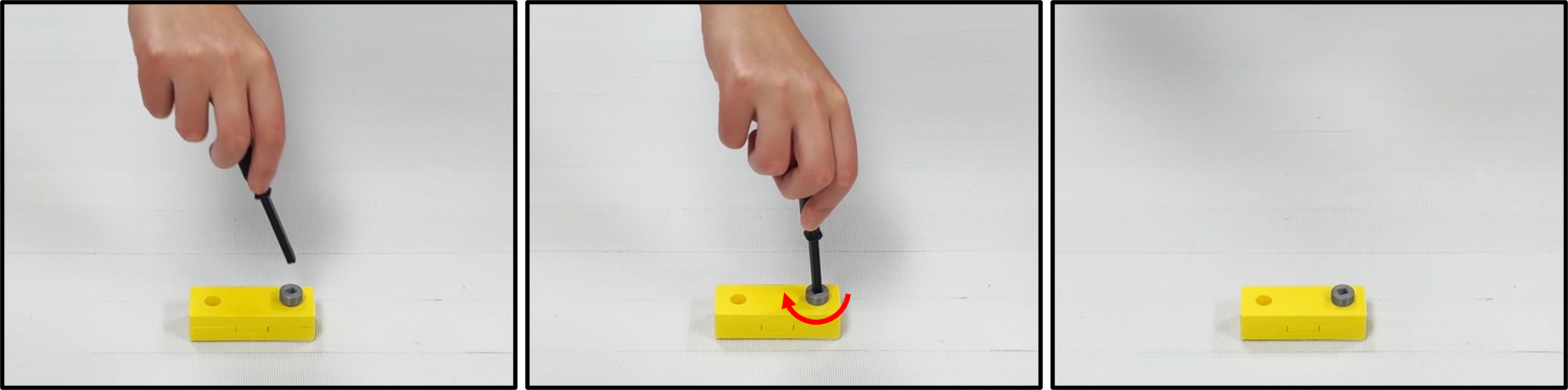}
    &
    \textbf{Task:} Tighten screw with screwdriver

    \textbf{Skills:} Extrinsic contact control, insertion and alignment

    \textbf{Description:} A screw is placed into the screw slot of an assembly block in the workspace, but the threads are unengaged. Use the screwdriver grasped in hand to tighten the screw.

    \textbf{Success criteria:} The screw is tightened and remains stationary when a small force is applied in an attempt to rotate the screw. The robot movement stops after terminating the tightening motion, moving away from, and breaking contact with the screw.
    \\


    \entryrule
    \includegraphics[width=\linewidth]{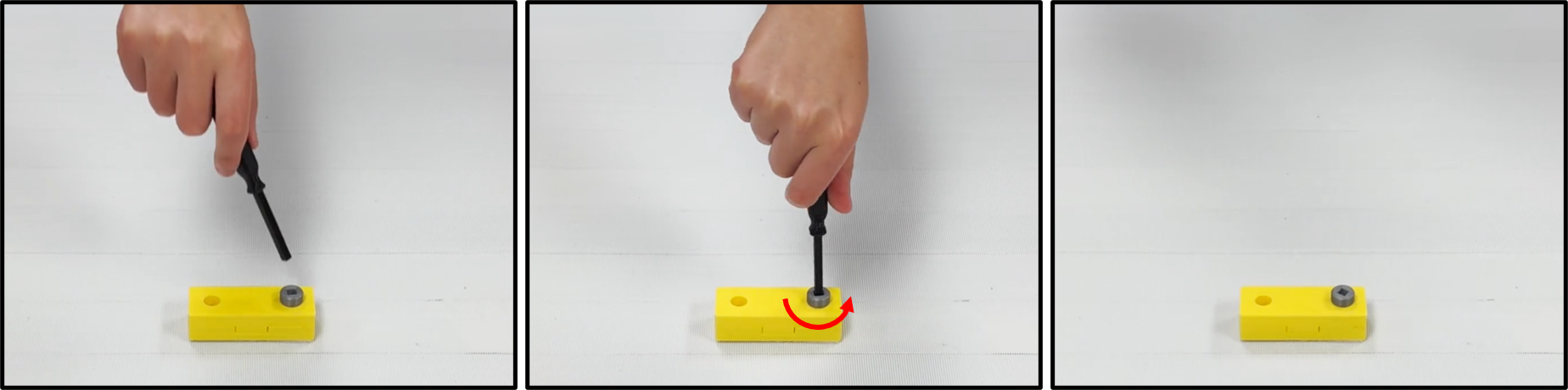}
    &
    \textbf{Task:} Loosen screw with screwdriver

    \textbf{Skills:} Extrinsic contact control, insertion and alignment

    \textbf{Description:} A screw is placed into the screw slot of an assembly block in the workspace and is fully tightened. Use the screwdriver grasped in hand to loosen the screw.

    \textbf{Success criteria:} The screw is loosened and can rotate freely when a small force is applied in an attempt to rotate the screw. The robot movement stops after terminating the loosening motion, moving away from, and breaking contact with the screw.
    \\


    \entryrule
    \includegraphics[width=\linewidth]{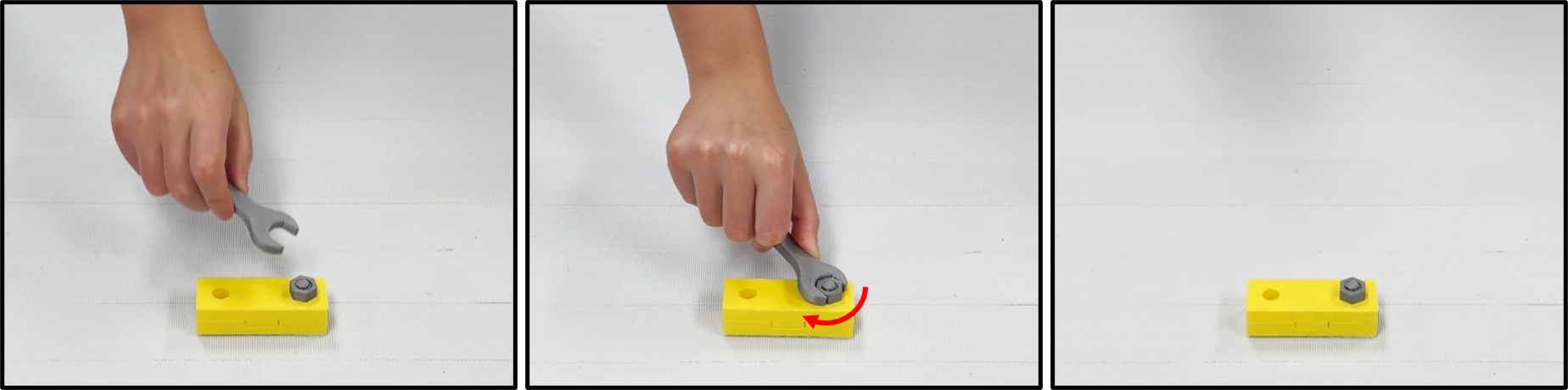}
    &
    \textbf{Task:} Tighten nut with wrench

    \textbf{Skills:} Extrinsic contact control, insertion and alignment

    \textbf{Description:} A nut is threaded onto a screw head in the workspace but is left untightened. Use the wrench grasped in hand to tighten the nut.

    \textbf{Success criteria:} The nut is tightened and remains stationary when a small force is applied in an attempt to rotate the nut. The robot movement stops after terminating the tightening motion, moving away from, and breaking contact with the nut.
    \\


    \entryrule
    \includegraphics[width=\linewidth]{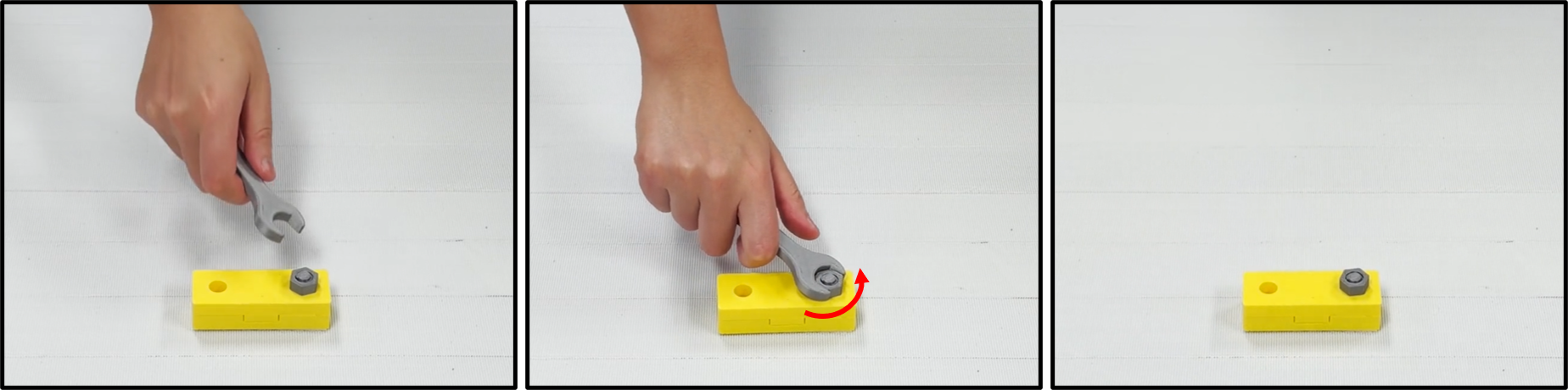}
    &
    \textbf{Task:} Loosen nut with wrench

    \textbf{Skills:} Extrinsic contact control, insertion and alignment

    \textbf{Description:} A nut is threaded onto a screw head in the workspace and is fully tightened. Use the wrench grasped in hand to loosen the nut.

    \textbf{Success criteria:} The nut is loosened and can rotate freely when a small force is applied in an attempt to rotate the nut. The robot movement stops after terminating the loosening motion, moving away from, and breaking contact with the nut.
    \\


    \entryrule
    \includegraphics[width=\linewidth]{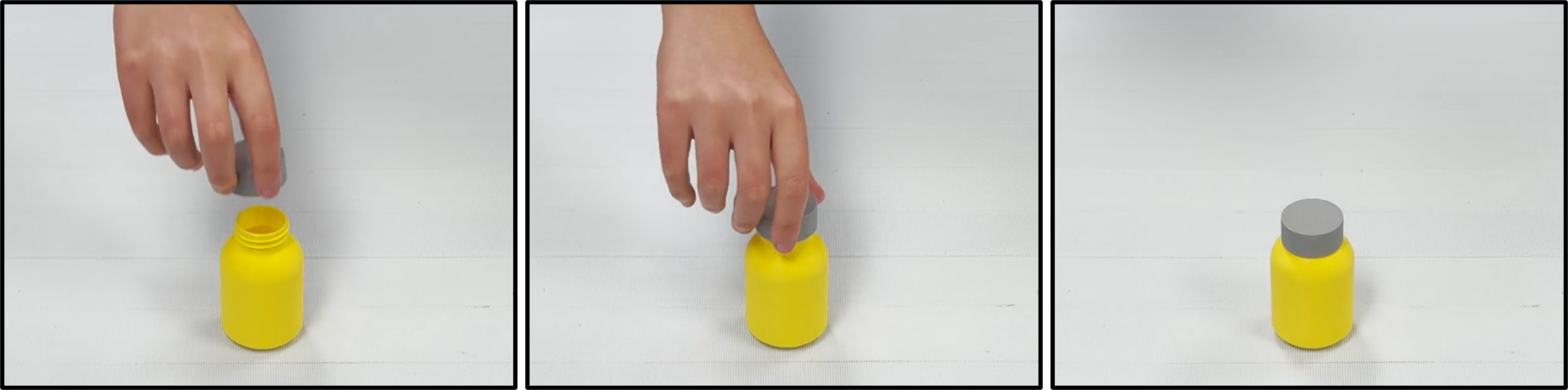}
    &
    \textbf{Task:} Put bottle cap onto bottle

    \textbf{Skills:} Insertion and alignment, articulated object manipulation

    \textbf{Description:} A bottle without a cap is placed in the workspace. Screw the bottle cap held in hand onto the bottle and tighten it.

    \textbf{Success criteria:} The bottle cap is tightened and remains stationary when a small force is applied in an attempt to rotate the cap. The robot movement stops after terminating the tightening motion, moving away from, and breaking contact with the task objects.
    \\







    \entryrule
    \includegraphics[width=\linewidth]{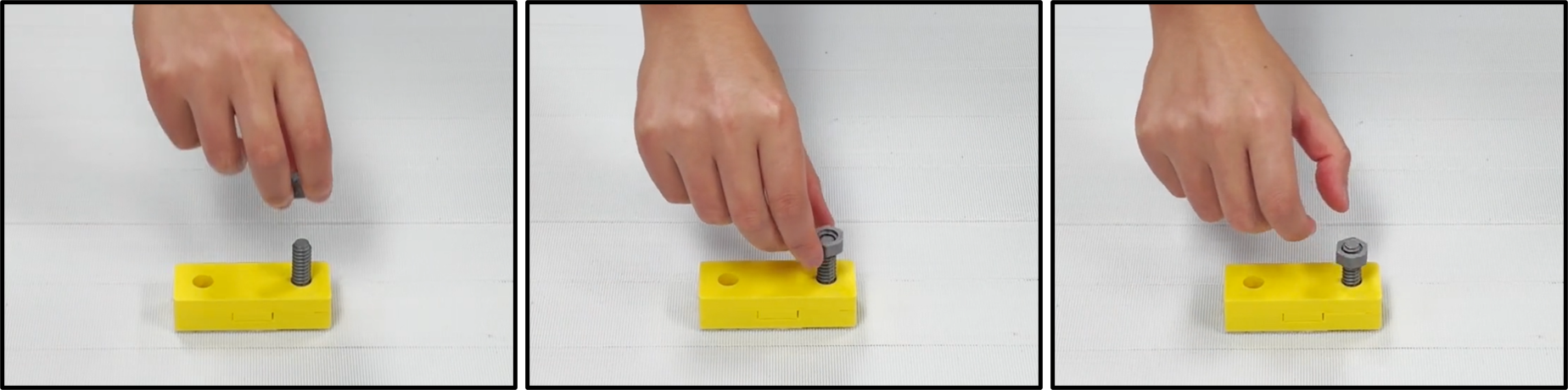}
    &
    \textbf{Task:} Thread nut onto screw

    \textbf{Skills:} Insertion and alignment, articulated object manipulation

    \textbf{Description:} Screw the nut held in hand onto the screw head placed in the workspace such that their threads fully engage.

    \textbf{Success criteria:} The nut and screw threads engage. The robot movement stops after terminating the rotation motion, moving away from, and breaking contact with the task objects.
    \\


    \entryrule
    \includegraphics[width=\linewidth]{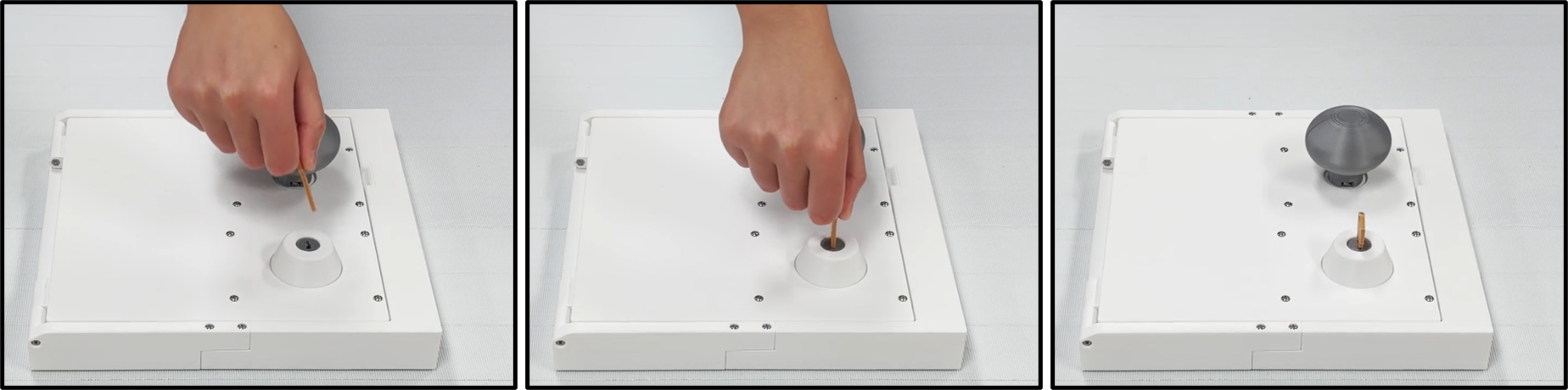}
    &
    \textbf{Task:} Insert key into lock

    \textbf{Skills:} Insertion and alignment

    \textbf{Description:} A lock is placed in the workspace. Insert the key held in hand into the lock.

    \textbf{Success criteria:} The key is inserted into the lock. The robot movement stops after terminating the insertion motion, moving away from, and breaking contact with the task objects.
    \\


    \entryrule
    \includegraphics[width=\linewidth]{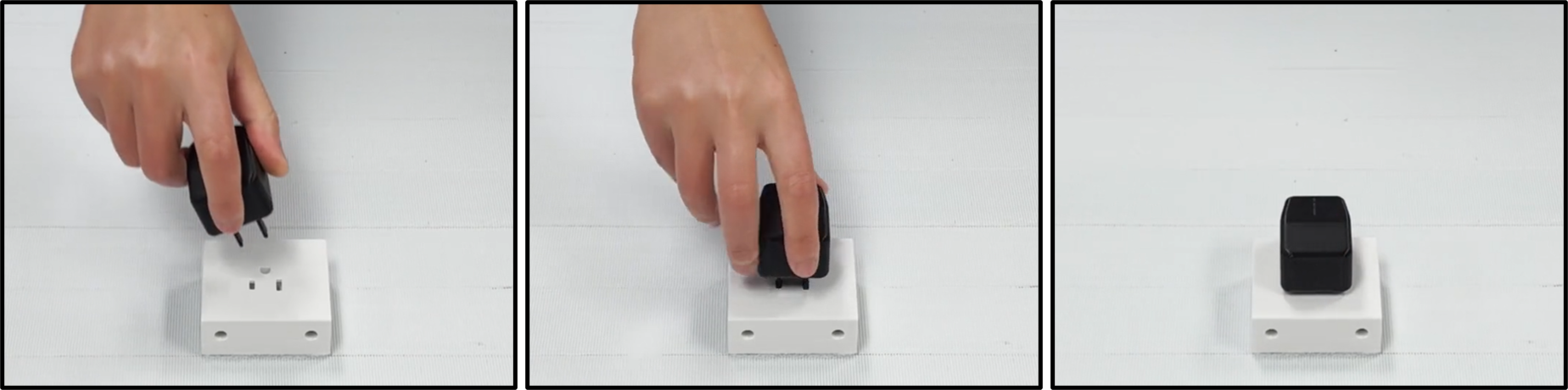}
    &
    \textbf{Task:} Plug in charger

    \textbf{Skills:} Insertion and alignment

    \textbf{Description:} An outlet is placed in the workspace. Fully insert the charger held in hand into the outlet.

    \textbf{Success criteria:} The charger is fully inserted into the outlet, such that it does not noticeably move when further pushed toward the outlet. The robot movement stops after terminating the insertion motion, moving away from, and breaking contact with the task objects.
    \\


    \entryrule
    \includegraphics[width=\linewidth]{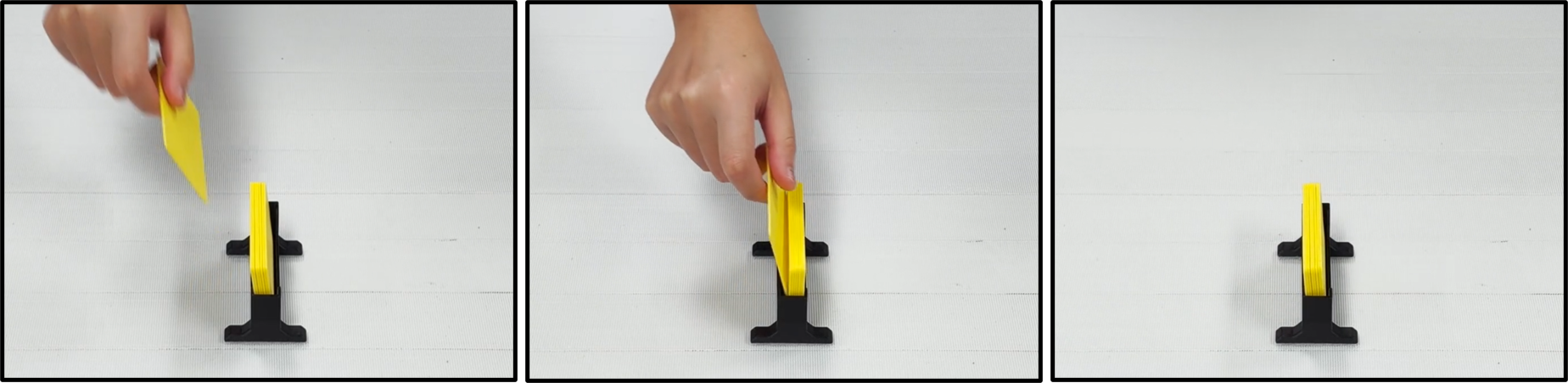}
    &
    \textbf{Task:} Place card into holder

    \textbf{Skills:} Placing under environmental constraints

    \textbf{Description:} A card holder that already contains other cards is placed in the workspace. Place the card held in hand into the card holder.

    \textbf{Success criteria:} The robot movement stops after placing the card in the card holder, with existing cards remaining in the card holder. The robot should not remain in contact with the task objects.
    \\


    \entryrule
    \includegraphics[width=\linewidth]{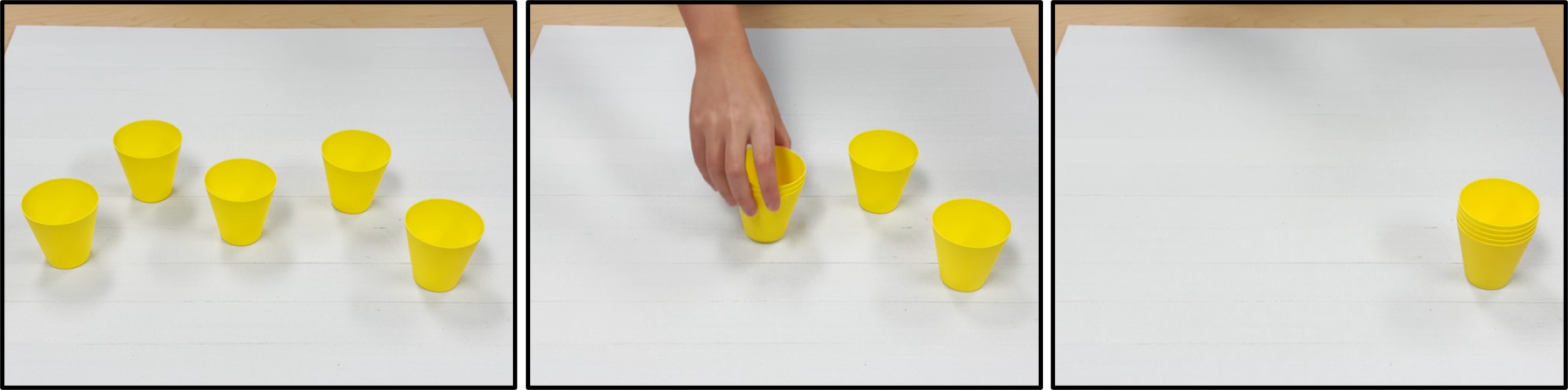}
    &
    \textbf{Task:} Stack cups

    \textbf{Skills:} Placing under environmental constraints

    \textbf{Description:} Multiple cups are scattered around the workspace. Collect the cups and place them inside one another to form a tight stack.

    \textbf{Success criteria:} The robot movement stops after all cups form a tight, upright stack (not knocked over). The robot should not remain in contact with the task objects.
    \\


    \entryrule
    \includegraphics[width=\linewidth]{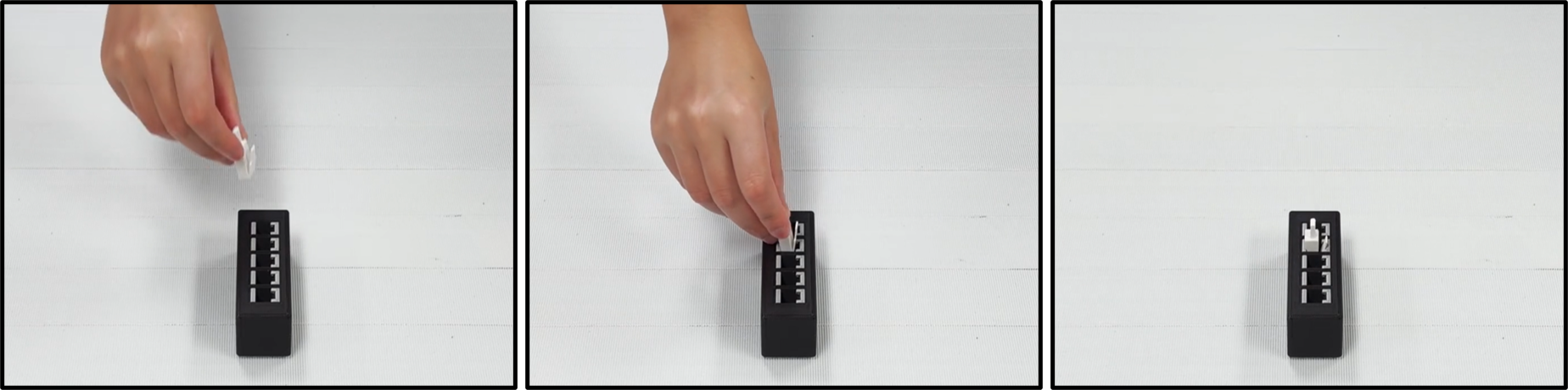}
    &
    \textbf{Task:} Plug in Ethernet

    \textbf{Skills:} Insertion and alignment

    \textbf{Description:} Fully insert the Ethernet plug held in hand into a socket placed in the workspace.

    \textbf{Success criteria:} The Ethernet plug is fully inserted into the socket with its latch engaged, such that the plug can’t be withdrawn by a light pull. The robot movement stops after terminating the insertion motion, moving away from, and breaking contact with the task objects.
    \\


    \entryrule
    \includegraphics[width=\linewidth]{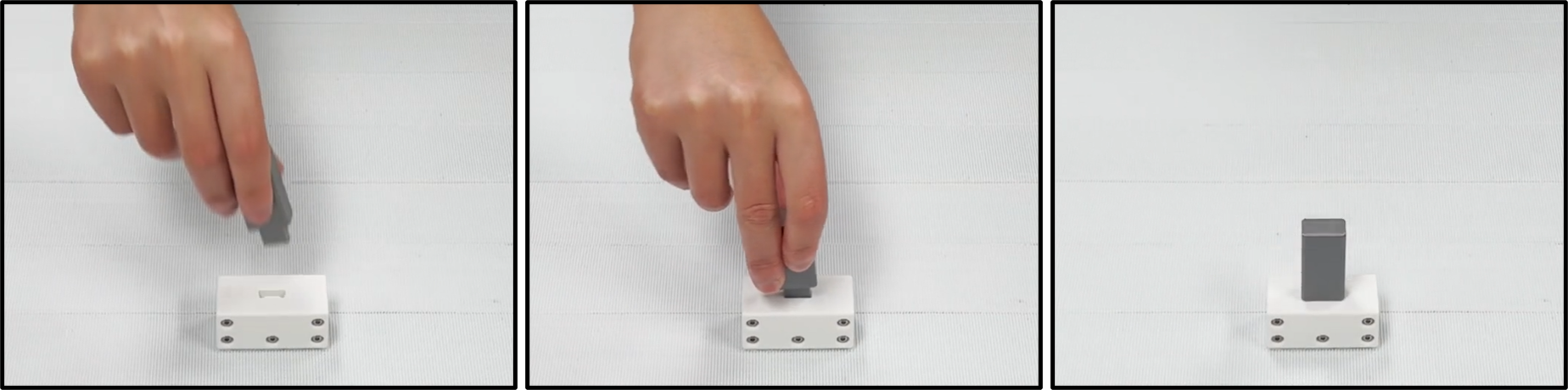}
    &
    \textbf{Task:} Plug in USB

    \textbf{Skills:} Insertion and alignment

    \textbf{Description:} Fully insert the USB stick held in hand into the USB slot placed in the workspace.

    \textbf{Success criteria:} The USB stick is fully inserted into the slot, such that it does not noticeably move when further pushed toward the slot. The robot movement stops after terminating the insertion motion, moving away from, and breaking contact with the task objects.
    \\


    \entryrule
    \includegraphics[width=\linewidth]{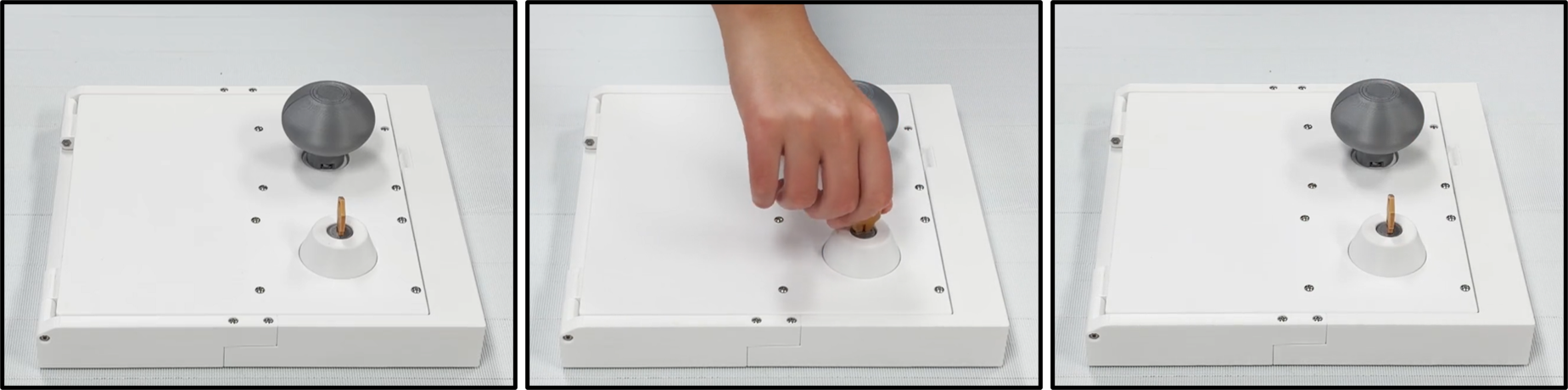}
    &
    \textbf{Task:} Turn key to unlock door

    \textbf{Skills:} Articulated object manipulation

    \textbf{Description:} A locked door module is placed in the workspace with a key already inserted into the lock. Turn the key to unlock the door.

    \textbf{Success criteria:} The key is fully turned such that the door can be opened. The robot movement stops after terminating the rotation motion, moving away from, and breaking contact with the task objects.
    \\


    \entryrule
    \includegraphics[width=\linewidth]{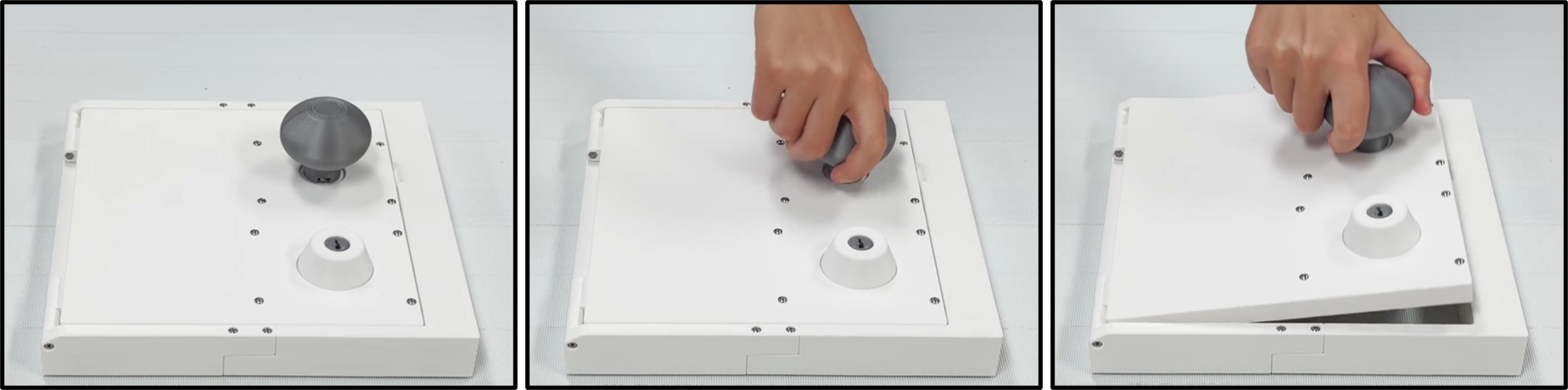}
    &
    \textbf{Task:} Turn doorknob to open door

    \textbf{Skills:} Articulated object manipulation

    \textbf{Description:} Turn the doorknob fully to unlock the door, then open the door.

    \textbf{Success criteria:} The door is opened by at least 5 degrees, such that it is visually obvious that the door has been unlocked.
    \\


    \entryrule
    \includegraphics[width=\linewidth]{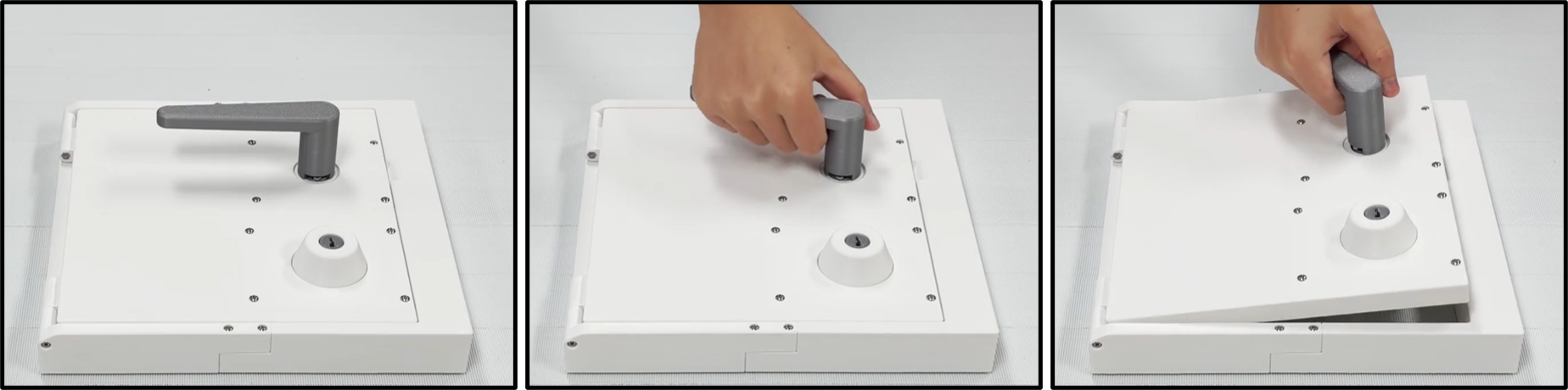}
    &
    \textbf{Task:} Turn door handle to open door

    \textbf{Skills:} Articulated object manipulation

    \textbf{Description:} Turn the door handle fully to unlock the door, then open the door.

    \textbf{Success criteria:} The door is opened by at least 5 degrees, such that it is visually obvious that the door has been unlocked.
    \\


    \entryrule
    \includegraphics[width=\linewidth]{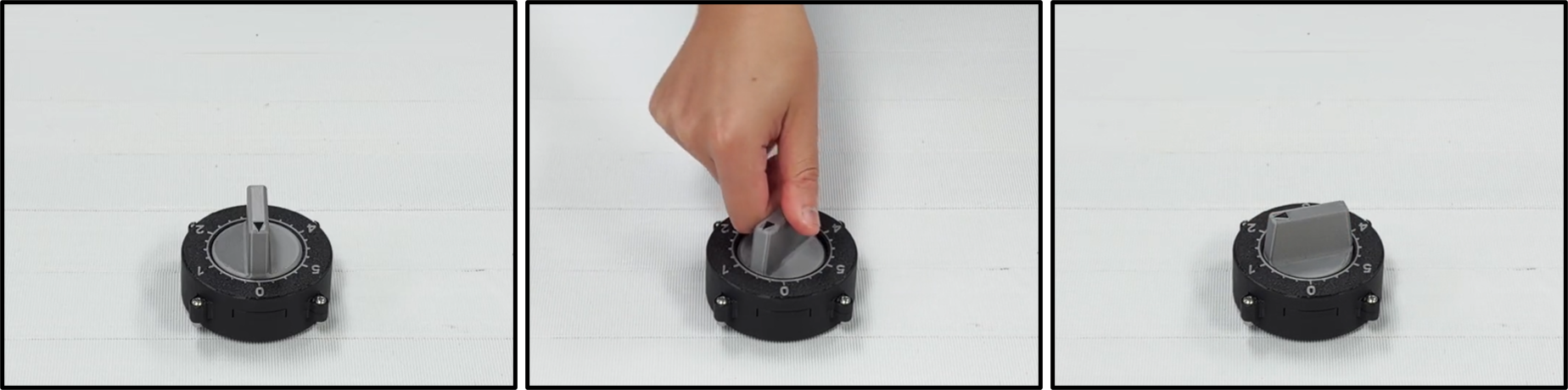}
    &
    \textbf{Task:} Turn dial to target position

    \textbf{Skills:} Articulated object manipulation

    \textbf{Description:} Operate the dial to turn it to the specified target position.

    \textbf{Success criteria:} The dial is turned to the target position. For dials that can be turned continuously, the final position should be within half a tick interval from the target position. The robot movement stops after terminating the rotation motion, moving away from, and breaking contact with the task objects.
    \\


    \entryrule
    \includegraphics[width=\linewidth]{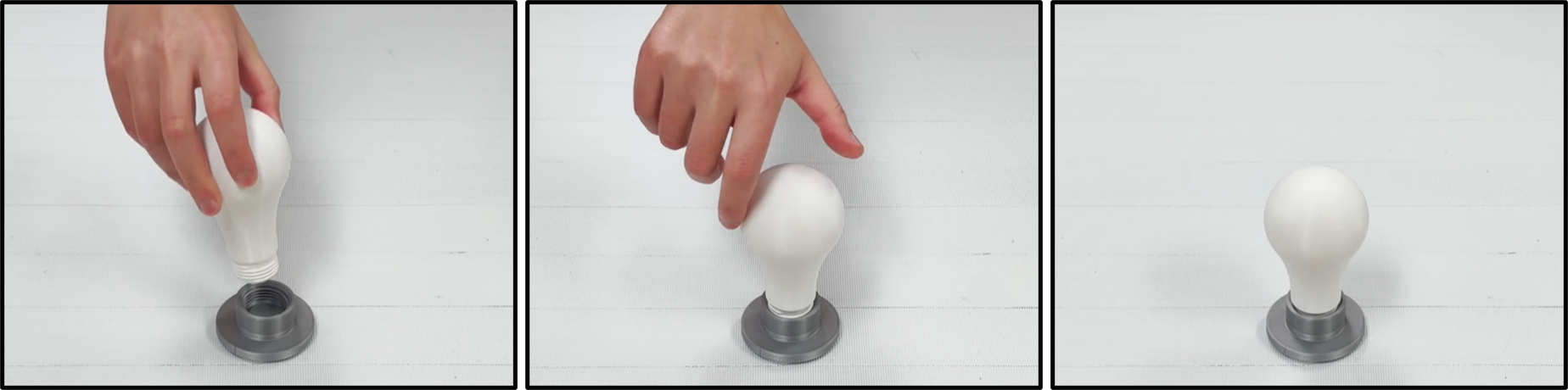}
    &
    \textbf{Task:} Install light bulb

    \textbf{Skills:} Insertion and alignment, articulated object manipulation

    \textbf{Description:} A light bulb socket is placed in the workspace. Screw the light bulb held in hand into the socket and tighten it.

    \textbf{Success criteria:} The light bulb is tightened and remains stationary when a small force is applied in an attempt to rotate it. The robot movement stops after terminating the tightening motion, moving away from, and breaking contact with the task objects.
    \\

  \end{xltabular}
\endgroup

\begingroup
  \footnotesize  
  \renewcommand{\arraystretch}{1.2}
  \setlength{\aboverulesep}{4pt}
  \setlength{\belowrulesep}{4pt}
  \renewcommand{\tabularxcolumn}[1]{m{#1}}  

  \begin{xltabular}{\textwidth}{@{} >{\centering\arraybackslash}m{0.4\textwidth} T @{}}

    \caption{Longer-horizon benchmark tasks that demand more dexterity and chain several skills together}
    \label{tab:additional-benchmark-tasks} \\
    \toprule
    \textbf{Task Execution Sequence} & \multicolumn{1}{c@{}}{\textbf{Task Specs}} \\
    \endfirsthead

    \toprule
    \textbf{Task Execution Sequence} & \multicolumn{1}{c@{}}{\textbf{Task Specs}} \\
    \endhead

    \bottomrule
    \endfoot
    \bottomrule
    \endlastfoot


    \entryrule
    \includegraphics[width=\linewidth]{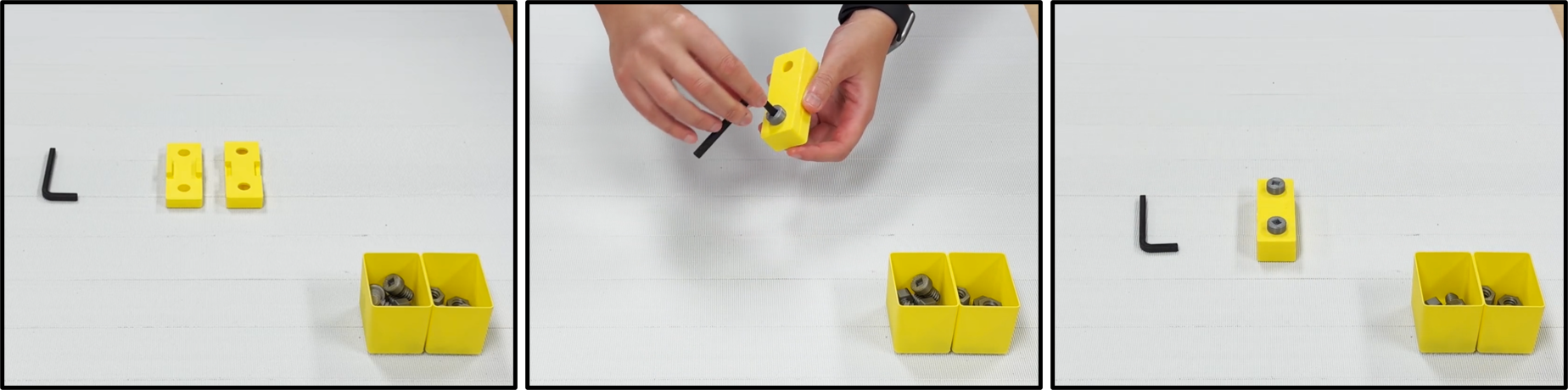} 
    &
    \textbf{Task:} Assemble toy block

    \textbf{Skills:} Grasping thin or small objects, grasping from clutter, in-hand repositioning, insertion and alignment, extrinsic contact control

    \textbf{Description:} Components of the block to be assembled, required tools, and bins of screws and nuts are placed in the workspace. Using the tools available, assemble the two halves of the block together with two screws and two nuts.

    \textbf{Success criteria:} The robot movement stops after fully assembling the block with screws tightened and placing the assembled block onto the table. The robot should not remain in contact with the block.
    \\


    \entryrule
    \includegraphics[width=\linewidth]{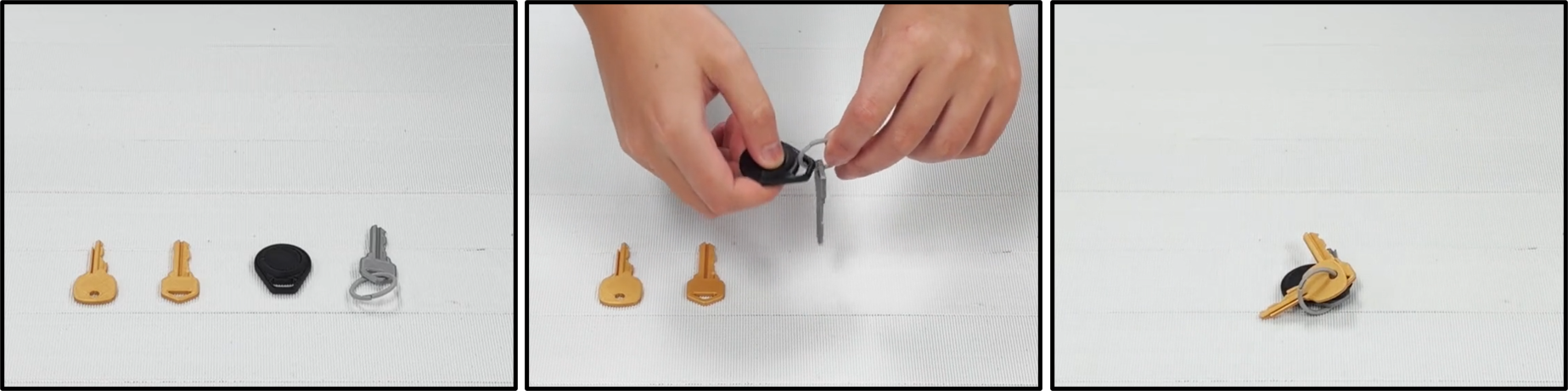} 
    &
    \textbf{Task:} Add keys to keyring

    \textbf{Skills:} Grasping thin or small objects, insertion and alignment

    \textbf{Description:} A keyring and several keys are scattered around the workspace. Add all loose keys onto the keyring.

    \textbf{Success criteria:} The robot movement stops after adding all loose keys to the keyring and placing the chained keys onto the table. The robot should not remain in contact with the task objects.
    \\


    \entryrule
    \includegraphics[width=\linewidth]{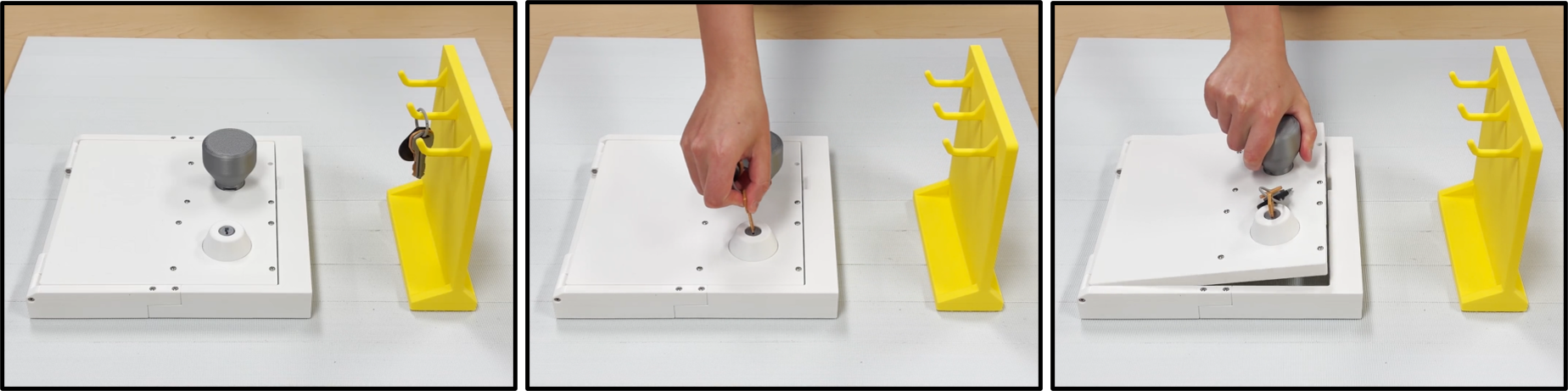} 
    &
    \textbf{Task:} Unlock and open door

    \textbf{Skills:} Grasping thin or small objects, grasping from clutter, in-hand repositioning, insertion and alignment, articulated object manipulation

    \textbf{Description:} A set of keys on a hanging board and a locked door are placed in the workspace. Retrieve the keys from the hanging board, use the correct key to unlock the door, then turn the doorknob to open the door.

    \textbf{Success criteria:} The door is opened by at least 5 degrees, such that it is visually obvious that the door has been unlocked.
    \\


    \entryrule
    \includegraphics[width=\linewidth]{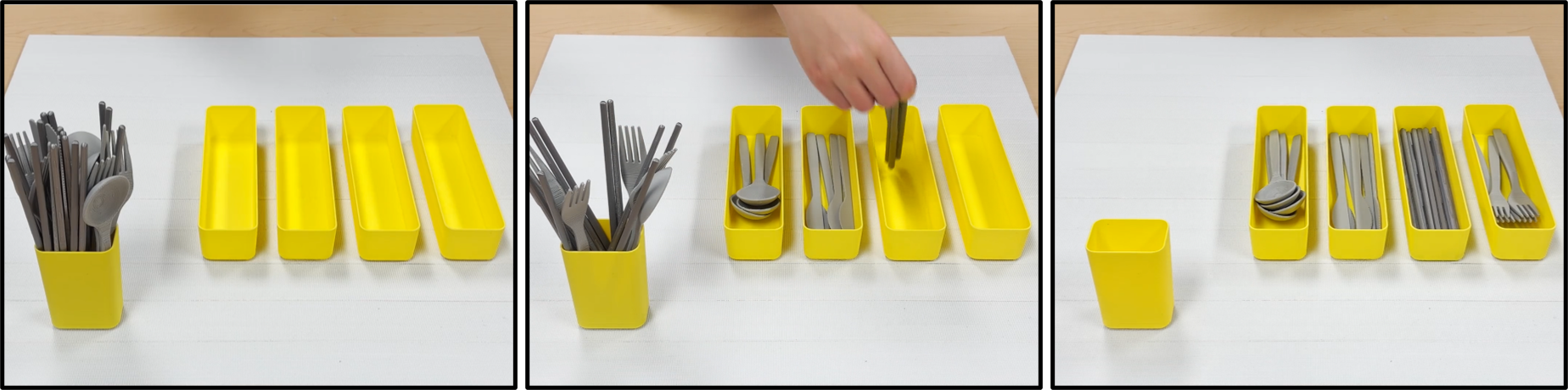} 
    &
    \textbf{Task:} Sort utensils

    \textbf{Skills:} Grasping thin or small objects, grasping multiple objects, grasping from clutter, placing under environmental constraints

    \textbf{Description:} A tight pile of mixed utensils is placed in the workspace or a storage bin in the workspace. Sort the utensils and place different types of utensils into their corresponding storage bins. Being able to hold multiple utensils simultaneously will greatly increase task completion efficiency.

    \textbf{Success criteria:} The robot movement stops after sorting every utensil in the pile into its corresponding bin. Only utensils of the same type should be in the same bin. The robot should not remain in contact with the task objects.
    \\


    \entryrule
    \includegraphics[width=\linewidth]{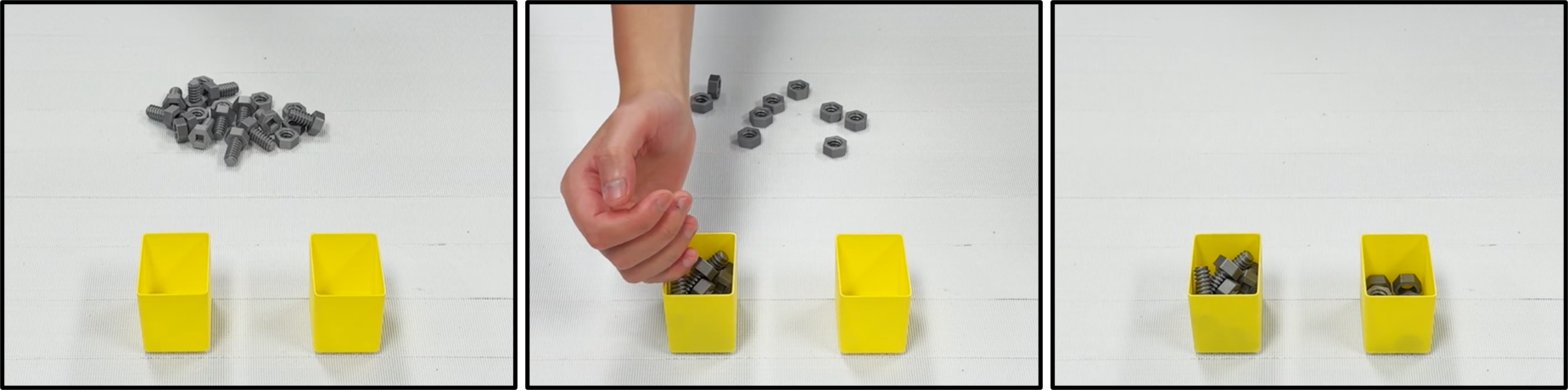} 
    &
    \textbf{Task:} Sort fasteners

    \textbf{Skills:} Grasping thin or small objects, grasping multiple objects, grasping from clutter, placing under environmental constraints

    \textbf{Description:} A tight pile of mixed fasteners is placed in the workspace or a storage bin in the workspace. Sort the fasteners and place different types of fasteners into their corresponding storage bins. Being able to hold multiple fasteners simultaneously will greatly increase task completion efficiency.

    \textbf{Success criteria:} The robot movement stops after sorting every fastener in the pile into its corresponding bin. Only fasteners of the same type should be in the same bin. The robot should not remain in contact with the task objects.
    \\

  \end{xltabular}
\endgroup
\twocolumn
\else
\twocolumn
{\footnotesize
\bibliographystyle{IEEEtran}
\bibliography{IEEEabrv,references}
}
\fi

\end{document}